\documentclass{iopjournal}

\usepackage{amssymb}
\usepackage{amsmath}
\usepackage{mathrsfs,amsfonts,amsthm,amsmath,amsbsy}

\usepackage{booktabs}
\usepackage{mwe}
\usepackage{caption}
\usepackage{subcaption}
\usepackage{graphicx}
\usepackage{subcaption}
\usepackage{todonotes}
\usepackage{float}
\usepackage{url}
\usepackage{multirow}
\usepackage{tabularx}
\usepackage{dsfont}
\usepackage{makecell}
\usepackage{rotating}
\usepackage{array}
\usepackage{ragged2e}
\usepackage{pbox}
\usepackage{arydshln}
\usepackage{siunitx}
\usepackage{adjustbox}
\usepackage{ragged2e}
\usepackage{pdflscape}
\usepackage{changepage} 
\usepackage{multicol}
\usepackage{xcolor}
\usepackage{soul}
\usepackage{lineno}

\usepackage{ulem}
\newcommand\rev[1]{\textcolor{black}{#1}}

\begin{document}

\articletype{Research paper} 

\title{\rev{Pure and Physics-Guided Deep Learning Approaches for Spatio-Temporal Groundwater Level Prediction}}

\author{Matteo Salis$^{1,3,*}$,
Gabriele Sartor$^1$,
Rosa Meo$^{1}$,
Stefano Ferraris$^{2}$,
Abdourrahmane M. Atto$^{3}$ 
}

\affil{$^1$Computer Science Department, University of Turin, Turin, Italy}

\affil{$^2$Interuniversity Department of Regional and Urban Studies and Planning, Politecnico di Torino and University of Turin, Turin, Italy}

\affil{$^3$LISTIC Laboratory, Université Savoie Mont Blanc, Annecy-le-vieux, France}

\affil{$^*$Corresponding author.}








            


\email{matteo.salis@unito.it or salis.matteo98@gmail.com (corresponding author)}



\begin{abstract}
\justifying
\rev{Groundwater represents a key element of the water cycle, yet it exhibits intricate and context-dependent relationships that make its modeling a challenging task. Theory-based models have been the cornerstone of scientific understanding. However, their computational demands, simplifying assumptions, and calibration requirements limit their use. In recent years, data-driven models have emerged as powerful alternatives. In particular, deep learning has proven to be a promising approach for its design flexibility and ability to learn complex relationships directly from the data without requiring extensive domain information.}

\rev{\noindent We proposed an attention-based pure deep learning model, named STAINet, to predict weekly groundwater levels of 28 piezometers in the Cuneo and Torino provinces in Piedmont (Italy), leveraging both irregular groundwater time series and weather image sequences. With the aim of enhancing the model's trustworthiness and generalization ability, we merged the theory and data-driven approaches by considering physics-guided strategies to inject the groundwater flow equation into the model. Firstly, we restructured the tail of the architecture to predict the three terms of the governing equation, named the autoregressive, diffusion, and residual components -- we thus obtained the PSTAINet-IB.
Then, we further injected physics priors by adding loss terms related to the estimated equation components, obtaining the PSTAINet-ILB model.
Lastly, we developed the PSTAINet-ILRB by imposing a loss term specific to the residual component, which forces the groundwater recharge to occur within a specific zone, noted as the groundwater body recharge zone, which is identified by domain experts.}

\rev{The four models were evaluated using RMSE, NBIAS, MAPE, NSE, and KGE, both by feeding true lagged values as input and by iterating their own predictions (rollouts) over the whole test set. 
With the physics-guided approach, we obtained better results, in particular, the PSTAINet-ILB model performed the best, achieving remarkable test performance (median MAPE 0.16\%, KGE 0.58 in the rollout setting), and generating equation components in line with domain experts' expectations.}
\end{abstract}

\justifying

\section{Introduction}
\rev{Accounting for water resources is pivotal for planning water policies and thus to meet human water needs sustainably \cite{UNUnited2023,usgs_gw2001}.
In the hydrological cycle, groundwater represents the second most relevant component, storing about 30\% of the world’s freshwater resources, just behind glaciers and ice caps, which account for 68.7\% ~\cite{ShiklomanovWorld1998}.
Furthermore, groundwater represents a more stable source of freshwater than rivers and lakes, whose storage responds promptly to the weather conditions~\cite{FamigliettiGlobal2014}.}

\rev{To model and account for water resources, hydrologists have largely relied on process-based models that map the cause-effect relationship among hydrological variables~\cite{Sabzipour2023,Tripathy2024}.
The drawback of such theory-based models is that they often rely on simplifying assumptions, either because of incomplete knowledge of certain processes or to make the model computationally tractable.
Furthermore, many internal parameters are usually required and thus estimated for the specific application, a procedure usually named model calibration.
The calibration of a model requires many geophysical data of the aquifer under study and a deep knowledge of the hydro-geophysical processes involved.
However, it is not always feasible to obtain such data, either because of the expensive measurements or observational restrictions, making additional simplifying assumptions necessary and thus limiting models' performance and applicability in real contexts~\cite {gupta_2014,karpatne_2017,Clark2017,BolsterRecent2019,willard_2022}.}

In recent years, deep learning models have achieved striking results in modeling many natural phenomena, including water resources~\cite{TaoGroundwater2022,zanoni_2022,andrychowicz_2023,lamGraphCastLearningSkillful2023}.
\rev{However, especially when the aim is forecasting (both nowcasting and long-term projections), relying only on past observations may not be effective} because the process might evolve following unseen patterns -- a situation well illustrated by the inductivist turkey\footnote{The inductive turkey (also known as the turkey illusion) exemplifies the problem of induction, originally described by Bertrand Russell in The Problems of Philosophy. It tells of a turkey that, being fed daily, infers that this pattern will persist indefinitely—until, on the day before Thanksgiving, it is slaughtered, exposing the weakness of its assumption}.

\rev{Furthermore, low-quality and scarce data may provide only a partial representation of the underlying processes, making the models trained on such incomplete data focused on spurious associations, and thus preventing them from good generalization abilities~\cite{karpatne_2017,daw_2022}.
Unfortunately, hydrological phenomena (but more generally all environmental processes) suffer from this data limitation, because of sensor failures or limited sensor coverage, but also because of complex and non-stationary dynamics that evolve in time, influenced by exogenous factors, such as, for example, climate change.}

Given that theory-based and data-driven approaches have their own advantages and disadvantages, a consistent number of researchers have proposed a mixed strategy to leverage the strengths of both, i.e., theory-guided data science~\cite{karpatne_2017,daw_2022,willard_2022}, and specifically for physics and machine learning, the physics-guided (or informed) machine learning~\cite{raissi_2019a,reichstein_2019,karniadakis_2021}.
The idea is to inject prior scientific knowledge, a bias (or guidance), into a data-driven model, and to this aim, several approaches have been proposed, and concerning the physics-guided literature, three main pathways \rev{have been defined in} \cite{karniadakis_2021} and named observational bias, inductive bias, and learning bias. 
These three approaches differ in terms of where the bias is applied in the modeling pipeline, i.e., in the data, in the model's computations, or in the learning algorithm, respectively.


\rev{In this work, we aimed to develop models to predict the weekly groundwater level (GWL) measured by 28 piezometers between the Provinces of Cuneo and Torino in Piedmont (Italy).
In particular, we compared pure data-driven and physics-guided deep learning approaches leveraging the groundwater flow equation and using an inductive and a mixed inductive-learning bias strategy. 
To the best of our knowledge, this is the first study to test different physics-guided deep learning approaches in a real-world scenario, rather than on simulated groundwater data.}

\rev{For our models, we considered the groundwater levels measurements of the 28 piezometers from 2001-01-01 to 2023-12-31 as the target, while as input we adopted an autoregressive component (i.e. past historical GWL values) and an exogenous term represented by the weather data over the whole Region of Interest (ROI) in the form of a sequence of raster images (i.e. a video) at 0.1° spatial resolution.}

\rev{
\subsection{Physics Injection Strategies}
In the last decade, an increasing number of research studies aim to inject prior knowledge into data-driven models~\cite{karpatne_2017,reichstein_2019}. 
Especially in the physics-guided machine learning literature, three main strategies have emerged to incorporate physics knowledge, differing in the stage at which the physics constraints are introduced, namely through the data, the model architecture, or the learning process~\cite{karniadakis_2021,meng2025physics,ruiPGDLsurvey2025}. Following the terminology adopted in \cite{karniadakis_2021}, we refer to these approaches as observational, inductive, and learning biases, respectively.}
 
\paragraph{Observational Bias}
The observational bias strategy is conceptually the simplest, \rev{as it involves providing sufficient data that} truly and completely represent the underlying physical process.
\rev{Indeed, data are the cornerstone of machine learning and can be used as a weak mechanism to inject physics into the model during training~\cite{karniadakis_2021}.
However, this strategy requires a large amount of high-quality data, whereas, as already pointed out, environmental observations are often scarce and expensive to obtain. 
As an alternative, theory-based simulations could be employed in place of measurements or in addition to them as a data augmentation strategy~\cite {wu_2024}. 
Nevertheless, this approach may be computationally inefficient, as it requires running the underlying physics model first to generate additional data. Moreover, when simulated data substantially outnumber observations, the resulting model may effectively act as a surrogate or emulator of the theory-based model, thereby inheriting its assumptions and limitations.}


\rev{
\paragraph{Inductive Bias}
The second strategy entails using an inductive bias, a general concept in machine learning that refers to the assumptions incorporated into a model during its design.
These assumptions generally concern the structure of the data and the relation to be learned, and they are the basis upon which the model performs induction ~\cite{mitchell1997}. 
For example, Convolutional Neural Networks (CNNs) incorporate inductive biases tailored to grid-structured data, such as images, by exploiting locality, spatially compact patterns, and translation invariance through parameter sharing~\cite{lecun1998,GoodfellowDeep2016}.
Differently, Graph Neural Networks (GNNs) are designed for graph-structured data, where relationships between entities (nodes) are encoded by edges. GNNs assume that relevant information is propagated through neighboring nodes and that the representation should be invariant to permutations of node ordering~\cite{goriGNN2005,zhouGraph2020}.}

\rev{Even though the concept of inductive bias is very broad and encompasses designing decisions made at every stage of the modelling pipeline, in the physics-guided literature, the term inductive bias has been used to describe those approaches focused mainly on developing specialized architectures that implicitly embed any prior physics knowledge~\cite{karniadakis_2021,reichstein_2019}.
These can be specific layers or operations performed inside the network (such as convolution in CNNs), leaving the learning algorithm untouched.
For example, authors in ~\cite{debezenac_2019} developed a neural network to predict sea-surface temperature, injecting the advection-diffusion equation. 
In particular, they developed a specific architecture that, through a warping scheme, predicts future sea-surface temperature images based on an autoregressive component and the motion field estimate. 
The proposed warping scheme is built upon the solution of a general class of partial differential equations that is defined by a radial basis function kernel.
Another example is that of authors in~\cite{darbon_2021}, who developed specific architectures representing the viscosity solutions of two sets of Hamilton–Jacobi partial differential equations.}

Even though this strategy could constrain the model to strictly adhere to the domain prior, it requires ad hoc and \rev{elaborate implementations, and usually, it is adopted in tasks characterized by relatively simple and well-defined physics~\cite{karniadakis_2021}.}

\rev{
\paragraph{Learning Bias}
The third strategy describes those approaches aiming to inject prior physics knowledge by focusing on the learning algorithm, independently of the architecture adopted. 
From a broader perspective, acting on the learning algorithm, it could still be considered an inductive bias.
However, the idea behind these approaches is completely different from the previously described inductive bias strategy, because the learning bias is a general strategy to both fit the data and adhere to a physics prior, regardless of the adopted architecture.
This approach is thus suited for those cases in which the domain knowledge is not complete or uncertain, and soft constraints derived from a prior are preferable~\cite{reichstein_2019,raissi_2019a,karniadakis_2021}.
Moreover, when a governing equation contains some unknown parameters, these could be jointly estimated with the model's own parameters during the training process.
After the training, analysing those estimated parameters could then offer a way to investigate the rationale of the model from a physics perspective.}

\rev{
More practically, imposing a learning bias involves augmenting the data loss term ($\mathcal{L}_{\textit{data}}$), which measures the discrepancy between observations and predictions, with additional regularization terms ($\mathcal{L}_{\textit{eq}}$) derived from the residuals of the governing equations\footnote{Governing equations generally describe the evolution of a physical process and are commonly expressed as Partial Differential Equations (PDEs).} evaluated on the model outputs.
Within this framework, the model is trained to simultaneously fit the observed data and satisfy the physics constraints imposed by prior knowledge, with weighting coefficients that can be adjusted to emphasize one objective over the other.
Although this flexibility allows the relative importance of data and physics to be tuned, it may also lead to solutions that do not fully satisfy the governing equations if too much importance is given to the data fitting objective.}

\rev{
This approach was proposed for the first time by \cite{raissi_2019a}, who demonstrated that neural networks constrained by PDE residuals, and thanks to universal approximation capabilities, can solve forward and inverse problems in several physics systems, including fluid flow, reaction-diffusion processes, and nonlinear wave propagation.
In the literature, the specific name Physics-Informed Neural Networks (PINNs) has been adopted to describe all neural networks specifically designed to approximate the solutions of partial differential equations~\cite{raissi_2019a,cuomo_2022a}.
While the term physics-guided has remained more general, embracing different approaches to inject physics into machine learning models.}

Even though all three strategies have been demonstrated to be effective, the inductive and learning biases are the most adopted in recent years~\cite{reichstein_2019,karniadakis_2021,daw_2022}, because of the possibility of introducing hard constraints \rev{(inductive bias), and for the flexibility of using different architectures and leveraging prior knowledge even if incomplete (learning bias).
Furthermore, it is possible to combine those approaches and build, in this way, a mixed strategy, leveraging both the advantages of the inductive and learning biases.}


\subsection{Groundwater principles}
\label{sec:gw_princ}

Our study addressed shallow groundwater bodies, also named phreatic aquifers or unconfined aquifers, which are portions of the terrain saturated with water and directly influenced by surface activities, and whose upper surface, denoted as the water table, is at atmospheric pressure. 
To quantify groundwater resources, the groundwater level (GWL) is frequently adopted, and it represents the distance between the reference datum and the water table.\\
Groundwater dynamics are usually described by using Darcy's law, which states that the specific discharge $q_x$ in the $x$-direction is proportional to the gradient of the total hydraulic head $h$ multiplied by the saturated hydraulic conductivity $K_x$.
More specifically, the specific discharge $q_x$ is the volume rate of flow per unit of area, and the hydraulic head is the potential mechanical energy per unit weight of fluid.
The hydraulic head is the sum of the gravitational head, given by the elevation of the point above an arbitrary horizontal datum, and the pressure head, which is the pressure of the fluid at the point divided by the specific weight of the fluid.
In the case of an unconfined aquifer, whose upper surface is at atmospheric pressure, the hydraulic head is equivalent to the water table height (i.e., groundwater level).
Therefore, it is possible to write Darcy's law as in Equation~\ref{eq:darcy}:

\begin{equation}
    q_x = - K_x \frac{\partial h}{\partial x}
\label{eq:darcy}
\end{equation}

Combining Darcy's law with the mass conservation principle, it is possible to derive a general PDE, a form of diffusion equation, for the saturated groundwater flow in the three spatial dimensions $(x,y,z)$, as in Equation~\eqref{eq:gw_pde_3d}.
The term $S_s$ represents the specific storage\footnote{As common in hydrology, to describe a unit of measurement, we adopted the terminology of the fundamental physical dimensions, thus length [L] and time [T] -- in our case meters [m] and weeks [w]} [L\textsuperscript{-1}], that is the volume of water entering/leaving a unit volume per unit variation in hydraulic head, and $K_x$, $K_y$, $K_z$ are the conductivities in the specific directions, all expressed in [LT\textsuperscript{-1}].

\begin{equation}
    \frac{\partial h}{\partial t} S_s = K_x\frac{\partial^2 h}{\partial x^2} + K_y\frac{\partial^2 h}{\partial y^2} + K_z\frac{\partial^2 h}{\partial z^2}
\label{eq:gw_pde_3d}
\end{equation}

However, attention is typically directed toward horizontal flow along the aquifer plane of coordinates $(x,y)$.
Considering a simplified formulation and defining the transmissivity $T = H \cdot K$ [L\textsuperscript{2}T\textsuperscript{-1}] and the storage coefficient $S = H \cdot S_s$ [1], in which $H$ [L] is the saturated thickness of the aquifer and $K$ is the horizontal conductivity, it is possible to define the groundwater flow equation for the horizontal flow (hereafter groundwater flow equation) as in Equation~\eqref{eq:gw_pde}:

\begin{equation}
    \frac{\partial h}{\partial t} S = T \left(\frac{\partial^2 h}{\partial x^2} + \frac{\partial^2 h}{\partial y^2}\right)
\label{eq:gw_pde}
\end{equation}

We reformulated Equation~\eqref{eq:gw_pde}, moving $S$ to the right-hand side obtaining Equation~\eqref{eq:gw_pde_new}. The resulting quantity $\frac{T}{S}$ is related to the ease with which water moves through the soil, and it is expressed in [L\textsuperscript{2}T\textsuperscript{-1}]. 

\begin{equation}
    \frac{\partial h}{\partial t} = \frac{T}{S} \left(\frac{\partial^2 h}{\partial x^2} + \frac{\partial^2 h}{\partial y^2}\right)
\label{eq:gw_pde_new}
\end{equation}

\section{Related Works}

An increasing number of studies have explored the implementation of physics-guided approaches across a wide range of domains, including healthcare~\cite{arzani_2021a,kissas_2020}, computer vision~\cite{banerjee_2025}, geoscience~\cite{debezenac_2019,toms_2020,tsai_2021,shen_2023}, weather~\cite{das_2024,song_2024}, air-pollution~\cite{kim_2024}, and also hydrology~\cite{jiang_2020,tsai_2021,cuomo_2023a,secci_2024a}.


Focusing on hydrology, authors in \cite{jiang_2020} adopted an inductive bias strategy to develop a physics-guided neural network to model runoff across the conterminous United States.
In particular, they proposed an architecture comprising two consecutive blocks.
The first is composed of a restructured version of an RNN layer, named P-RNN, which embeds the spatially lumped hydrological model described by two discrete state-space representations.
Then the second block, made of usual 1D convolutional layers, takes the original input along with the P-RNN output to model the residual component with respect to the physics. 
The proposed physics-guided model achieved better performance compared to pure deep learning implementations.

A different strategy was followed by authors in \cite{cuomo_2023a,secci_2024a} who adopted learning biases to estimate simulated groundwater flow data. 

In the first study~\cite{cuomo_2023a}, a multilayer perceptron (MLP) was trained to estimate the solution of the groundwater flow equation in one and two-dimensional cases in the presence of pumping wells, \rev{incorporating the residual of the governing PDE and the boundary conditions into the loss function.}
The trained PINN was revealed to be more accurate than the finite difference method adopted as a comparison.

In the second study~\cite{secci_2024a}, the authors aimed to estimate the phreatic surface (i.e., water table) and the piezometric heads in a vertical cross-section of a homogeneous and isotropic aquifer and a heterogeneous and anisotropic one.
To this aim, they proposed a physics-informed neural network made of two sub-networks, each implemented as an MLP.
The first is responsible for estimating the piezometric head value along the vertical cross-section over time, while the second network is designed to predict the phreatic surface. 
In this study, MODFLOW~\cite{hughes_2017} simulations were performed and considered as the true data in the training process, along with the residuals of the groundwater flow equation.
Even though the architecture appears to be intricate, the authors achieved better performance than a pure deep learning architecture without the physics loss. Furthermore, they highlighted the ability of PINNs to work in mesh-free domains (i.e., without discretization) and that the incorporation of physics constraints can dramatically reduce the number of required observations for training, suggesting the PINNs' potential to generalize well also in data-scarce environments.

Even if the mentioned studies reveal the potential of physics-guided deep learning approaches, our application context is different. 
Indeed, in \cite{debezenac_2019}, the adopted target (sea surface temperature) is dense in space and time, and this contrasts with our scenario, where groundwater level observations are spatially sparse.

\noindent On the other hand, even if the PINN approach is flexible, it is usually implemented for theoretical analyses, and thus models are frequently developed to take as input only the coordinates of the system (spatial, temporal, or both), and less frequently, exogenous and real inputs are considered.
Furthermore, feeding multi-modal data (e.g., video and tabular data) to the models in this framework is non-immediate, and a careful design is needed.

In our study, to inject the groundwater flow equation, we implemented and tested two of the above-presented strategies.
Firstly, we introduced an inductive bias into the pure deep learning model by making the neural network estimate the different components of the governing equation. Afterwards, we added learning biases into the loss function to further condition the model.

\rev{Specifically, with the learning bias, we aimed to enforce the groundwater flow equation over the discretized domain of the ROI. However, to compute the residual with respect to the groundwater flow equation, we needed a corresponding grid of predictions.
Instead of adopting post-processing (e.g., interpolation) methods, we developed a neural architecture able to predict at a variable number of points. Thus, for the data loss $\mathcal{L}_{\textit{data}}$, the models predict only on the 28 sensor locations, while for the physics loss $\mathcal{L}_{\textit{eq}}$, the models predict a regular grid directly.
To develop such models,} we leveraged the attention mechanism~\cite{VaswaniAttention2017}. 
Indeed, attention maps \textit{values} indexed by \textit{keys} to new values indexed by \textit{queries} (in a database fashion), and thus offers a straightforward solution to leverage input data indexed by spatio-temporal coordinates (keys) to generate predictions at an arbitrary number of new locations (queries).
Other works, such as \cite{andrychowicz_2023,song_2024}, leverage spatially sparse data as well; however, in the first case~\cite{andrychowicz_2023}, the authors interpolated data into a grid, while in the second case~\cite{song_2024}, the authors adopted an MLP to perform the upsampling, and thus the number of input and output locations must be fixed in advance.

It is worth noting that developing models that automatically predict at different locations than the input ones is in contrast with traditional approaches, which require input features to be co-located with the target variable and then use post hoc interpolation methods to generate predictions in new locations. 

\section{Data}
\label{sec:pi_data}

\begin{figure*}[h]
    \centering
    \begin{subfigure}[c]{0.5\textwidth}
        \centering
        \includegraphics[width=\textwidth]{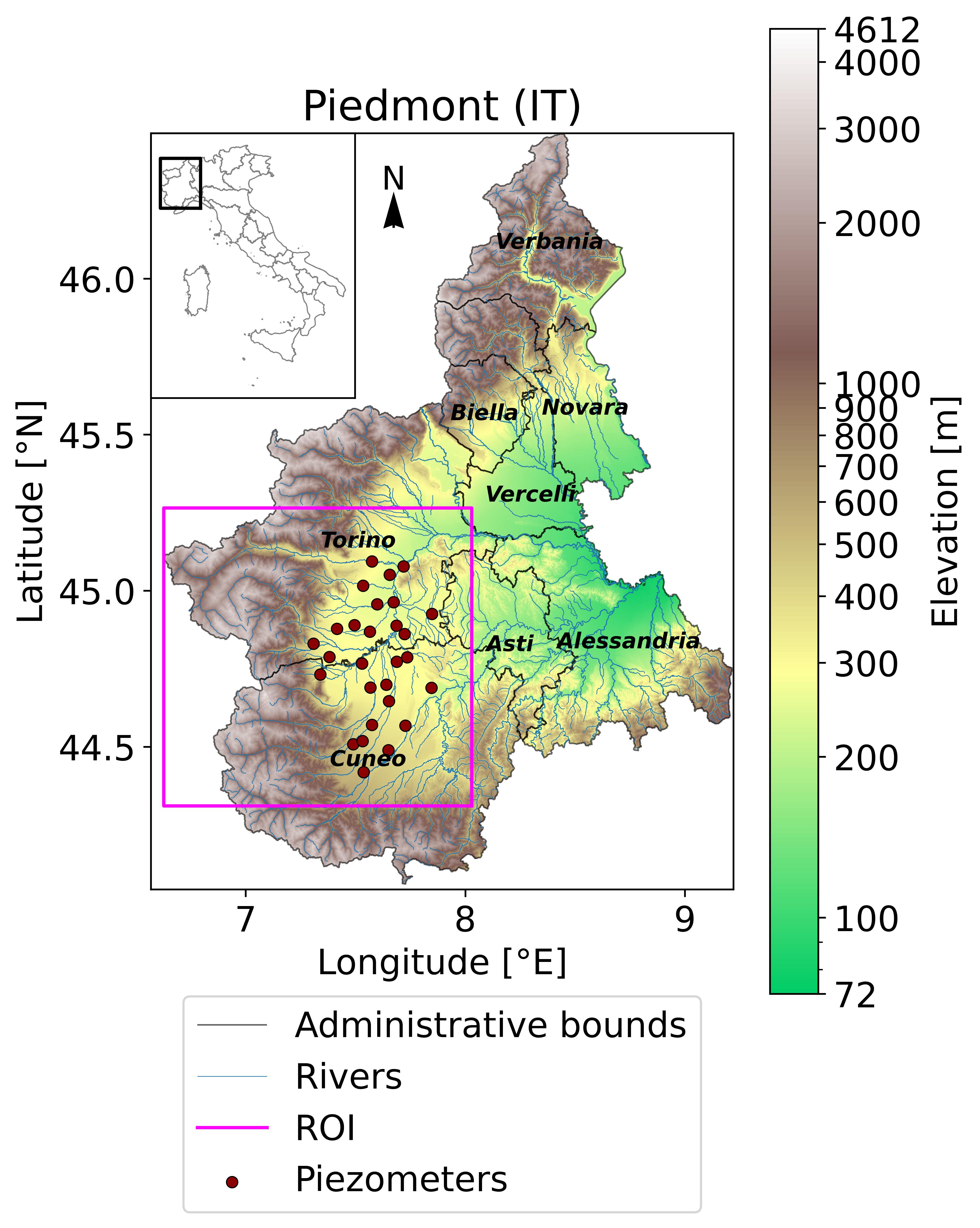}
    \end{subfigure}
    \hfill
    \begin{subfigure}[c]{0.49\textwidth}
        \centering
        \includegraphics[width=\textwidth]{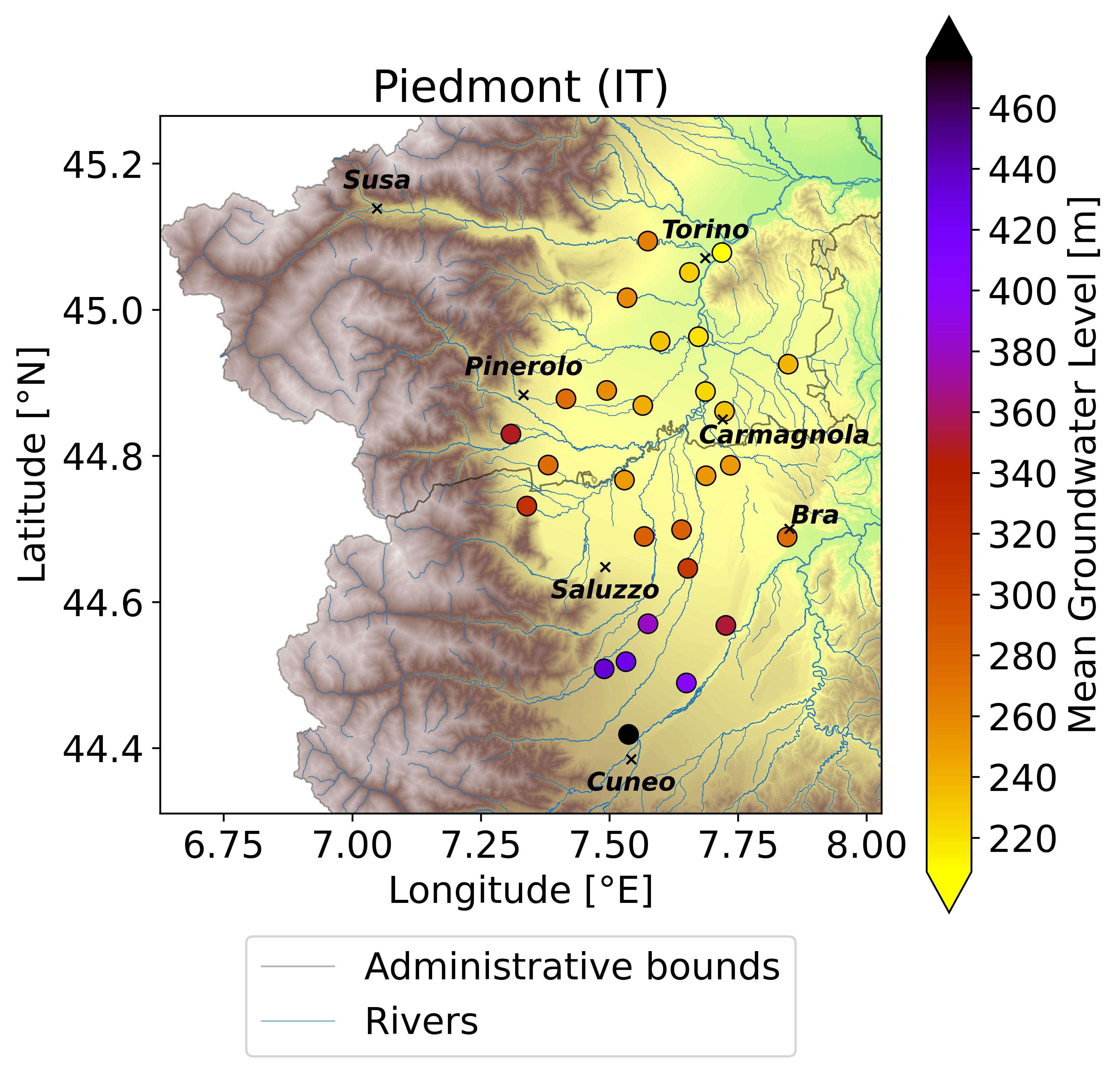}
    \end{subfigure}
    \caption{On the left, the Piedmont region with the ROI box and the selected piezometers with a colormap representing the elevation; bold names represent Piedmont's Provinces. On the right, a zoom over the ROI with an additional colormap representing the mean groundwater levels; bold names represent cities.}
    \label{fig:piedmont_roi}
\end{figure*}
\begin{figure}[h]
    \begin{subfigure}{\linewidth}
        \centering
        \includegraphics[width=0.8\textwidth]{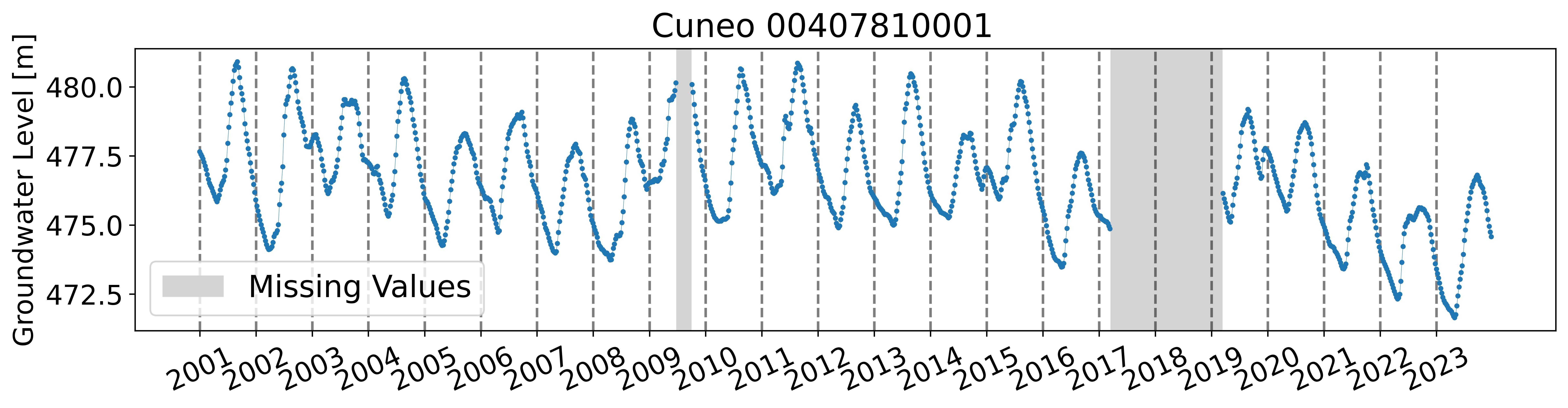}
        \caption{}
        \label{fig:cuneo_ts}
    \end{subfigure}
    \begin{subfigure}{\linewidth}
        \centering
        \includegraphics[width=0.8\textwidth]{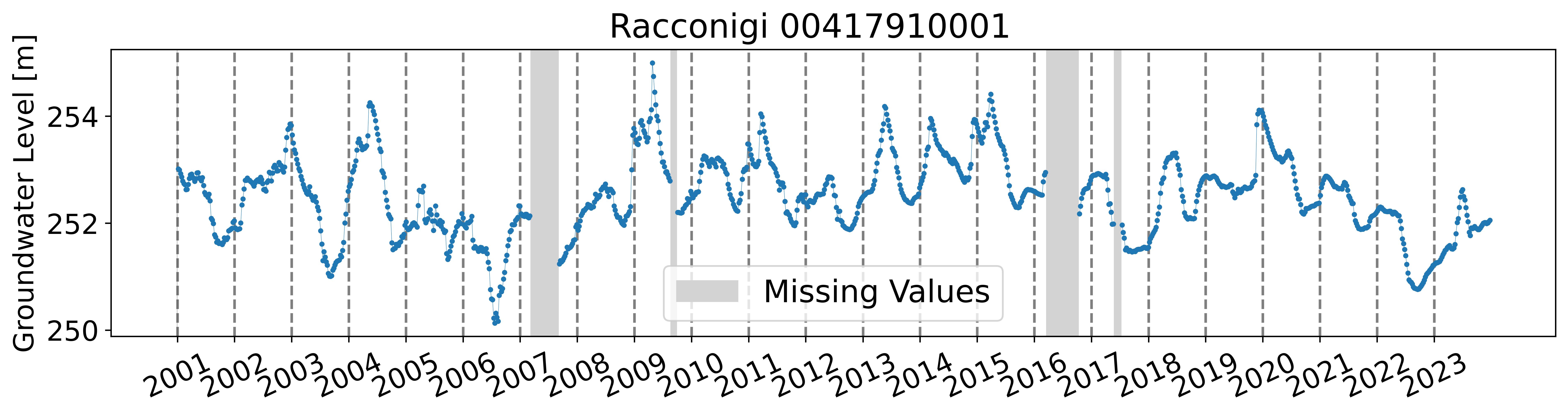}
        \caption{}
        \label{fig:racc_ts}
    \end{subfigure}
    \begin{subfigure}{\linewidth}
        \centering
        \includegraphics[width=0.8\textwidth]{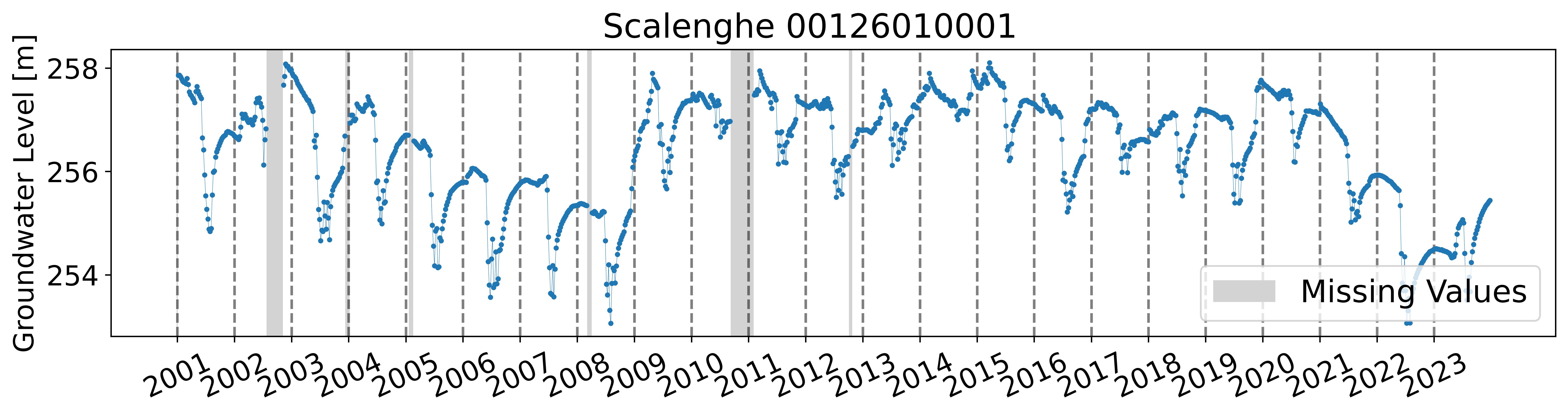}
        \caption{}
        \label{fig:sca_ts}
    \end{subfigure}
    \begin{subfigure}{\linewidth}
        \centering
        \includegraphics[width=0.8\textwidth]{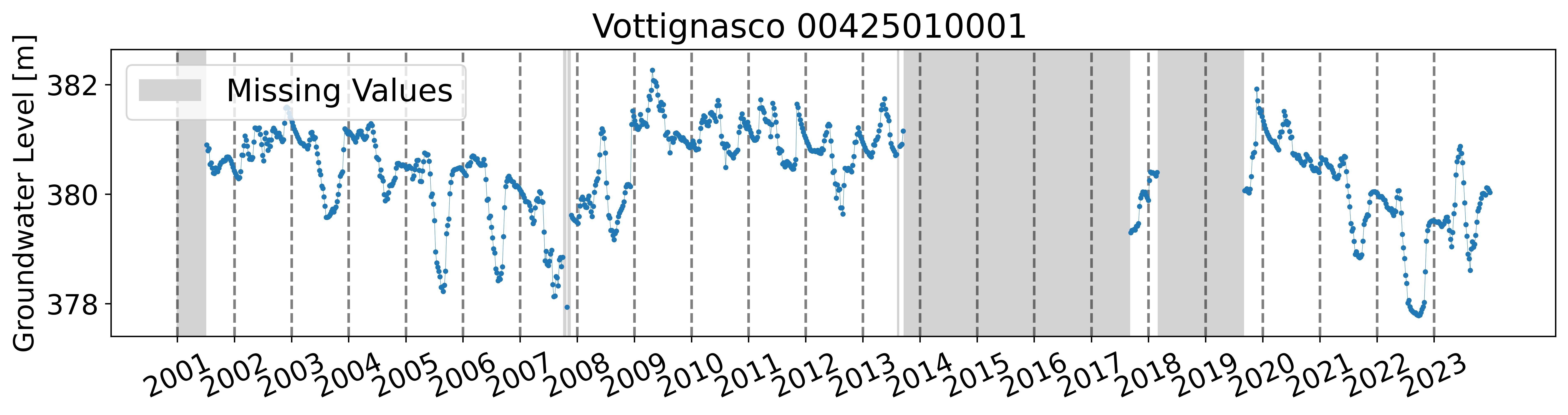}
        \caption{}
        \label{fig:vott_ts}
    \end{subfigure}
     \caption{Time series of the sensors in \textbf{a)} Cuneo, \textbf{b)} Racconigi, \textbf{c)} Scalenghe, and \textbf{d)} Vottignasco.}
     \label{fig:focus_ts}
\end{figure}

In this study, we \rev{focused on an area between the provinces of Cuneo and Torino in Piedmont, Italy, characterized by uniformly permeable soils that do not limit rain and snowmelt infiltration.
The area also exhibits intensive agricultural activity and a high degree of urbanization, making water resource assessment and sustainable water management particularly relevant.
In more detail, we defined} our ROI as a square region with longitudes ranging from $6.6267$° to $8.0292$° and latitudes from $44.3108$° to $45.2650$°, of approximately 16700 km$^2$ (see Figure~\ref{fig:piedmont_roi}).
\rev{This square region was defined to encompass the hydrological catchments monitored by the piezometers while producing square meteorological images that can be readily processed by deep learning models.}
Concerning groundwater data, in this area, we retrieved 28 piezometers' weekly time series, measuring the groundwater level from 2001-01-01 to 2023-12-31. These sensors are part of the Regional Environmental Agency of Piedmont (Agenzia Regionale per la Protezione Ambientale, ARPA Piemonte in Italian) network\footnote{Free access is provided at \url{https://shorturl.at/F3KzQ}.}.
The time series of sensors \rev{localized in the municipalities of} Cuneo, Racconigi, Scalenghe, and Vottignasco \rev{in Italy} are reported in Figure~\ref{fig:focus_ts}; all the other time series, along with their means, standard deviations, and percentages of missing values, are shown in Section~\ref{apx:sensor_stat_ts} in Figure~\ref{fig:all_ts} and Table~\ref{tab:all_ts_stats}.
The right side of Figure~\ref{fig:piedmont_roi} represents the location of all the sensors with a colour scale indicating their mean over time.

A visible aspect of these time series is that they show very different dynamics and missing periods. This is a complex issue to handle because it suggests the presence of different processes involved in the groundwater level data.
In particular, some exogenous phenomena, like abstraction for irrigation or the response to precipitation, might act differently in different locations.\\
\noindent Given the high percentage of missing values among all sensors (see Table~\ref{tab:all_ts_stats} and Figure~\ref{fig:miss_perc} in Section~\ref{apx:sensor_stat_ts} of the Supplementary Materials) with an \rev{median} value of around \rev{22}\%, we decided not to discard any sensors but simply fill the missing data with the last available data for every sensor, i.e. a \rev{forward filling strategy}. 
However, these filled data were then masked out, and so not considered, in the loss computation during training.

We adopted \rev{the fifth ECMWF ReAnalysis land dataset (ERA5-land)}~\cite{era5land} as the source data for weather information. ERA5-land is a reanalysis dataset providing different environmental variables at a spatial resolution of 0.1° and an hourly temporal resolution.
In particular, we downloaded total precipitation, 2 m height temperature, potential evaporation, and snowmelt, and performed a weekly average aggregation.
\rev{It is worth noting that such a data structure made of a sequence of raster images can be considered as a video.
Indeed, a video is made of frames (in our case, the rasters) evolving over time.}

\rev{
We initially used data from 2001-01-01 to 2019-12-31 for training and from 2020-01-01 to 2021-12-31 for validation and hyperparameter tuning.
Once the optimal configuration was identified, we retrained the models using both the training and validation sets, and we evaluated their performance on the test set covering the period from 2022-01-01 to 2023-12-31.
Figure \ref{fig:num_obs} of the Supplementary Material reports the number of observations for each sensor in all the dataset splits.}
Furthermore, we normalized all data by computing z-scores $z = \frac{F - \mu_F }{\sigma_F}$ for all features $F$, where all means $\mu_F$ and standard deviations $\sigma_F$ were computed on the training set.

\section{Methods}

\begin{figure*}[!h]
    \centering
    \includegraphics[width=0.8\linewidth]{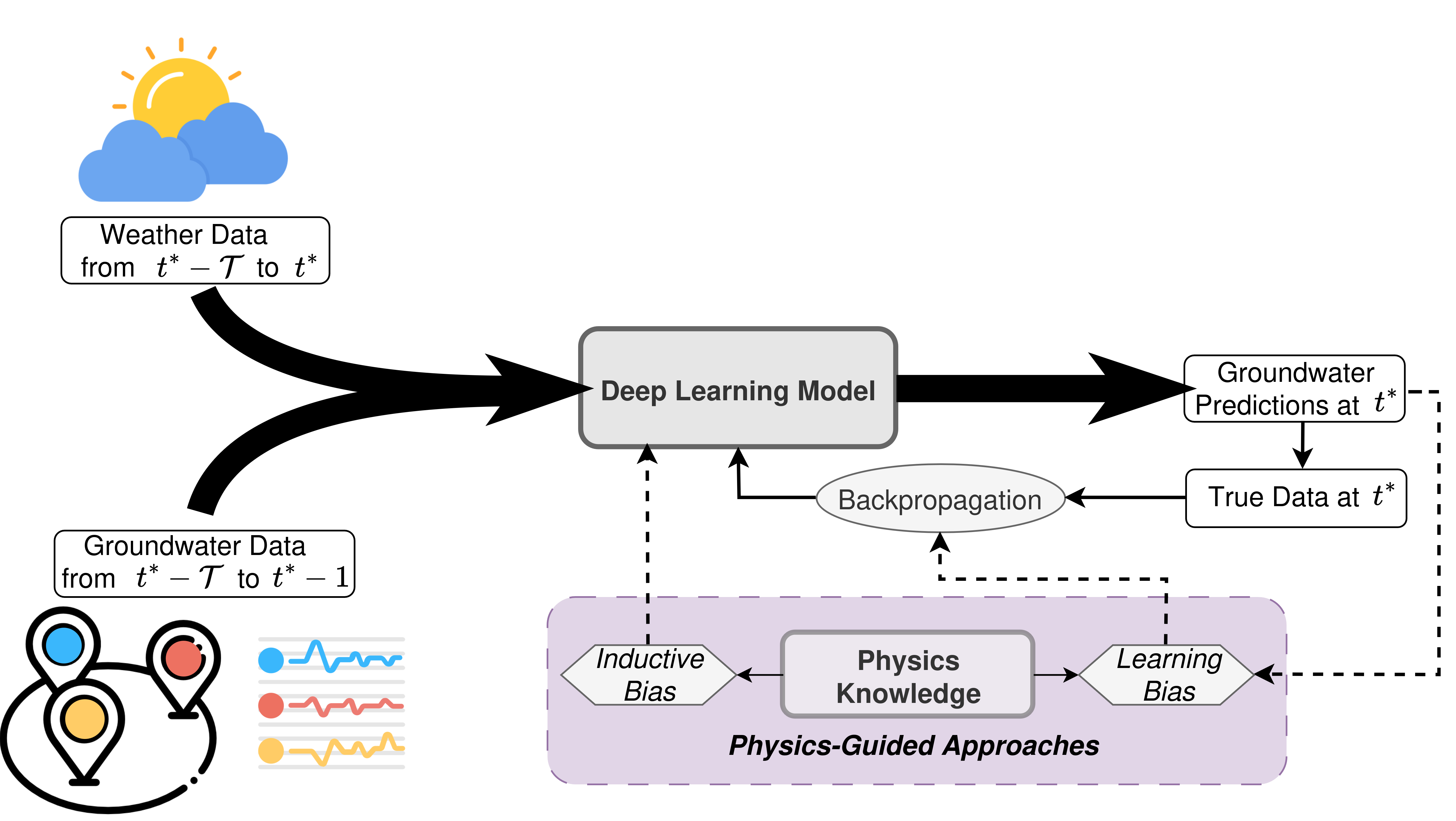}
    \caption{The adopted modeling pipeline. Each of our models takes as input the weather and groundwater data from a window of length $\mathcal{T} + 1$ and $\mathcal{T}$, respectively, from the target time $t^*$.
    The predictions are compared with true measurement data, and through backpropagation, the weights of the model are updated to lower such differences. 
    What differs between models is whether or not the physics knowledge is leveraged (dashed lines). This could be done by applying an inductive bias or a learning bias that acts on the model architecture and on the learning algorithm (backpropagation), respectively.}
    \label{fig:pipeline}
\end{figure*}

\rev{Our proposed models follow the pipeline illustrated in Figure~\ref{fig:pipeline}.
In particular, for a target time $t^*$, the deep learning models take as input the weather data, in the form of a sequence of raster images covering the whole ROI from $t^* - \mathcal{T}$ to $t^*$, and the past groundwater level measurements from $t^* - \mathcal{T}$ to $t^* -1$, with $ \mathcal{T} = 4$. 
The choice of using 4 weeks as an input lag window was to provide the model with monthly information on both weather and groundwater levels, which, from the domain knowledge, is a reasonable amount of time to infer major patterns.
This choice was further validated during the fine-tuning performed on the validation set.}

\rev{The models then predict the weekly groundwater level for the week $t^*$ for all the target points, which could be the sensors' locations or arbitrary points over the whole ROI. 
It is worth noting that the supervision of the groundwater levels against true data was applied only to the sensor location points for which we had true measurement data. 
Meanwhile, we generated a regular grid of predictions to apply the learning biases and to perform a qualitative evaluation of the estimated equation components.
}

\rev{We developed 4 different models, the first with a pure deep learning approach trained just using a data loss $\mathcal{L}_{\textit{data}}$ supervising the groundwater level predictions against measurements (and so without the Physics Knowledge block in the pipeline in Figure \ref{fig:pipeline}).
Then we restructured the tail of the architecture (applying an inductive bias), inspired by the groundwater flow equation, making the output a summation of three different terms, namely the autoregressive, diffusion, and residual components. 
Subsequently, we applied learning biases to the restructured architecture, supervising these components. In particular, we added a loss term related to the residuals of the groundwater flow equation evaluated on the model output.
Finally, for the last model, we leveraged the domain knowledge of the groundwater body recharge by applying a loss term that penalizes recharge of groundwater outside the estimated recharge zones. 
}

\rev{
\subsection{Deep Learning Overview}
In the following, we describe the building blocks of our neural network architectures, for which we have been inspired by~\cite{arnab_2021,andrychowicz_2023,allen_2025}.} 
\subsubsection{Multi-Head Attention Mechanism}
\rev{Given the overwhelming performance obtained from architectures based on Multi-Head Attention (MHA)~\cite{VaswaniAttention2017}, we decided to adopt MHA as the backbone of our models.
Furthermore, since attention generates predictions indexed at query points, taking as input values indexed at key points, it enabled us to} develop learning models that can generate \rev{predictions also at points different from the piezometers' locations.}

The MHA is based on the scaled dot-product attention that computes attention weights by taking the dot product between query ($Q$) and key ($K$) vectors, scaled by the inverse square root of the key dimension ($d_k$) to maintain numerical stability.
These weights are normalized using a softmax function and applied to the corresponding value vectors ($V$) to produce a weighted sum representing the attention output (Equation~\ref{eq:dpatt}).
When the queries, keys, and values are all derived from the same input sequence, the mechanism is referred to as self-attention.
This mechanism enables efficient computation of dependencies within a sequence by emphasizing relevant contextual information while mitigating gradient instability in high-dimensional spaces.

\begin{equation}
    \text{Attention}(Q, K, V) = \text{softmax}\left( \frac{QK^{T}}{\sqrt{d_k}} \right) V
    \label{eq:dpatt}
\end{equation}

The Multi-Head formulation consists of performing $n$ parallel and independent scaled dot-product attention (heads) on different linear transformations of the \rev{query, value, and key}. All heads' outputs are then concatenated, and a linear transformation is performed to produce the final output, as in Equation~\eqref{eq:multihead}:

\begin{equation}
\begin{split}
    \text{Multi-Head Attention}(Q, K, V) = \text{Concat}(\text{head}_1,...,\text{head}_n)W^O
\end{split}
    \label{eq:multihead}
\end{equation}

Where the generic $\text{head}_i$ is $\text{Attention}(Q_i,K_i,V_i)$.

This structure enables the model to extract different but complementary meaningful information from the input sequence, thereby learning more complex relationships.

The MHA layer is usually adopted in blocks, followed by feed-forward layers, as in the Transformer architecture~\cite{VaswaniAttention2017,arnab_2021}.
For our purpose, and inspired by \cite{arnab_2021}, we defined an MHA block as depicted in Figure~\ref{fig:mha_block} that consists of an MHA layer followed by an MLP made of two fully connected layers.
We used Instance Normalization~\cite{ulyanov_2017} as the normalization procedure (Norm layer in Figure~\ref{fig:mha_block}), and LeakyReLU as the activation function after each layer.
Two residual connections are then employed to ease the gradient flow.

\subsubsection{Values, Keys, and Queries Embedding}

\rev{Within the attention framework, the input and output are objects made of tokens, which could be the words in a sentence, the patches in an image, or the values in a time series~\cite{VaswaniAttention2017,autoformer2021,arnab_2021}.
Every token is associated with the above-mentioned query, key, and value vectors. 
The query represents the information that a token seeks from the rest of the sequence, or in other sequences, the key vector describes the type of information that each token provides, and the value vector contains the information that can be transferred to other tokens. 
Every output token is then the aggregation of the input values, weighted by the similarity of the query to all the input keys.
This allows the attention mechanism to predict a number of tokens different from the number of input tokens; in other words, the number of queries can be different from the number of keys and the corresponding values.
The query, key, and value are typically obtained from the input tokens through embedding layers, which are usually implemented as linear layers‌~\cite{VaswaniAttention2017}.}

Conceptually, in our framework, the keys and queries are related to the spatio-temporal coordinates of the input data and the prediction locations, respectively. 
\rev{Consequently, adopting the attention mechanism, we can generate predictions at a variable number of locations (queries), even at ungauged ones, i.e. when the target points are different from the locations (keys) of the input lagged groundwater data (values). 
Therefore, predictions at ungauged locations are generated by leveraging the information provided by the available piezometer -- a sort of implicit interpolation of the groundwater level, performed on the latent representation of the network.}

\rev{To create the most informative spatio-temporal coordinates} embeddings (ST-CEmb), we augmented the coordinates (longitude and latitude) with the elevation, \rev{and the sine and cosine of the day of the year (doy)} -- the functions $sin(\frac{2\pi\ \text{doy}}{366})$ and $cos(\frac{2\pi\ \text{doy}}{366})$ respectively. 
The usage of sine and cosine signals for the embedding creation has been adopted in the original Transformer architecture~\cite{VaswaniAttention2017} to inform about the position of each token in the sequence, but also in environmental applications to ease the seasonality extraction~\cite{WunschKarst2022,allen_2025}.
The augmented spatio-temporal coordinates of each point, either in input or to be predicted, were fed to a linear layer that gives as output the spatio-temporal coordinates embeddings -- the keys, or the queries, respectively.

Concerning the computation of the value embeddings (ST-VEmb), for the autoregressive component, we adopted a linear layer that takes as input the groundwater level value concatenated with the augmented spatio-temporal coordinates described before. 
Differently, for the value embedding of the weather data, given the image-like structure, we \rev{concatenated the weather features with the augmented spatio-temporal coordinates and we fed them into} a convolutional layer with a 3×3 kernel, followed by an average pooling layer.

\subsection{Formalization}
Our target variable is the hydraulic head $h(x,y,t)$ indexed by time $t$, and spatial coordinates $x,y$. Given that we considered the unconfined aquifer, $h$ corresponds to \rev{the groundwater level.
The piezometers measure $h$ at $M = 28$ points inside the ROI, labeled as $\Omega$}, and we can thus define the spatial measuring domain $\mathcal{M}$ as the set of sensors' locations $\mathcal{M} = \{(x_{m},y_{m})|m \in [1;M]\}$ and $\mathcal{M} \subset \Omega$.

\noindent We aimed to develop a model that can predict an arbitrary number $P$ of points $\mathcal{P}$ inside the ROI at time $t^*$, more precisely $\mathcal{P} = \{(x_{p},y_{p})|p \in [1;P]\}$.
Our target can thus be defined as $h_{\mathcal{P},t^*} = \{h(x,y,t)|(x,y) \in \mathcal{P}, t = t^* \}$, note that $\mathcal{P} \subset \Omega$.

\noindent As input data, we leveraged an autoregressive component $H_{\mathcal{P},t^*-1}$ made of $\mathcal{T}$ lagged values of the target, i.e., $H_{\mathcal{P},t^*-1} = \{h_{\mathcal{\mathcal{P}},t}|t \in [t^* - \mathcal{T};t^*-1]\}$. 
Furthermore, we used \rev{a sequence of weather images over the ROI (a video) from the time of the former $\mathcal{T}$ lag up to the time $t^*$ of the prediction.
We defined the weather video} as $\mathbf{V}_{t^*} = \{v(x,y,c,t) \in \mathbb{R} | (x,y) \in \mathcal{H} \times \mathcal{W}, c \in [1;\mathcal{C}], t \in [t^* - \mathcal{T};t^*]\}$, where $\mathcal{H}$,$\mathcal{W}$ are the horizontal and vertical centroids of each pixel of the frame of the video, and $\mathcal{C}$ are the channels, i.e. total precipitation, 2 m height temperature, potential evaporation, and snowmelt.

\noindent \rev{We aimed to develop models that takes as input $H_{\mathcal{P}, t^*-1}$ and $\mathbf{V}_{t^*}$, and generate a prediction for $h_{\mathcal{P},t^*}$.}

\subsection{Deep Learning Architecture}

\begin{figure*}[ht]
    \centering
    \begin{subfigure}[c]{0.30\textwidth}
        \centering
        \includegraphics[width=\textwidth]{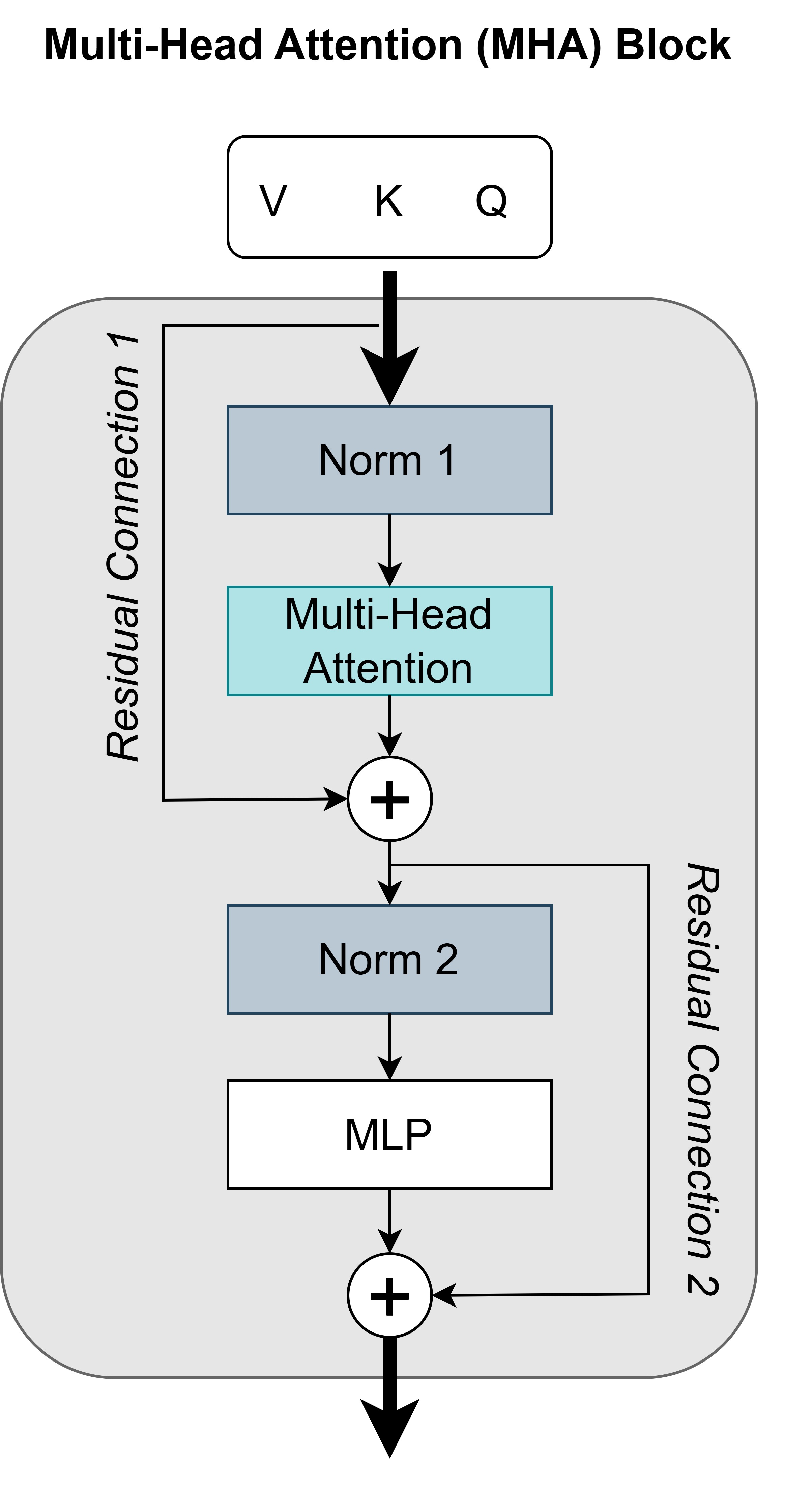}
        \caption{}
        \label{fig:mha_block}
    \end{subfigure}
    \begin{subfigure}[c]{0.30\textwidth}
        \centering
        \includegraphics[width=\textwidth]{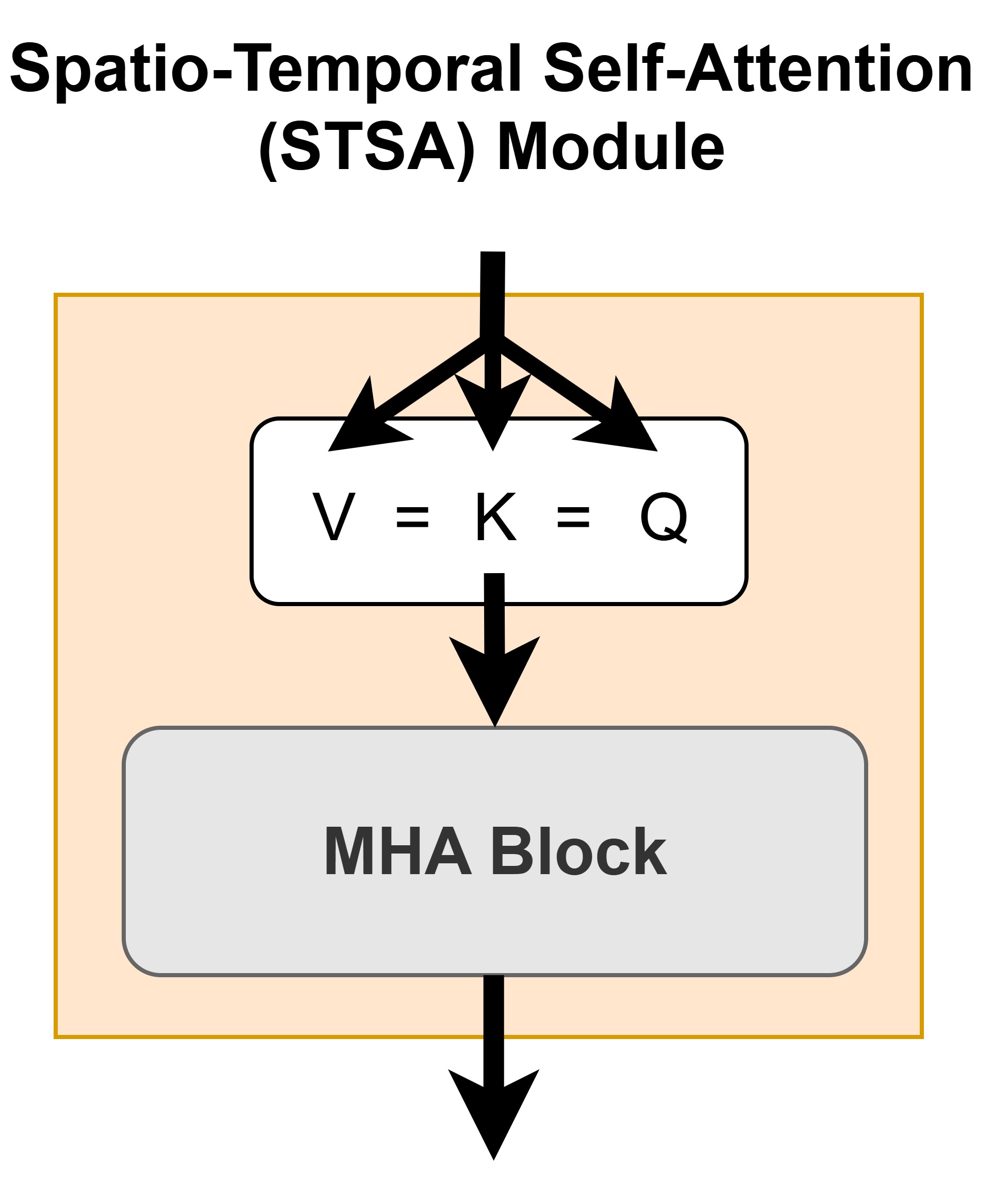}
        \caption{}
        \label{fig:arch_stsa}
    \end{subfigure}
    \begin{subfigure}[c]{0.30\textwidth}
        \centering
        \includegraphics[width=\textwidth]{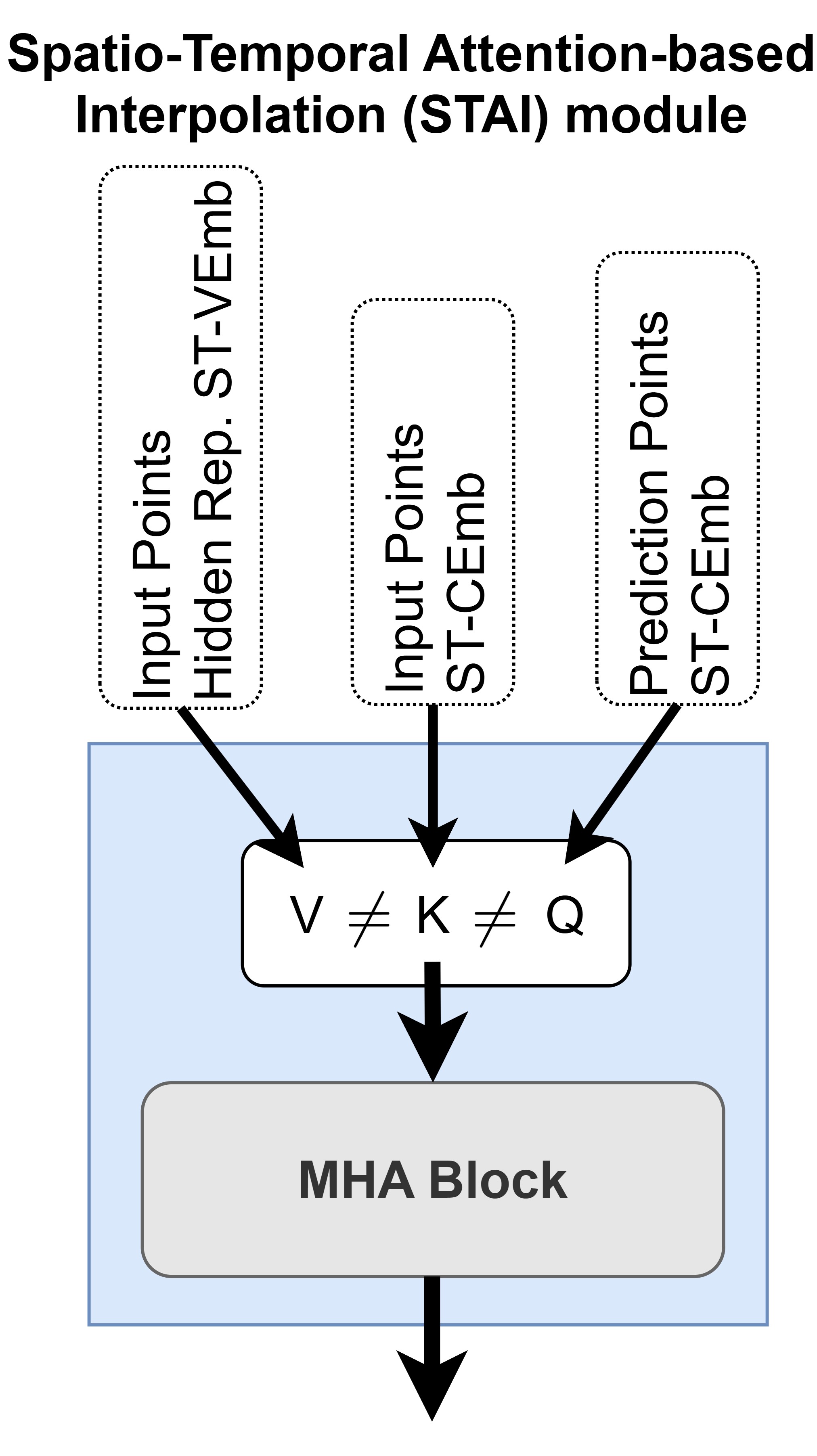}
        \caption{}
        \label{fig:arch_stai}
    \end{subfigure}
    \caption{\textbf{a)} Multi-Head Attention block adopted in the models. \textbf{b)} STSA Module \textbf{c)} STAI Module that takes as values the hidden representation of the input points' value embeddings (ST-VEmb), as keys the input points' spatio-temporal coordinates embeddings (ST-CEmb), and as queries the prediction points' spatio-temporal coordinates embeddings (ST-CEmb).}
    \label{fig:arch_stnet_mhablock}
\end{figure*}

Upon the MHA block, we defined \textbf{a)} the Spatio-Temporal Self-Attention (STSA, in Figure~\ref{fig:arch_stsa}) module made of one MHA block adopting self-attention; and \textbf{b)} the Spatio-Temporal Attention-based Interpolation (STAI, in Figure~\ref{fig:arch_stai}) module composed of one MHA block in which the keys are the spatio-temporal coordinates embeddings of the input points, the values are the hidden representation of the value embeddings, and the queries are the spatio-temporal coordinates embeddings of the prediction points -- given that the keys and queries could be of different lengths, we dropped the first residual connection in the MHA block.

\rev{Thanks to the STAI module, we can predict at a variable number of locations, even at ungauged ones.
Indeed, leveraging attention, it projects the input groundwater data (values and keys) into the prediction points (queries), which is a form of automatic interpolation performed in the latent space by the neural network.
However, such a possibility is used just to apply the physics learning bias and not to perform a systematic evaluation of the predictions at ungauged locations.}

To further condition the model on the spatio-temporal location of the prediction points, we adopted the Feature-wise Linear Modulation (FiLM) conditioning~\cite{perez_2018}. The FiLM layer performs a conditional feature normalization by scaling and shifting every feature channel.
In particular, the scale and shift factors are computed by a linear layer from the conditioning information -- in our case, the augmented spatio-temporal coordinates of the prediction points.
This mechanism enables adaptation of neural representations to the conditioning information, facilitating the learning of complex context-dependent relations and improving the generalization ability of the model.

\begin{figure*}[!h]
    \centering
    \includegraphics[width=0.95\linewidth]{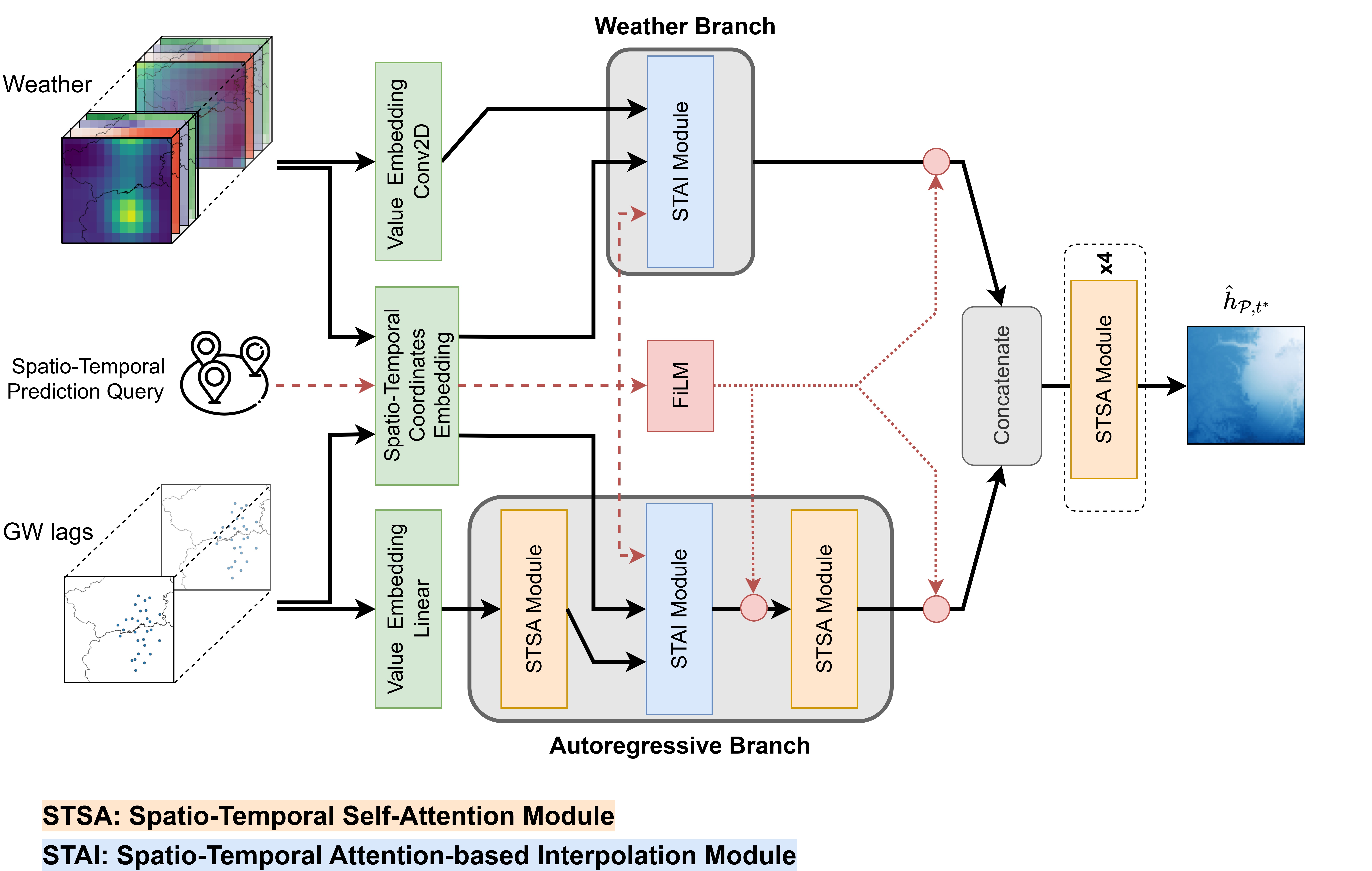}
    \caption{STAINet architecture.}
    \label{fig:arch_stainet}
\end{figure*}

We named our pure deep learning model Spatio-Temporal Attention-based Interpolation neural Network (STAINet), and its architecture is depicted in Figure~\ref{fig:arch_stainet}.
We decided to process the autoregressive component and the weather component with two parallel branches.
In particular, we designed an Autoregressive Branch that begins with one STSA module.
During training, the MHA layer adopted in this first module employs dropout to randomly omit part of the input key–value pairs, thereby enhancing the model’s generalization capability and preventing it from focusing on specific input locations~\cite{andrychowicz_2023}.
Subsequently, to extract the relevant information \rev{from the input groundwater locations}, an STAI module projects the hidden representation of the spatio-temporal value embedding (i.e., the output of the previous STSA module) onto the spatio-temporal prediction query, followed by an additional STSA module.
The FiLM conditioning is adopted both after the STAI module and after the last STSA module to force the hidden representations to capture context-dependent relationships and facilitate the model in differentiating among prediction points.

\noindent This configuration was found to be the optimal choice for the Autoregressive Branch, balancing the model's complexity with the ability to learn complex dependencies from in situ spatially sparse measurements.

To handle the exogenous weather information, we defined the Weather Branch, which is made up of only one STAI module, whose outputs are conditioned by the FiLM layer.
The inclusion of STSA modules in this branch did not prove effective; on the contrary, it made the training process more difficult and reduced the models’ generalization capability. This behaviour is likely attributable to the use of the time-distributed convolutional embedding layer for constructing the ST-VEmb, which already encapsulates the relevant spatial information.

For the pure deep learning model, the two branches were concatenated, and the resulting hidden representation was passed through four stacked STSA modules.
The output is then processed by a last linear layer that produces the predictions for all the desired points.

Due to differences in the time series across sensors, we adopted the Mean Absolute Percentage Error (MAPE) as the data loss term ($\mathcal{L}_{data}$) since it proved more effective than other non-relative loss functions. 
Furthermore, we introduced an orthogonality regularization term $\mathcal{L}_{ortho}$ to enforce orthogonality on the weights of the embedding layers.
This constraint encouraged the embeddings to capture non-redundant features~\cite{vorontsov_2017,bansal_2018}.

\newpage

\subsection{Physics-Guided Approach}

Considering the PDE in Equation~\eqref{eq:gw_pde_new}, applying the Euler integration method, we then expressed $h$ at a time $t^*$ ($h_{t^*}$) as a function of its $\Delta t$ lag ($h_{t^*-\Delta t}$), as in Equation~\eqref{eq:gw_pde_new_euler}.


\begin{equation}
    h_{t^*} = h_{t^*-\Delta t} + \Delta t \left[ \frac{T}{S} \left(\frac{\partial^2 h}{\partial x^2} + \frac{\partial^2 h}{\partial y^2}\right)\right]
\label{eq:gw_pde_new_euler}
\end{equation}

To simplify the notation, we defined the diffusion parameter $\mathcal{D} = \frac{T}{S}$ expressed in [L\textsuperscript{2}T\textsuperscript{-1}], and $\Delta_{GW_{t^*}} = \Delta t\left[\mathcal{D}\left(\frac{\partial^2 h}{\partial x^2} + \frac{\partial^2 h}{\partial y^2}\right)\right]$ expressed in [L]. We further added a residual (sink/source) term $\mathcal{R}_{t^*}$ expressed in [L] related to all exogenous factors occurring in the interval $\Delta t$, which can be either anthropogenic (e.g., water abstraction for irrigation) or natural (e.g., rainfall or snowmelt recharge).
We thus obtained Equation~\ref{eq:gw_pde_lag}:

\begin{equation}
h_{t^*} = h_{t^*-1} + \Delta_{GW_{t^*}} + \mathcal{R}_{t^*}
    \label{eq:gw_pde_lag}
\end{equation}

Hereafter, we refer to $h_{t^*-1}$ as the autoregressive component and to $\Delta_{GW_{t^*}}$ as the diffusion displacement component. 
In our setting, we considered $\Delta t$ equal to one week.

\subsubsection{Inductive Bias Strategy}

\begin{figure*}[h]
    \centering
    \includegraphics[width=0.95\linewidth]{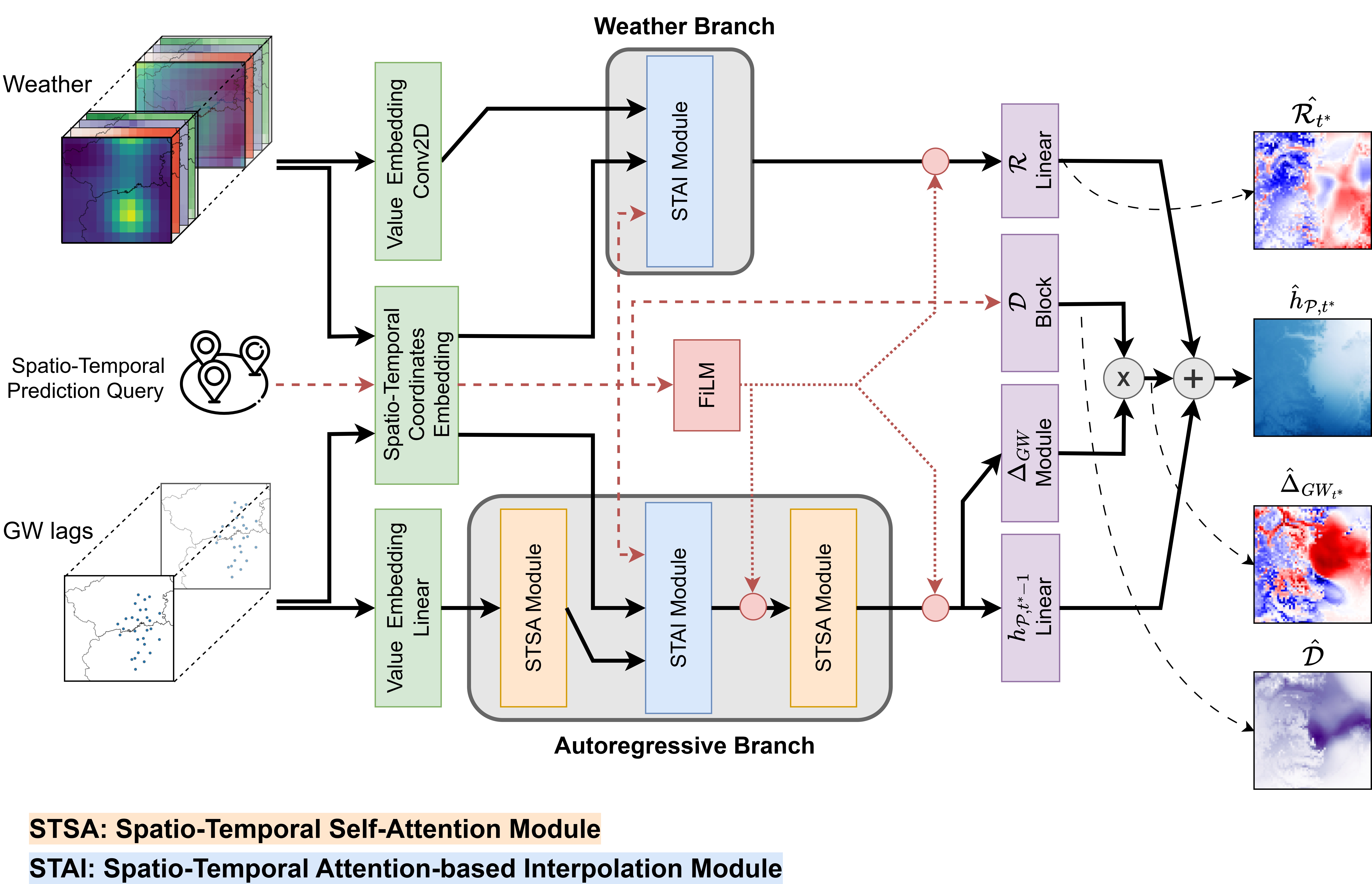}
    \caption{PSTAINet-IB architecture.}
    \label{fig:arch_pstainet}
\end{figure*}

As an initial approach to incorporate physics knowledge into the deep learning model, we modified STAINet to explicitly output the three components defined earlier, computing the final output as the sum of these components.
To this end, we developed the PSTAINet-IB (Physics-guided STAINet with Inductive Bias), which is depicted in Figure~\ref{fig:arch_pstainet}.

The initial part of the architecture follows the same structure as the STAINet; however, PSTAINet-IB does not concatenate the two branches.
In particular, the $\mathcal{R}_{t^*}$ and $h_{\mathcal{P},t^*-1}$ terms are estimated by two linear layers.
Differently, since the diffusion component $\Delta_{GW_{t^*}}$ is more complex, we employed a more advanced neural structure for its estimation.

\begin{figure*}[h]
    \centering
    \begin{subfigure}[c]{0.25\textwidth}
        \centering
        \includegraphics[width=\textwidth]{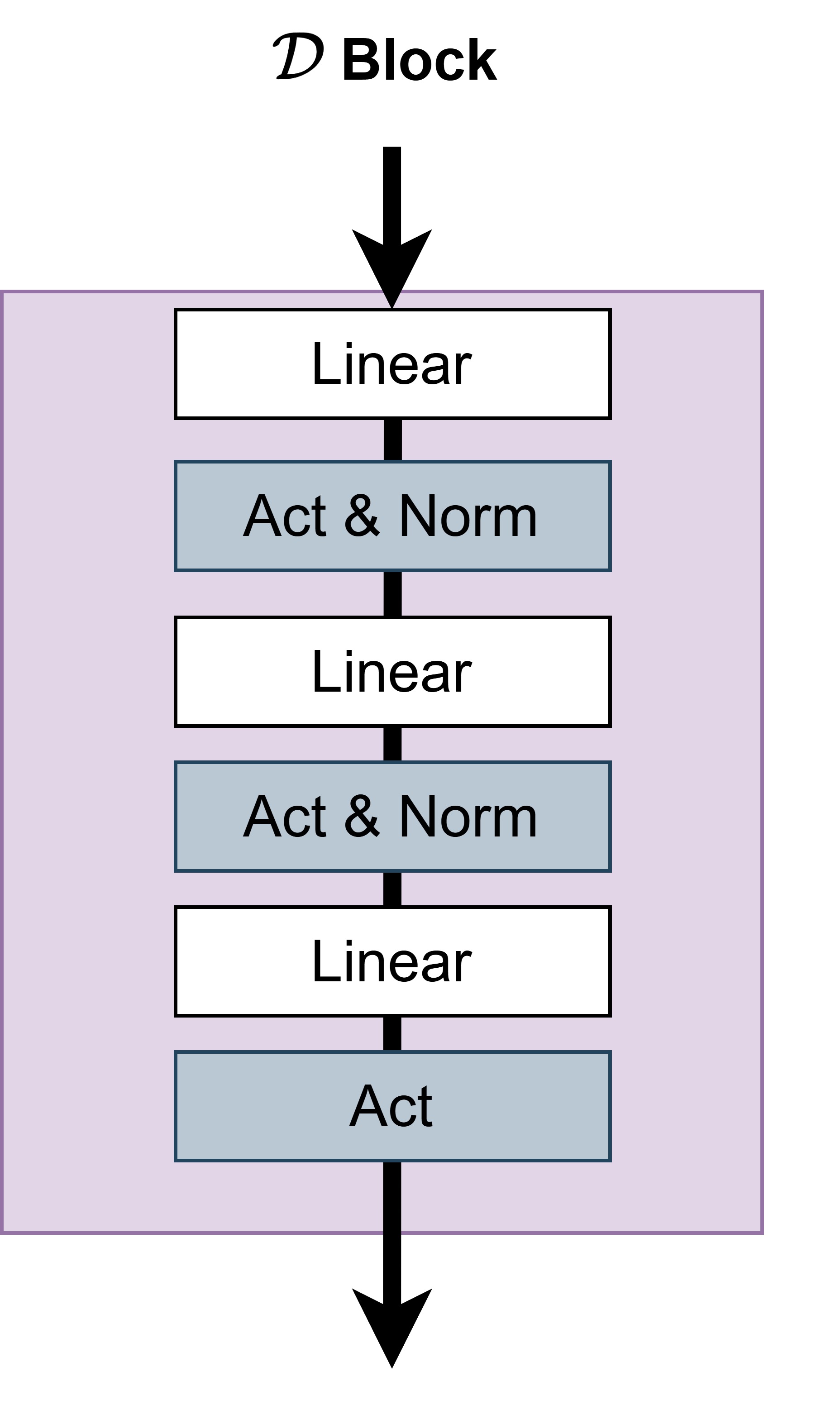}
        \caption{}
        \label{fig:arch_dblock}
    \end{subfigure}
    \begin{subfigure}[c]{0.25\textwidth}
        \centering
        \includegraphics[width=\textwidth]{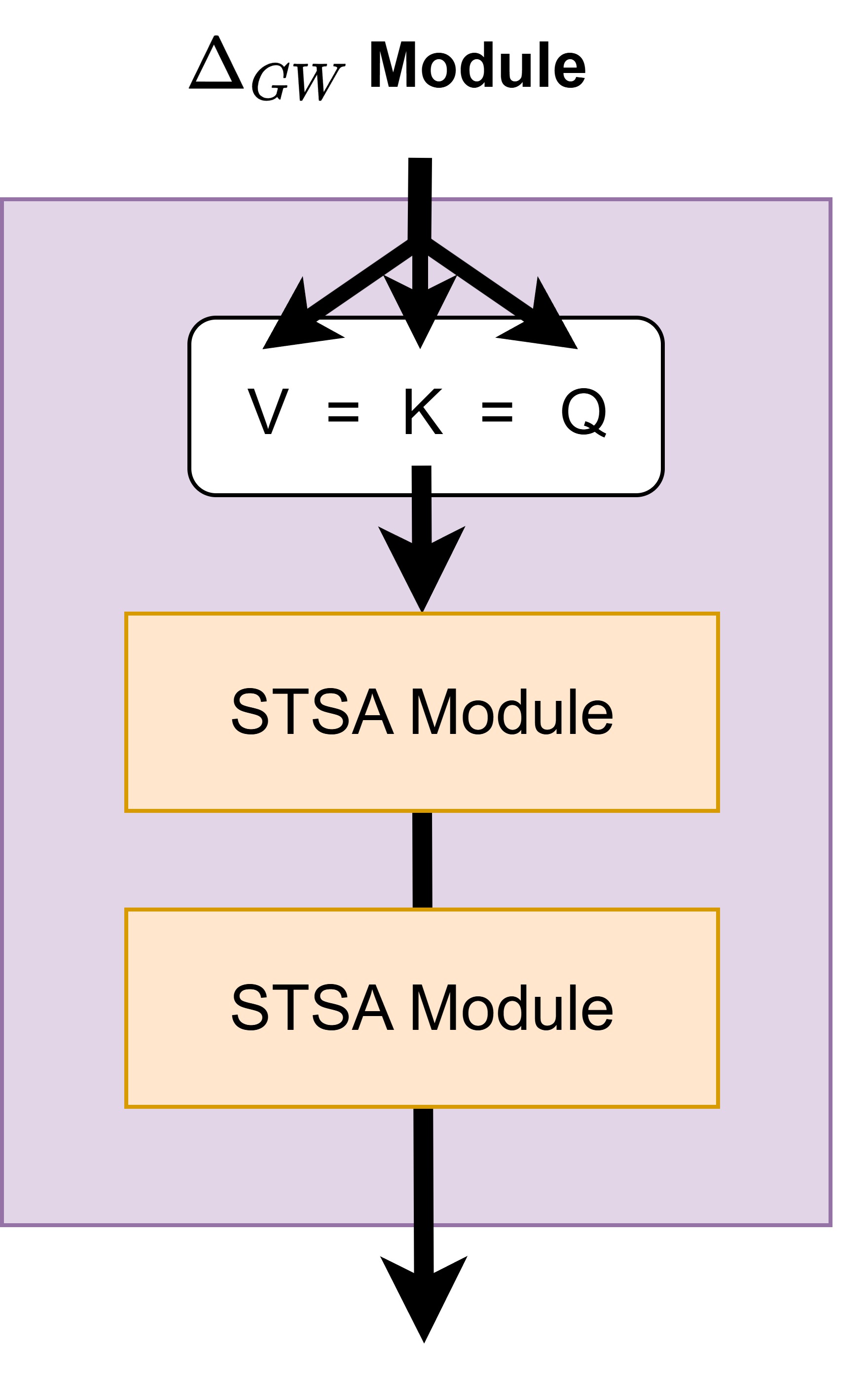}
        \caption{}
        \label{fig:arch_dgw_block}
    \end{subfigure}
     \hfill
    \caption{\textbf{a)} $\mathcal{D}$ Module in which Act \& Norm stand for activation and normalization respectively; \textbf{b)} $\Delta_{GW}$ Module.}
    \label{fig:arch_ib_blocks}
\end{figure*}

Firstly, the $\mathcal{D}$ Block estimates the parameter $\mathcal{D}$, taking as input the spatial coordinates and elevation of the prediction points. It is implemented as an MLP comprising three linear layers (Figure~\ref{fig:arch_dblock}), followed by Instance Normalization and LeakyReLU activation. 
We designed the $\mathcal{D}$ Block under the assumption that $\mathcal{D}$ is a time-invariant property of the soil, and thus can be estimated based on spatial coordinates. Although simple in structure, this block provides hydrologically consistent estimations of $\mathcal{D}$ while keeping the number of trainable parameters low.
We then developed the $\Delta_{GW}$ module (Figure~\ref{fig:arch_dgw_block}), consisting of a stack of two STSA modules, to estimate the divergence term $\frac{\partial^2 h}{\partial x^2} + \frac{\partial^2 h}{\partial y^2}$. 
In the $\Delta_{GW}$ module, we adopted STSA modules in place of MLPs to enable the estimation of more complex relationships among prediction points.
The diffusion component is then computed as the multiplication of the estimated $\hat{\mathcal{D}}$ and the divergence term ($\Delta t$ is omitted as we set it equal to one).

During the training of the model PSTAINet-IB, the three terms were not directly supervised; only their sum, corresponding to the final output of the model, was used in the calculation of the data loss function $\mathcal{L}_{data}$.

\vspace{1cm}

\vspace{2cm}

\subsubsection{Learning Bias Strategy}
\label{sec:lb_strategy}
We then tested a second strategy consisting of adding learning biases to the PSTAINet-IB.
Firstly, we supervised the estimated autoregressive component using the MAPE between $\hat{h}_{\mathcal{P},t^*-1}$ and the true data $h_{\mathcal{P},t^*-1}$, to make the lag estimation coherent with the available data, we named this loss term $\mathcal{L}_{coh}$.

Subsequently, we computed the Mean Squared Error (MSE) of the residuals between the estimated diffusion displacement and its discretized version computed on the estimated lag -- we named the loss built on these residuals $\mathcal{L}_{\text{diff}}$.
In detail, we applied the finite difference method with a central difference approximation~\cite{Tadmor_2012} to compute the second-order spatial derivatives of \rev{the gridded prediction $\hat{h}_{\mathcal{P},t^*-1}$.
We decided to generate a 70x70 grid of predictions that corresponds to a map with a spatial resolution of $1.5$ km.
It is worth noting that this loss is computed on all the gridded prediction points, named control points in the PINN literature, and does not require true data for supervision.
The choice of using 70x70 images was because of a trade-off between higher resolution and memory load; indeed, higher resolution maps mean more control points and thus more computations during training.
We found producing $1.5$ km maps to be a good compromise between memory load and resolution of the output maps on which we performed a qualitative analysis.}

\noindent We further introduced some regularization terms on the $\ell^1$ and $\ell^2$ norms of the $\hat{\mathcal{R}}$ to constrain its mean value toward zero and mitigate high-magnitude responses, termed $\mathcal{L}_{||{\mathcal{R}||}_1}$ and $\mathcal{L}_{||{\mathcal{R}||}_2}$, respectively.
We named the model with these learning biases PSTAINet-ILB (Physics-guided STAINet with Inductive and Learning Biases).

Finally, we leveraged the domain knowledge of recharge zones, which are the zones where the groundwater body recharges due to precipitation and snowmelt.  
In our ROI, ARPA experts defined the recharge zones for the groundwater body in an area at the base of the mountain\footnote{Available here: \url{https://shorturl.at/qTheu}.}.
We considered a buffer of 0.05° and defined a loss term $\mathcal{L}_{RCH}$ that penalizes positive values of $\mathcal{R}$ outside the recharge zones.
Formally, defining $\mathcal{R^+}$ as the positive value of $\mathcal{R}$, $RCH$ as the set of points inside the recharge zones, and its complement $RCH^c$, we defined $\mathcal{L}_{RCH} = \frac{1}{\sum\mathds{1}_{RCH^c}(\mathcal{R^+})}\sum\mathds{1}_{RCH^c}(\mathcal{R^+})\mathcal{R}$.
We named the model trained with this further regularization PSTAINet-ILRB (Physics-guided STAINet with Inductive, Learning, and Recharge zone Biases).

The final loss is then the summation of all terms previously described (if adopted by the considered model), as in Equation~\eqref{eq:tot_loss}:

\begin{equation}
\begin{split}
\mathcal{L} =\;& \mathcal{L}_{\text{\textit{data}}} 
+ \alpha_{\text{\textit{ortho}}}\mathcal{L}_{\text{\textit{ortho}}}
+ \alpha_{\text{\textit{coh}}}\mathcal{L}_{\text{\textit{coh}}} 
+ \alpha_{\text{\textit{diff}}}\mathcal{L}_{\text{\textit{diff}}} \\
&+ \alpha_{||\mathcal{R}||_1}\mathcal{L}_{||\mathcal{R}||_1}
+ \alpha_{||\mathcal{R}||_2}\mathcal{L}_{||\mathcal{R}||_2}
+ \alpha_{RCH}\mathcal{L}_{RCH}
\end{split}
\label{eq:tot_loss}
\end{equation}

\rev{
\subsection{Evaluation}
In the following, we explain the methods adopted to evaluate the predictions of the proposed models.}
\subsubsection{Metrics}
\rev{As a principal evaluation method, we compared some fitness metrics on the test set comprising groundwater measurements from 2022-01-01 to 2023-12-31.
In particular, we compared the models' performance considering the median over all sensors of NBIAS, RMSE, MAPE, Nash–Sutcliffe Efficiency (NSE), and Kling–Gupta Efficiency (KGE), described in Equations \ref{eq:rmse}, \ref{eq:nbias}, \ref{eq:mape}, \ref{eq:nse}, and \ref{eq:kge}, computed between true data $h$ and the prediction $\hat{h}$ for every sensor over the test set of length $N = 104$ (weeks), and with $\sigma$ and $\mu$ representing the variance and mean respectively. 
To evaluate further the model's generalization ability, we tested the models feeding both the true lags as input and in a rollout setting using the predictions at the previous time step as the lagged input for the whole test, i.e. $\hat{h}_{\mathcal{P},t^*} = \hat{f}_\theta(H_{\mathcal{P},t^*-1},\mathbf{V}_{t^*})$ and $\hat{h}_{\mathcal{P},t^*} = \hat{f}_\theta(\hat{H}_{\mathcal{P},t^*-1},\mathbf{V}_{t^*})$, respectively.
Thus, in the last case, models see only true groundwater level data at the beginning of the test set (2022-01-01), and then they iterate their predictions to produce the whole test set's predictions.}

\begin{equation}
RMSE = \sqrt{\frac{1}{N}\sum_{i}^{N} ( \hat{h_i} - h_i )^2} 
\label{eq:rmse}
\end{equation}

\begin{equation}
NBIAS = \frac{\frac{1}{N}\sum_{i}^{N} (\hat{h_i} - h_i)}{h_{max} - h_{min} } 
\label{eq:nbias}
\end{equation}

\begin{equation}
MAPE = 100\frac{1}{N}\sum_{i}^{N} \frac{|\hat{h_i} - h_i|}{h_i} 
\label{eq:mape}
\end{equation}

\begin{equation}
NSE = 1 - \frac{\sum_{i}^{N} ( \hat{h_i} - h_i )^2}{\sum_{i}^{N} ( h_i - \overline{h}  )^2}
\label{eq:nse}
\end{equation}

\begin{equation}
    \begin{gathered}
    KGE = 1 - \sqrt{(\rho-1)^2 + (\alpha - 1 )^2 + (\beta - 1)^2 } \\
    \alpha = \frac{\sigma_{\hat{h}} }{\sigma_h};
    \beta = \frac{\mu_{\hat{h}} }{\mu_h}
    \end{gathered} 
   \label{eq:kge}
\end{equation}

\rev{
Moreover, we further evaluated the ability of the best-performing model in reproducing the full available time series (starting from 2001-01-01) for all sensors.
In detail, using the model providing the best test metrics, we predicted all the time series in a rollout setting with a forecast horizon of 26 weeks (i.e., 6 months). 
This is in line with the real situation in our case study, in which the ARPA releases groundwater level data on a semester basis.}

\subsubsection{Correlation Analysis}

\rev{Two aspects characterize our case study: the substantial amount of missing data and the heterogeneous behavior of the groundwater level time series.
Therefore, we investigated the association between the performance metrics and both the proportion of missing values and the trend and seasonal components' strengths of the time series.}

\rev{In particular, we considered the Pearson's correlation coefficient $\rho$ in Equation \ref{eq:rho}}

\begin{equation}
    \rho (u,h) = \frac{\sum_{i=1}^{n} (u_i - \bar{u})(h_i - \bar{h})}
{\sqrt{\sum_{i=1}^{n} (u_i - \bar{u})^2}
 \sqrt{\sum_{i=1}^{n} (h_i - \bar{h})^2}}
\label{eq:rho}
\end{equation}

\rev{Where $u$ is the quantity of interest, i.e. the missing values quantity or the trend and seasonal components strengths.}

\rev{
Concerning the decomposition of the groundwater level data, we decomposed each time series with a Seasonal and Trend decomposition using LOESS\footnote{LOESS stands for LOcally Estimated Scatterplot Smoothing, which is a non-parametric local non-linear regression statistical method.} (STL)~\cite{cleveland_1990}.
This method assumes that each time series $Y$ is a summation of three terms, namely a trend component $T$, a seasonal component $S$, and a residual component $R$, in other words, $Y = T + S + R$.
Firstly, the trend component $T$ is estimated using LOESS to capture the overall direction of the series by smoothing short-term fluctuations.
Then, on the residual between $T$ and the original $Y$, the seasonal term $S$ is estimated to extract repeated patterns occurring at the yearly scale.
Lastly, the residual component $R$ is determined as the remaining fluctuations not captured by the other two terms\footnote{For more technical details, please refer to \cite{cleveland_1990}.}.}

\rev{
We then adopted the metrics in Equation~\ref{eq:trend_strength} and  \ref{eq:seasonal_strength} to quantify the strength of the seasonal and trend components, the Trend Strength $\mathcal{S}_T$ and the Seasonal Strength $\mathcal{S}_S$, respectively~\cite{wang_2006}:
}

\begin{equation}
    \mathcal{S}_T = max\left(0,1-\frac{Var(R)}{Var(T+R)}\right)
\label{eq:trend_strength}
\end{equation}
\begin{equation}
    \mathcal{S}_S = max\left(0,1-\frac{Var(R)}{Var(S+R)}\right)
\label{eq:seasonal_strength}
\end{equation}

\subsubsection{Qualitative Analysis}
\rev{
As already mentioned in Section~\ref{sec:lb_strategy}, to impose the learning biases, we generated a grid of predictions corresponding to maps with a spatial resolution of $1.5$km.
Given that we could not supervise these gridded predictions with real groundwater data, we performed visual inspections of the estimated diffusion and residual components supported by experts to evaluate the physical plausibility of the estimations.
We want to highlight that this evaluation is intended to evaluate the plausibility of the estimated values and not to state conclusions on the spatial generalization ability of the models, given that non-ground truth data are available for such considerations.
}

\section{Experiments}

\rev{To find the best settings for all our models, namely STAINet, PSTAINet-IB, PSTAINet-ILB, and PSTAINet-ILRB, we performed a fine-tuning with a manual grid search of the optimal hyperparameters, training from 2001-01-01 to 2019-12-31 and validating on data from 2020-01-01 to 2021-12-31.
Specifically, to ensure that the optimal settings were chosen for each model type, we looked at the convergence of the loss terms and the validation performance metrics. 
Additionally, for the models using learning biases, we performed visual inspections of the estimated equation components together with domain experts, tuning the weights of the additional physics loss terms to best meet the experts’ expectations.}

We thus set for all models the embedding dimension equal to 32 and the number of heads for each MHA layer to 8. 
Under these settings, STAINet contains a total of 62081 trainable parameters, while PSTAINet-IB, PSTAINet-ILB, and PSTAINet-ILRB each comprise 49666 trainable parameters.
The dropout rate in the initial Spatio-Temporal Self-Attention block of the Autoregressive Module was set to $0.15$ for all configurations.

All models were trained using AdamW optimizer~\cite{adamw2019} with a learning rate of $2.5\cdot 10^{-4}$ decreased to $10^{-4}$ at epoch 400, and a weight decay equal to $2.5\cdot 10^{-3}$.
The total number of epochs to achieve a relevant optimum was found to be 1000 for STAINet, 850 for PSTAINet-IB and PSTAINet-ILRB, and 600 for PSTAINet-ILB.
The weights assigned to the individual loss components are reported in Table~\ref{tab:hyperparams_pg}.

\rev{The choice of setting similar hyperparameters (e.g., embedding dimension and learning rate) for all the models was made to perform the fairest comparison, but also because of an overall similar structure of the models.
Indeed, even if the tail of the architecture changes between the STAINet and the three physics-guided models, the backbone remains the same.
Our choice was then supported by the optimal performance on the validation set of this configuration.}

\rev{
After fine-tuning, and for final evaluation on the test set, all models were retrained using both the training and validation data, from 2001-01-01 to 2021-12-31 (hereafter referred to as the training set for simplicity), and evaluated on the period from 2022-01-01 to 2023-12-31.\footnote{The code is available here \url{https://github.com/Matteo-Salis/physics-guided-gwl} along with GIFs representing the evolution of the prediction maps through time.}.}

\begin{table*}[h!]
\centering
\caption{Loss terms weights.}
\begin{tabular}{l|cccccc} 
\toprule
\textbf{Model} & $\alpha_{\text{ortho}}$ & $\alpha_{\text{coh}}$ & $\alpha_{\text{diff}}$ & $\alpha_{||\mathcal{R}||_1}$ & $\mathbf{\alpha_{||\mathcal{R}||_2}}$ & $\alpha_{\text{RCH}}$ \\
\cmidrule(lr){1-7}
\morecmidrules
\cmidrule(lr){1-7}
STAINet           & $10^{-3}$ & -   & -        & -      & -      & - \\ 
PSTAINet-IB        & $10^{-3}$ & -   & -        & -      & -      & - \\ 
PSTAINet-ILB      & $10^{-3}$ & $1$ & $2.5\cdot 10^{-2}$ & $5\cdot 10^{-4}$ & $10^{-4}$ & - \\ 
PSTAINet-ILRB  & $10^{-3}$ & $1$ & $2.5\cdot 10^{-2}$ & $5\cdot 10^{-4}$ & $10^{-4}$ & $5\cdot 10^{-4}$\\ 
\bottomrule
\end{tabular}
\label{tab:hyperparams_pg}
\end{table*}

\section{Results}
\label{sec:results}
\begin{figure*}[!h]
    \centering
    \includegraphics[width=0.82\linewidth]{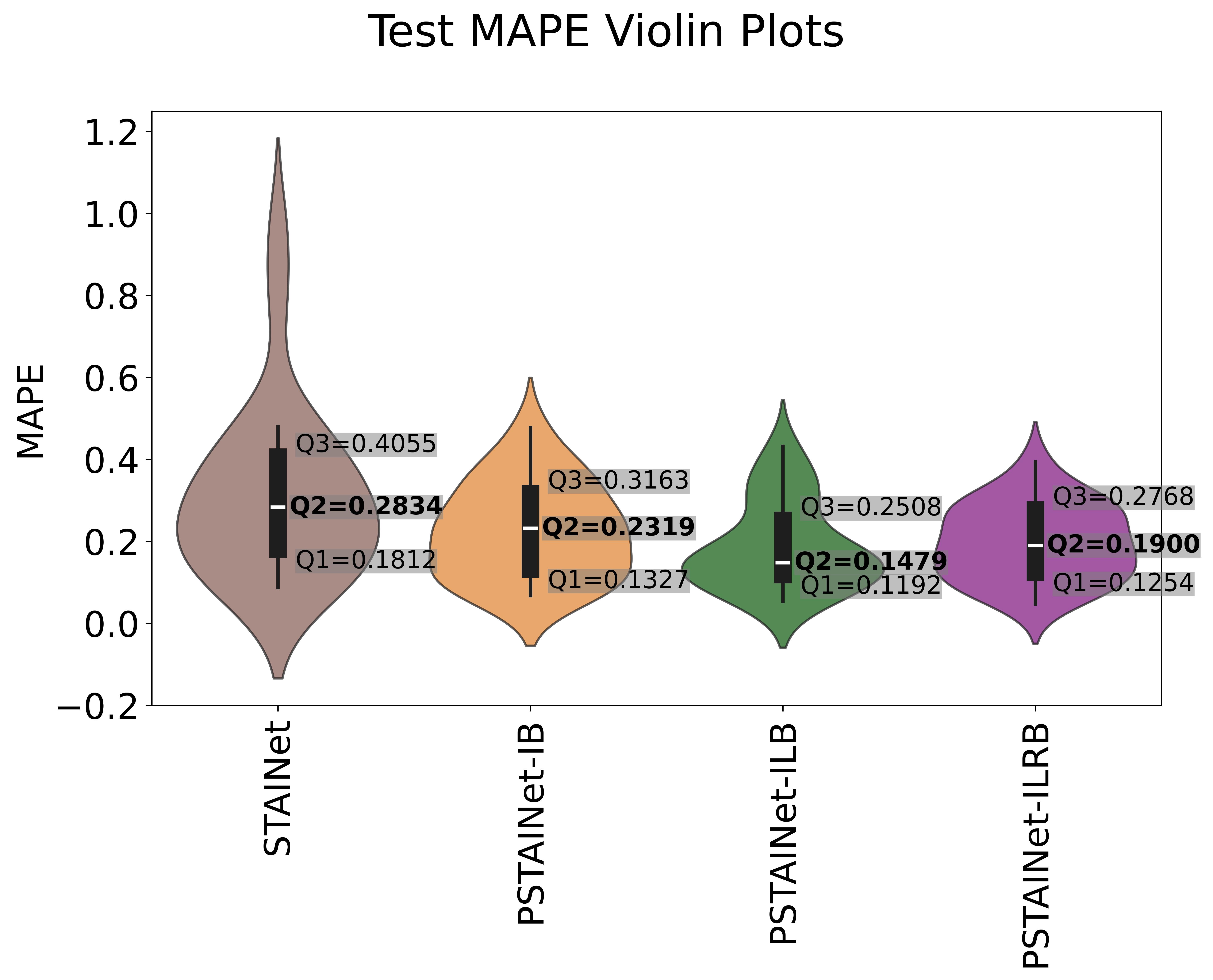}
    \caption{Violin plots representing MAPE values for the proposed models on the test set. \textbf{Q1}, \textbf{Q2}, \textbf{Q3} represent the first, second (median), and third quartiles, respectively. The shapes around the box plots are the KDE density estimates of the MAPE values over the sensors.}
    \label{fig:mape_violins}
\end{figure*}

Table~\ref{tab:PI_metrics} reports the median \rev{and the interquartile Range (IQR, i.e., the difference between the 3rd and 1st quartiles) for the} metrics over all sensors' predictions feeding real data and in the rollout setting on the test set.
\rev{Figure~\ref{fig:mape_violins} depicts the violin plots for the MAPE on the test set feeding true lagged groundwater data.}
Both in the rollout setting and using true data, the best-performing model was PSTAINet-ILB, \rev{providing better metrics with lower IQR values}. 
In general, all models with the physics prior performed better than the pure deep learning approach (STAINet), suggesting that both the inductive bias and the learning bias strategies provided useful information to the models.
\rev{In particular, PSTAINet-IB, PSTAINet-ILB, and PSTAINet-ILRB reduced the MAPE by 18\%, 48\%, and 33\%, respectively, when fed by true lagged groundwater data.
Furthermore, these models substantially reduced the long right tail of the MAPE distribution observed for STAINet (Figure~\ref{fig:mape_violins}).
In terms of KGE, the improvements were even more pronounced, with increases of 65\%, 129\%, and 94\%, respectively.}

\newcommand{\iqr}[1]{\scriptsize\textit{(#1)}}
\begin{table*}[h]
\centering
\caption{Median performance metrics on the test set feeding true data (True Data) and in the rollout setting (Rollout). Numbers in parentheses represent the Interquartile Range (IQR).}
\begin{tabular}{l|cc|cc|cc|cc}
\toprule
\textbf{}
& \multicolumn{2}{c|}{\textbf{STAINet}}
& \multicolumn{2}{c|}{\textbf{PSTAINet-IB}}
& \multicolumn{2}{c|}{\textbf{PSTAINet-ILB}}
& \multicolumn{2}{c}{\textbf{PSTAINet-ILRB}} \\
\cmidrule(lr){2-9}
& \textbf{\makecell{True \\ Data}} & \textbf{Rollout}
& \textbf{\makecell{True \\ Data}} & \textbf{Rollout}
& \textbf{\makecell{True \\ Data}} & \textbf{Rollout}
& \textbf{\makecell{True \\ Data}} & \textbf{Rollout} \\
\cmidrule(lr){1-9}
\morecmidrules
\cmidrule(lr){1-9}

\multirow{2}{*}{NBIAS}
& 0.1042 & 0.1713
& 0.0594 & 0.0862
& -0.0093 & 0.0124
& 0.0177 & 0.0685 \\[-4pt]
& \iqr{0.2297} & \iqr{0.2297} 
& \iqr{0.1741} & \iqr{0.2028}
& \iqr{0.1137} & \iqr{0.1351}
& \iqr{0.2212} & \iqr{0.1636} \\

\multirow{2}{*}{RMSE [m]}
& 0.8351 & 1.0292
& 0.6296 & 0.7336
& 0.5058 & 0.5851
& 0.6521 & 0.6783 \\[-4pt]
& \iqr{0.8635} & \iqr{1.2064}
& \iqr{0.8175} & \iqr{0.9495}
& \iqr{0.3031} & \iqr{0.4485}
& \iqr{0.5347} & \iqr{0.8296} \\

\multirow{2}{*}{MAPE [\%]}
& 0.2834 & 0.2868
& 0.2319 & 0.2606
& 0.1479 & 0.1604
& 0.1900 & 0.1953 \\[-4pt]
& \iqr{0.2243} & \iqr{0.3159}
& \iqr{0.1836} & \iqr{0.2240}
& \iqr{0.1316} & \iqr{0.1505}
& \iqr{0.1514} & \iqr{0.1984} \\

\multirow{2}{*}{NSE}
& 0.5612 & 0.2884
& 0.7059 & 0.6269
& 0.8025 & 0.7876
& 0.7986 & 0.7024 \\[-4pt]
& \iqr{0.4800} & \iqr{0.2613}
& \iqr{0.3584} & \iqr{0.2327}
& \iqr{0.2903} & \iqr{0.3036}
& \iqr{0.2073} & \iqr{0.1896} \\

\multirow{2}{*}{KGE}
& 0.2583 & 0.2236
& 0.4253 & 0.5196
& 0.5917 & 0.5800
& 0.5018 & 0.4074 \\[-4pt]
& \iqr{0.6213} & \iqr{0.5469}
& \iqr{0.5158} & \iqr{0.4690}
& \iqr{0.3046} & \iqr{0.3441}
& \iqr{0.3471} & \iqr{0.4266} \\

\bottomrule
\end{tabular}
\label{tab:PI_metrics}
\end{table*}

On most of the metrics, PSTAINet-ILRB performed better than PSTAINet-IB, but worse than PSTAINet-ILB. This might suggest that the recharge zones prior was too restrictive, limiting the model's representation ability.
The metrics computed on every sensor and every model in the rollout setting are reported in Section \ref{sec:appx_test_set_metr_pred} of the Supplementary Materials, specifically in Table~\ref{tab:PI_sens_metrics_STNet}, Table~\ref{tab:PI_sens_metrics_STDisNet}, Table~\ref{tab:PI_sens_metrics_STDisNetPI}, Table~\ref{tab:PI_sens_metrics_STDisNetPI-RCH}.

Figure~\ref{fig:focus_pred} depicts the predictions on the test set, feeding true data and iterating the predictions (rollout setting), respectively, for the sensors in Racconigi and Scalenghe.
In Section \ref{sec:appx_test_set_metr_pred}, we reported the predictions over the test set for all the sensors, in the rollout setting (Figure~\ref{fig:all_pred_iter}).

From Figure~\ref{fig:focus_pred}, the good generalization ability of the PSTAINet-ILB model is visible, especially in contexts of unprecedented drop, like in the summers of 2022 and 2023.
PSTAINet-ILRB also predicted the groundwater level ‌decreases, but in some cases (like in Scalenghe, Figure~\ref{fig:sca_pred_iter}) the drops were misquantified, and in other cases (like Racconigi, Figure~\ref{fig:racc_pred_iter}) a premature recharge was predicted.\\
\noindent Even if differences between feeding true data and the rollout setting are visible (Table \ref{tab:PI_metrics} and Figure \ref{fig:focus_pred}), the PSTAINet-ILB model maintained almost the same performance.
This further demonstrates the good generalization ability of the PSTAINet-ILB model.

\begin{figure*}[h]
    \centering
    \includegraphics[width=0.76\linewidth]{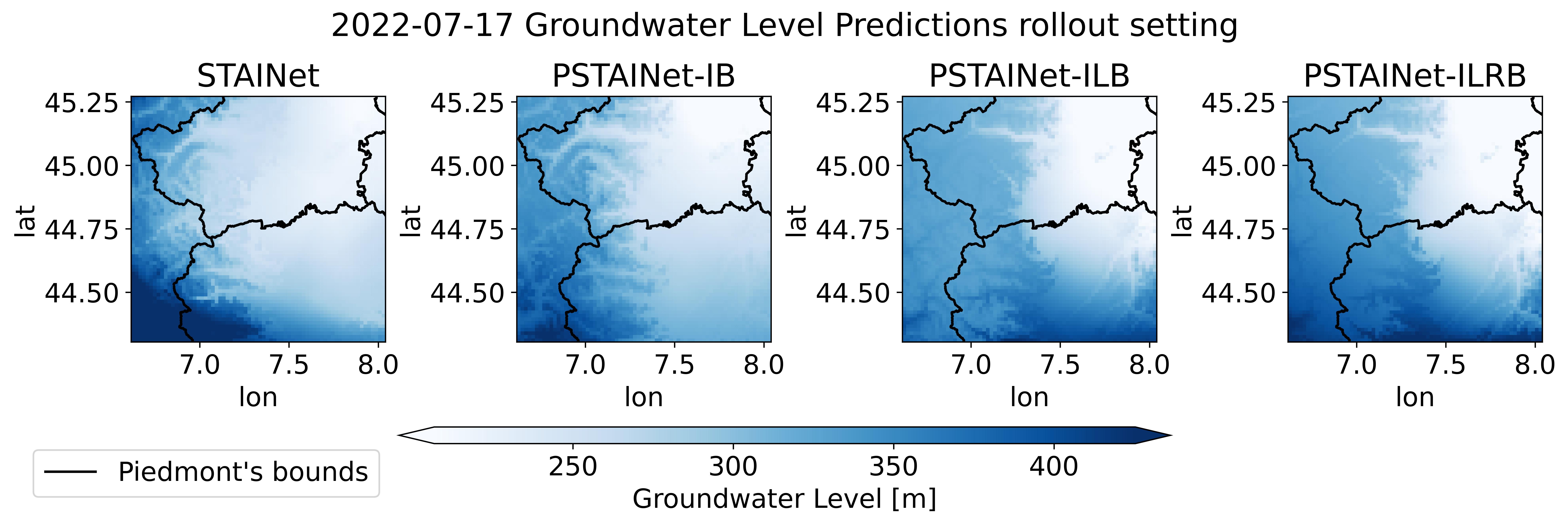}
    \caption{STAINet, PSTAINet-IB, PSTAINet-ILB, and PSTAINet-ILRB groundwater level prediction maps at $1.5$km/pixel for 2022-07-17.}
    \label{fig:pred_map_gwl}
\end{figure*}

Figure~\ref{fig:pred_map_gwl} shows the map predictions at a resolution of $1.5$ km/pixel for the week starting on 2022-07-17. 
All models predicted a map in line with domain experts' expectations (i.e., with higher groundwater levels in the mountain region).
However, the PSTAINet-ILB and PSTAINet-ILRB predictions appear to be less definite on the north-west mountains, but they better identified the rivers and valleys in the southeastern part.
Notwithstanding, a formal evaluation of these results is difficult because of the lack of true measured data all over the ROI, especially in the mountains -- a universal problem in hydrology.

\begin{figure}[h]
    \centering
    \begin{subfigure}{\linewidth}
        \centering
        \includegraphics[width=0.75\linewidth]{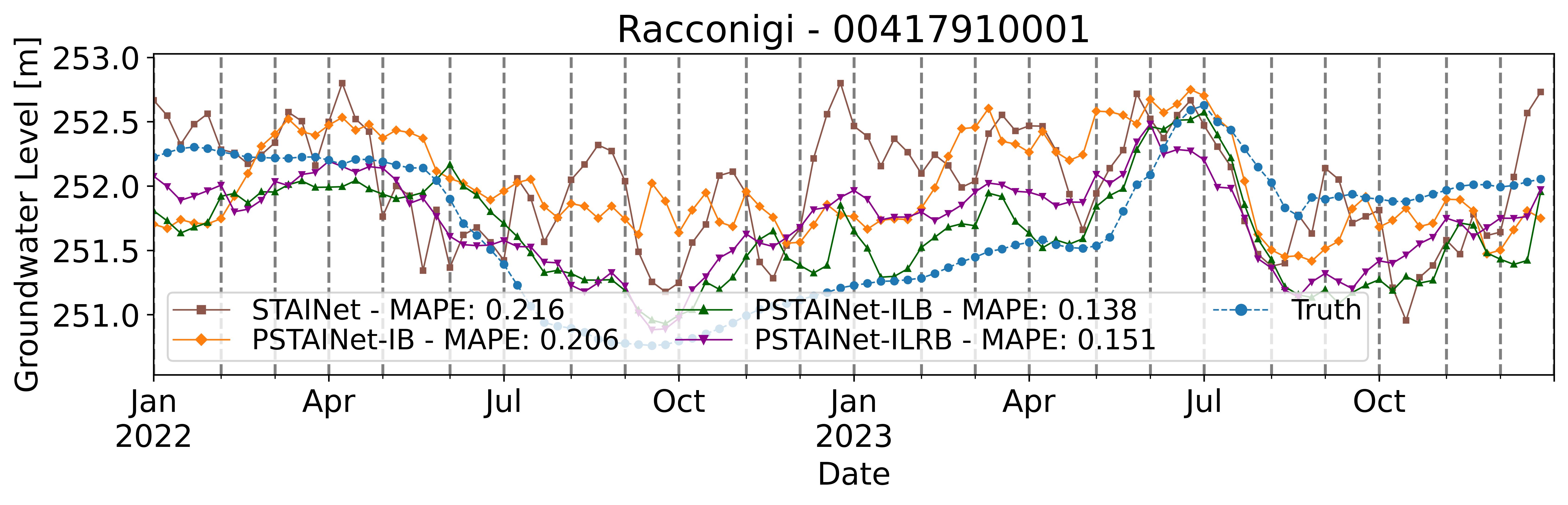}
        \caption{}
        \label{fig:racc_pred_true}
    \end{subfigure}
    \begin{subfigure}{\linewidth}
        \centering
        \includegraphics[width=0.75\linewidth]{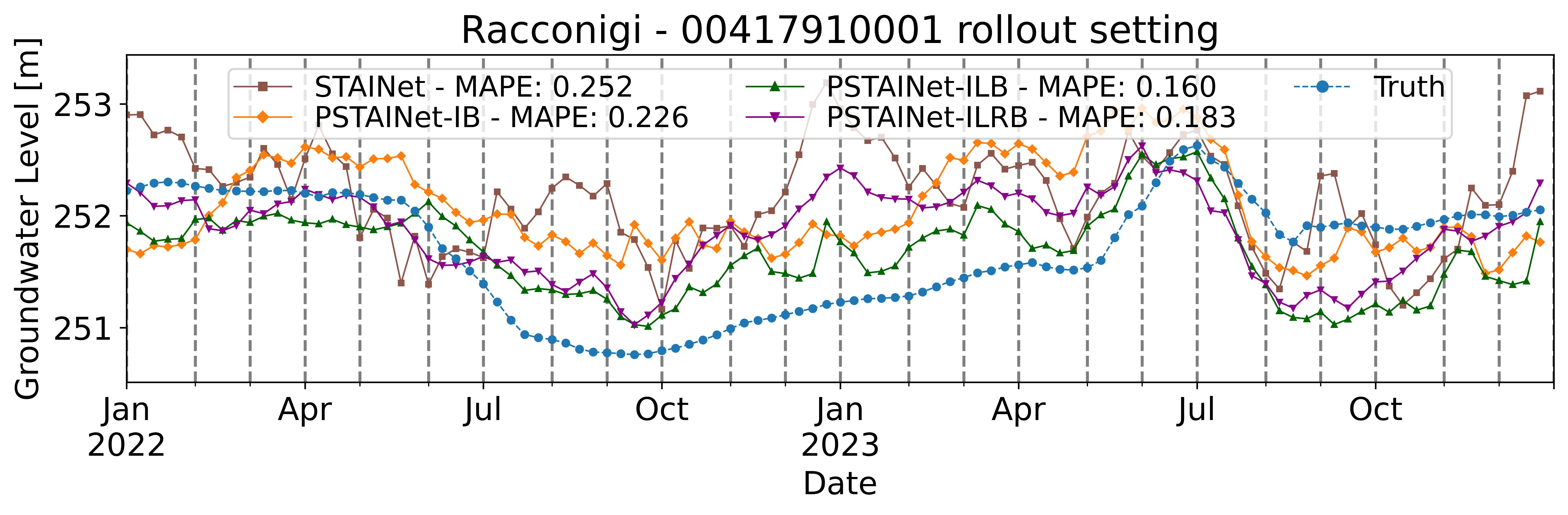}
        \caption{}
        \label{fig:racc_pred_iter}
    \end{subfigure}
    \begin{subfigure}{\linewidth}
        \centering
        \includegraphics[width=0.75\linewidth]{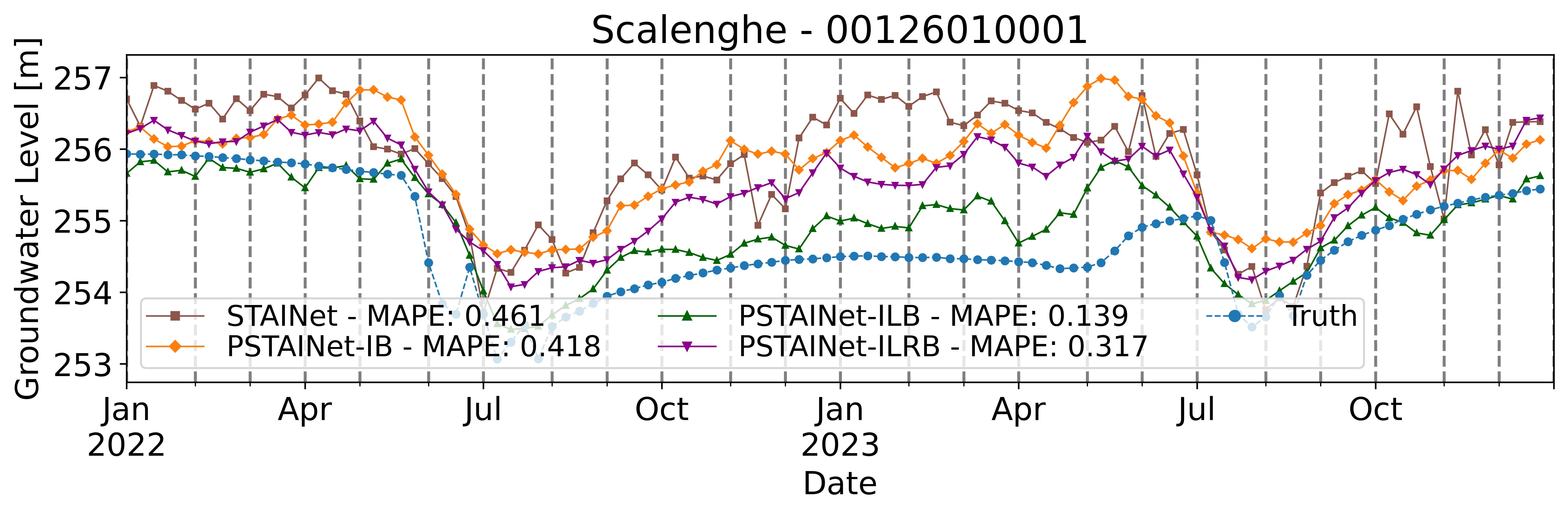}
        \caption{}
        \label{fig:sca_pred_true}
    \end{subfigure}
\end{figure}

\begin{figure}[h]
    \centering
    \addtocounter{figure}{-1}  
    \begin{subfigure}{\linewidth}
        \centering
        \addtocounter{subfigure}{3} 
        \includegraphics[width=0.75\linewidth]{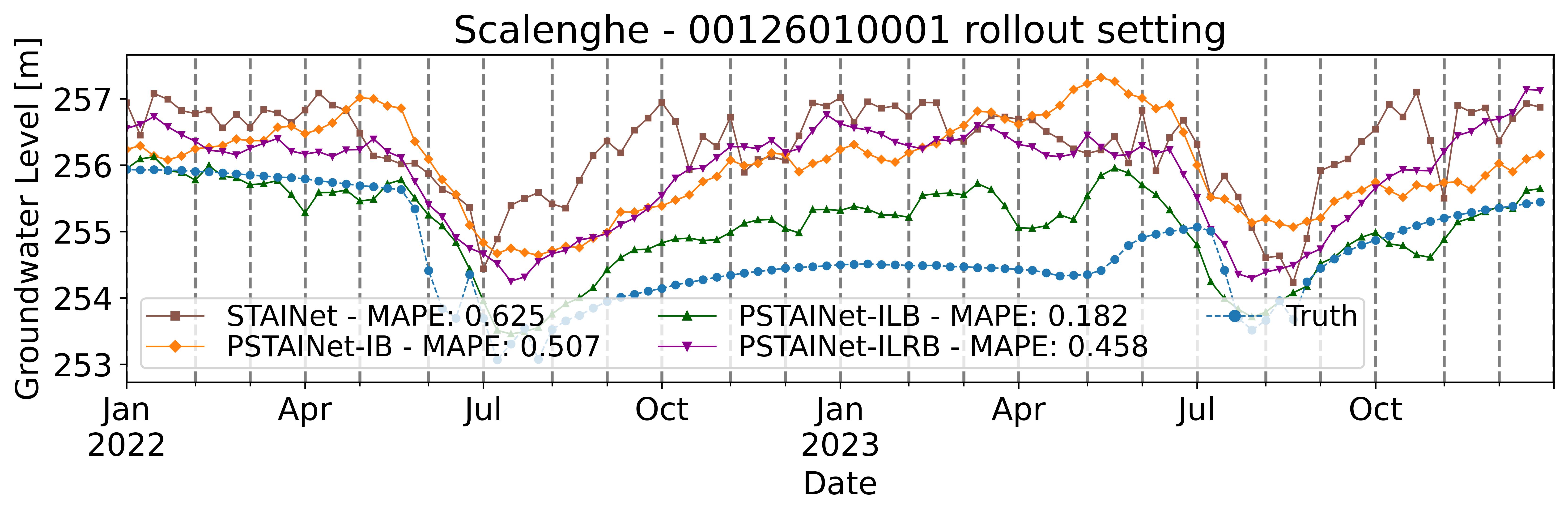}
        \caption{}
        \label{fig:sca_pred_iter}
    \end{subfigure}
    \caption{STAINet, PSTAINet-IB, PSTAINet-ILB, and PSTAINet-ILRB predictions on the test set feeding true data in \textbf{a)}, \textbf{c)} and in the rollout setting \textbf{b)}, \textbf{d)} in Racconigi and Scalenghe sensors.}
    \label{fig:focus_pred}
\end{figure}

\subsection{Predicted Equation Components}
\begin{figure}[h]
    \centering
    \begin{subfigure}{0.80\linewidth}
        \includegraphics[width=\linewidth]{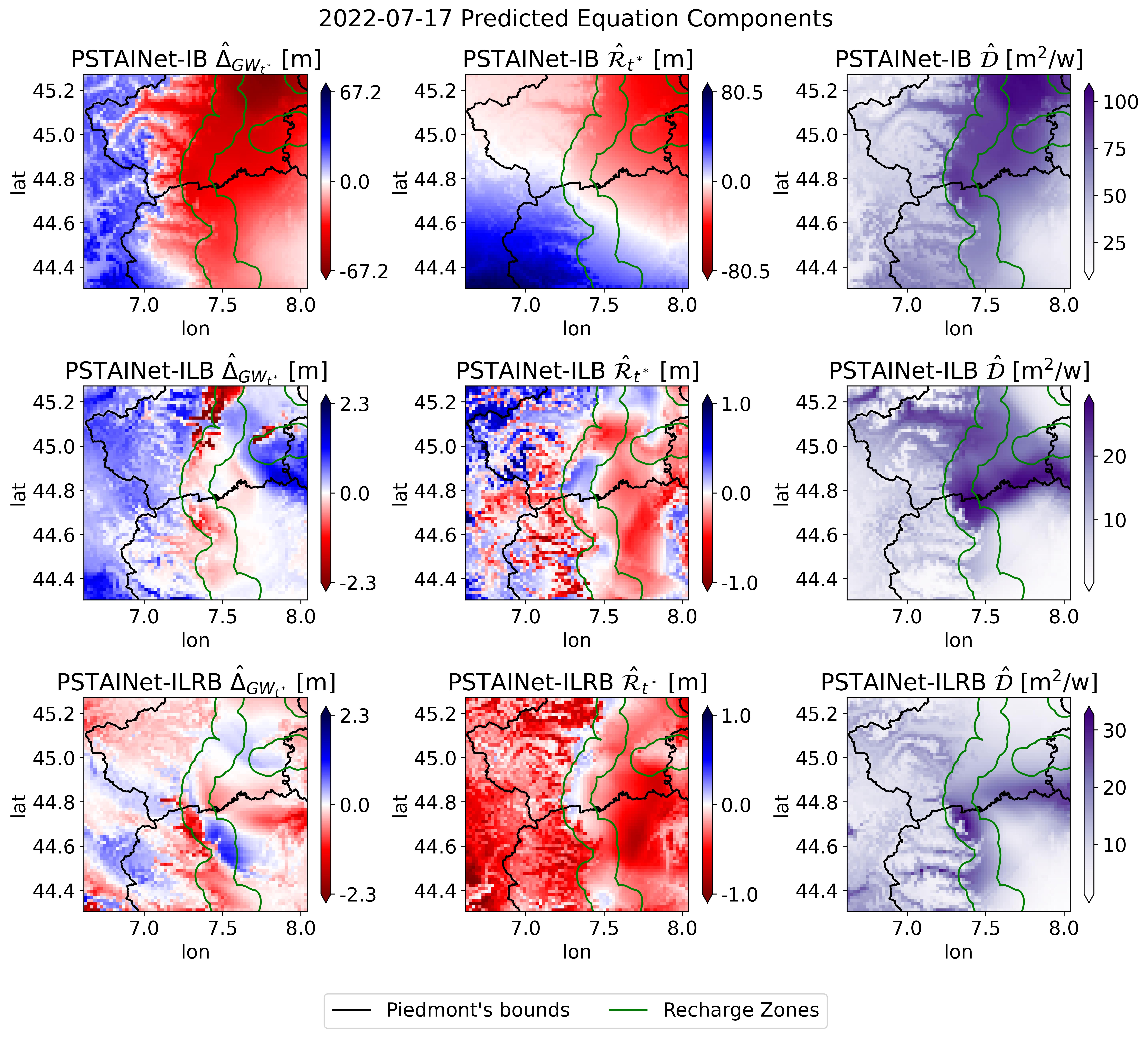}
        \caption{}
        \label{fig:pred_map_deltas_jul}
    \end{subfigure}
    \begin{subfigure}{0.80\linewidth}
        \includegraphics[width=\linewidth]{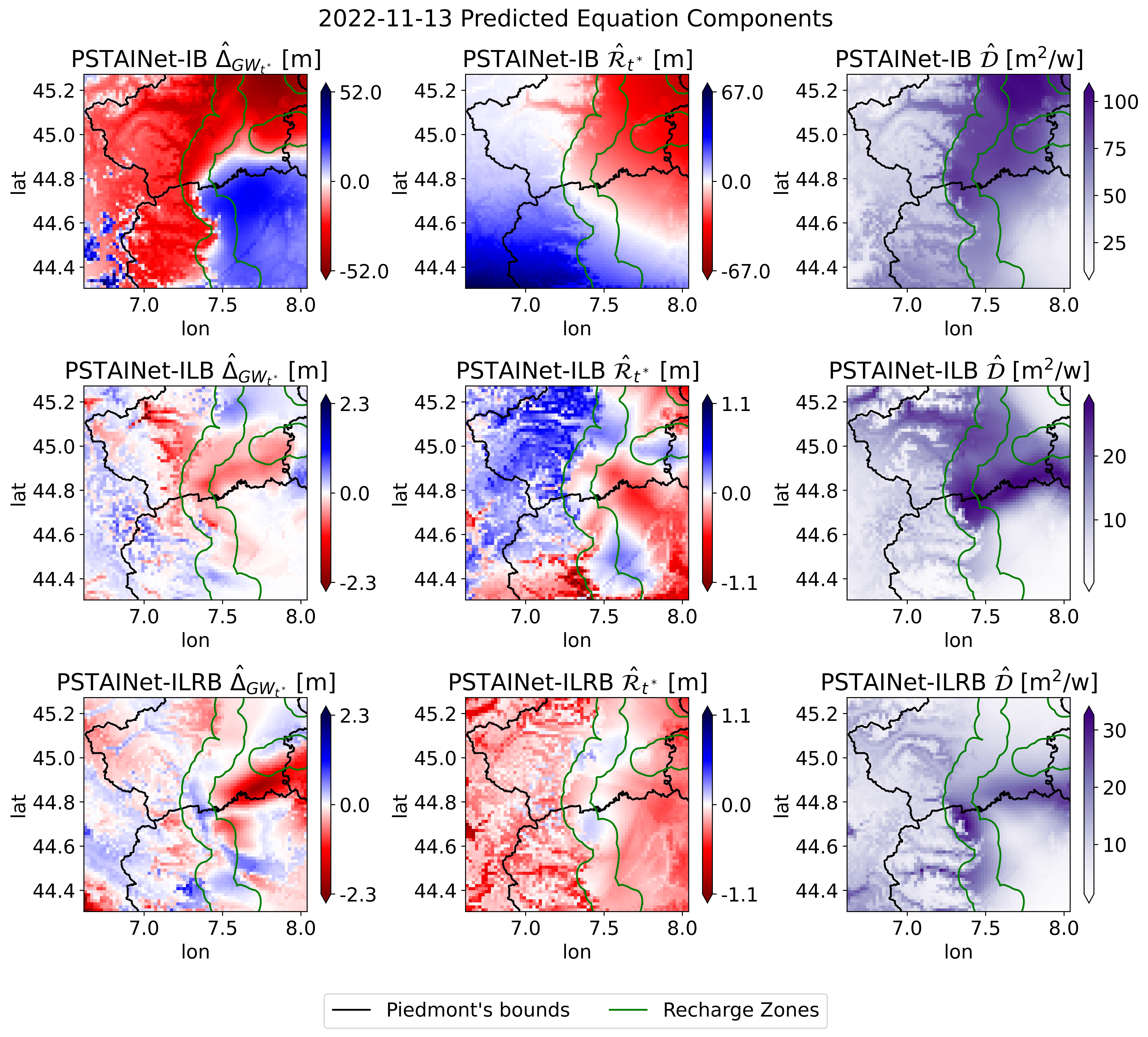}
        \caption{}
        \label{fig:pred_map_deltas_nov}
    \end{subfigure}
    \caption{Predicted equation components by PSTAINet-IB, PSTAINet-ILB, and PSTAINet-ILRB for the weeks starting on 2022-07-17 (\textbf{a}), and 2022-11-13 (\textbf{b}). The first column reports $\hat{\Delta}_{GW_{t^*}}$, the second $\hat{\mathcal{R}}_{t^*}$, and the last $\hat{\mathcal{D}}$. Green lines identify the recharge zones. For $\hat{\Delta}_{GW_{t^*}}$, a blue pixel means inflow while a red one means outflow; for $\hat{\mathcal{R}}_{t^*}$, a blue pixel means recharge while red represents a loss.}   
\end{figure}

Apart from the better performance achieved, the proposed physics-guided models also provided an estimation of the terms of Equation~\eqref{eq:gw_pde_lag}.
To assess the physical plausibility of the predictions, we analyzed the prediction maps of the equation components, in particular, in the rollout setting. Figures~\ref{fig:pred_map_deltas_jul} and \ref{fig:pred_map_deltas_nov} show the predicted PDE component on weeks starting on 2022-07-17 and 2022-11-13, respectively.

PSTAINet-IB predicted a general displacement of groundwater from the mountains to the plain.
However, the very high absolute value of $\hat{\Delta}_{GW_{t^*}}$ and $\hat{\mathcal{R}}_{t^*}$ undermined their physics soundness.\\
\noindent Differently, both PSTAINet-ILB and PSTAINet-ILRB provided insightful predictions also from the hydrological point of view.
In particular, concerning the diffusion component $\hat{\Delta}_{GW_{t^*}}$, PSTAINet-ILB predicted during the recharge period (e.g., 2022-11-13 in Figures \ref{fig:pred_map_deltas_nov}) a more pronounced groundwater discharge from mountains, while during the dry season, given the lower water content and precipitation, the discharge is limited, with water being more retained in the valleys and beneath the mountains. 
In contrast, the PSTAINet-ILRB model predicted a diffusion dynamic that remained more stable across different periods. Specifically, it captured a general outflow from the valleys toward the plain.\\
\noindent The two learning biased models learnt different physics relationships, and even though both appeared sensible, a rigorous evaluation of these processes is difficult, given the scarce knowledge of groundwater processes under mountains.

Looking at the sink/source predicted component $\hat{\mathcal{R}}_{t^*}$, the PSTAINet-ILB model detected more pronounced and extended recharge areas during the recharging season (Figures \ref{fig:pred_map_deltas_nov}), while a sparser pattern with alternating recharge and loss zones in the dry season (Figure~\ref{fig:pred_map_deltas_jul}) -- probably because of more abstractions or higher temperatures in summer.
Concerning the $\hat{\mathcal{R}}_{t^*}$ predictions made by PSTAINet-ILRB, the recharge zone regularization effect is clearly visible, limiting positive values of $\hat{\mathcal{R}}_{t^*}$ only into the delimited recharge zones (in green in Figures \ref{fig:pred_map_deltas_jul} and \ref{fig:pred_map_deltas_nov}).
Although this constraint is hydrologically reasonable, it may have been too restrictive in our case, where the shallow groundwater body is strongly influenced by surface forcing, both human and natural.\\
Concerning the values of the predicted term $\hat{\mathcal{D}}$, all models produced plausible values. 
However, PSTAINet-ILB and PSTAINet-ILRB reported a more sensible spatial pattern with higher values in the valley that gradually attenuated through the plain.

In conclusion, the predicted equation component analysis suggests that the absence of supervision on the equation components may be problematic.
Introducing regularization terms on $\hat{\Delta}_{GW_{t^*}}$ and $\hat{\mathcal{R}}_{t^*}$ enabled the models to generate results that are more in line with physics principles.

\clearpage
\subsection{Time Series Reconstruction}
\label{sec:rec_task}

\begin{figure}[h]
    \centering
    \begin{subfigure}{\linewidth}
        \centering
        \includegraphics[width=0.8\textwidth]{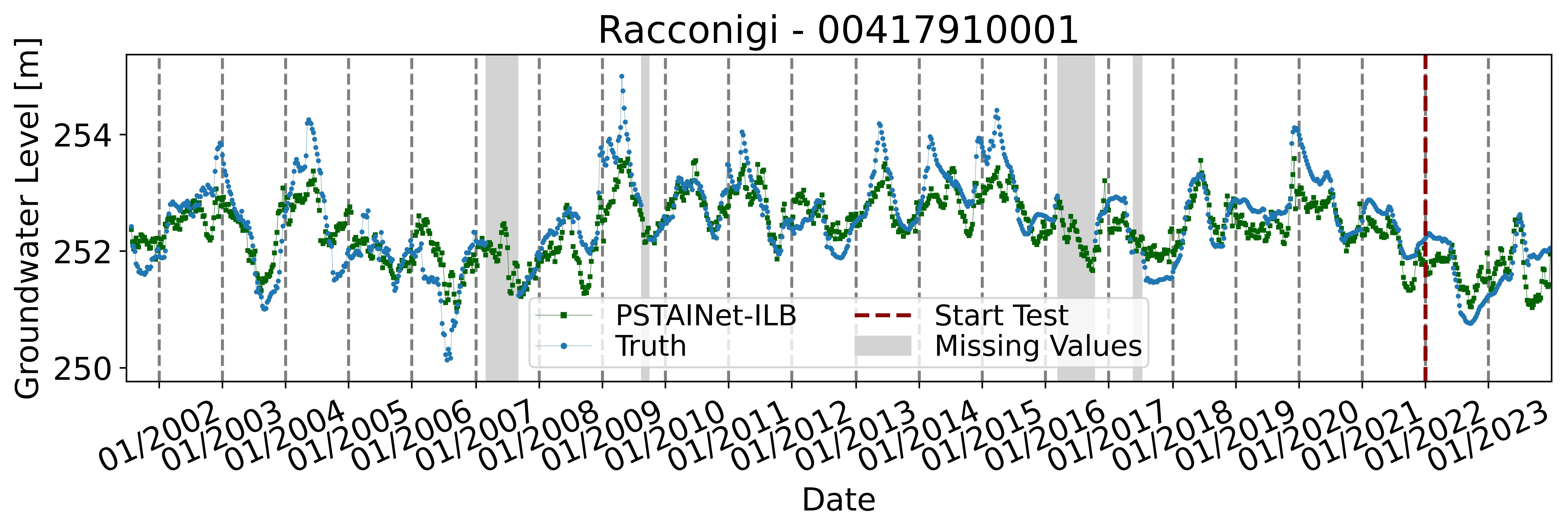}
        \caption{}
        \label{fig:racc_rec}
    \end{subfigure}
    \begin{subfigure}{\linewidth}
        \centering
        \includegraphics[width=0.8\textwidth]{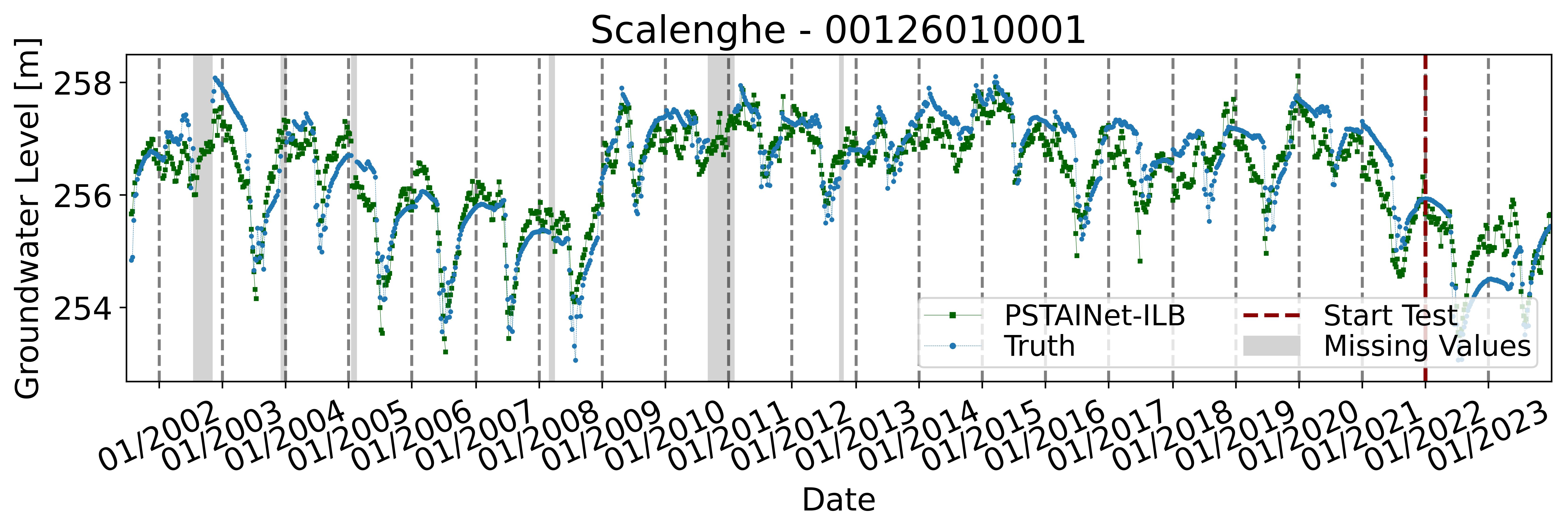}
        \caption{}
        \label{fig:sca_rec}
    \end{subfigure}
    \caption{PSTAINet-ILB time series reconstruction in the rollout setting with a 26 time step forecast horizon for \textbf{a)} Racconigi and \textbf{b)} Scalenghe piezometers.}
    \label{fig:focus_rec}
\end{figure}

Table~\ref{tab:PI_metrics_rec} reports the median evaluation metrics of the PSTAINet-ILB over all sensors, and Figure~\ref{fig:focus_rec} depicts the predicted time series for Racconigi and Scalenghe piezometers.
In Section \ref{sec:appx_ts_rec}, we reported the metrics (Table~\ref{tab:rec_metrics_all}) and the predicted time series (Figure~\ref{fig:all_rec}) for all sensors separately.
The model demonstrated strong capability in reconstructing the full time series, even during intervals with missing data, accurately reproducing seasonal dynamics as well as short-term fluctuations.

\begin{table*}[!h]
\centering
\caption{PSTAINet-ILB time series reconstruction median Metrics, in the rollout setting with a 26 time step forecast horizon.}
\begin{tabular}{l|ccccc}
\toprule
\textbf{} & \textbf{NBIAS} & \textbf{RMSE [m]} & \textbf{MAPE [\%]} & \textbf{NSE} & \textbf{KGE} \\
\cmidrule(lr){1-6}
\morecmidrules
\cmidrule(lr){1-6}
PSTAINet-ILB & -0.0278 & 0.4490 & 0.1404 & 0.6314 & 0.7547 \\
\bottomrule
\end{tabular}
\label{tab:PI_metrics_rec}
\end{table*}

\section{Discussion}
An extensive comparison with existing studies is challenging due to the specificity of our application and the scarcity of research employing physics-guided strategies on real groundwater level measurement data.

However, a reference study is available~\cite{salis_2024a}, in which the authors trained sensor-specific, purely data-driven neural networks on \rev{a small subset of groundwater sensors considered in our study, specifically the piezometers located in the municipality of Vottignasco, Savigliano, and Racconigi.
On these same piezometers, PSTAINet-ILB achieved comparable performance, with a median NBIAS of $-0.01$ versus $-0.05$ of \cite{salis_2024a}, an NSE of $0.84$ versus $0.90$, and a KGE of $0.66$ versus $0.43$.
Nevertheless, the two approaches differ substantially in their modeling strategy. In our study, a single model is trained jointly on all monitoring stations, simultaneously predicting groundwater levels at multiple locations through the attention mechanism. In contrast, the approach proposed in~\cite{salis_2024a} requires training a separate model for each piezometer. Consequently, extending the method to a larger monitoring network entails increased computational costs, as both training and inference must be performed independently for each sensor.
}

\rev{
Furthermore, the models in~\cite{salis_2024a} follow a purely data-driven paradigm and directly predict groundwater levels without providing additional physics information. In contrast, the proposed physics-guided approach not only achieves competitive predictive performance but also estimates physically interpretable equation components, thereby improving the trustworthiness of the model predictions.
}
Our performances are also consistent with those reported in other studies that adopt data-driven models to estimate groundwater levels in different geographical regions, such as \cite{LeeUsing2019} in the Yangpyeong riverside area (South Korea) and \cite{WunschGroundwater2021} in the Upper Rhine Graben area (central Europe), which achieved NSE values of nearly $0.8$ and $0.5$, respectively.

In the following, a more detailed analysis of the results of our proposed models is presented.

\subsection{Loss Analyses}

\begin{figure*}[h!]
    \centering
    \begin{subfigure}[b]{0.49\textwidth}
        \centering
        \includegraphics[width=\textwidth]{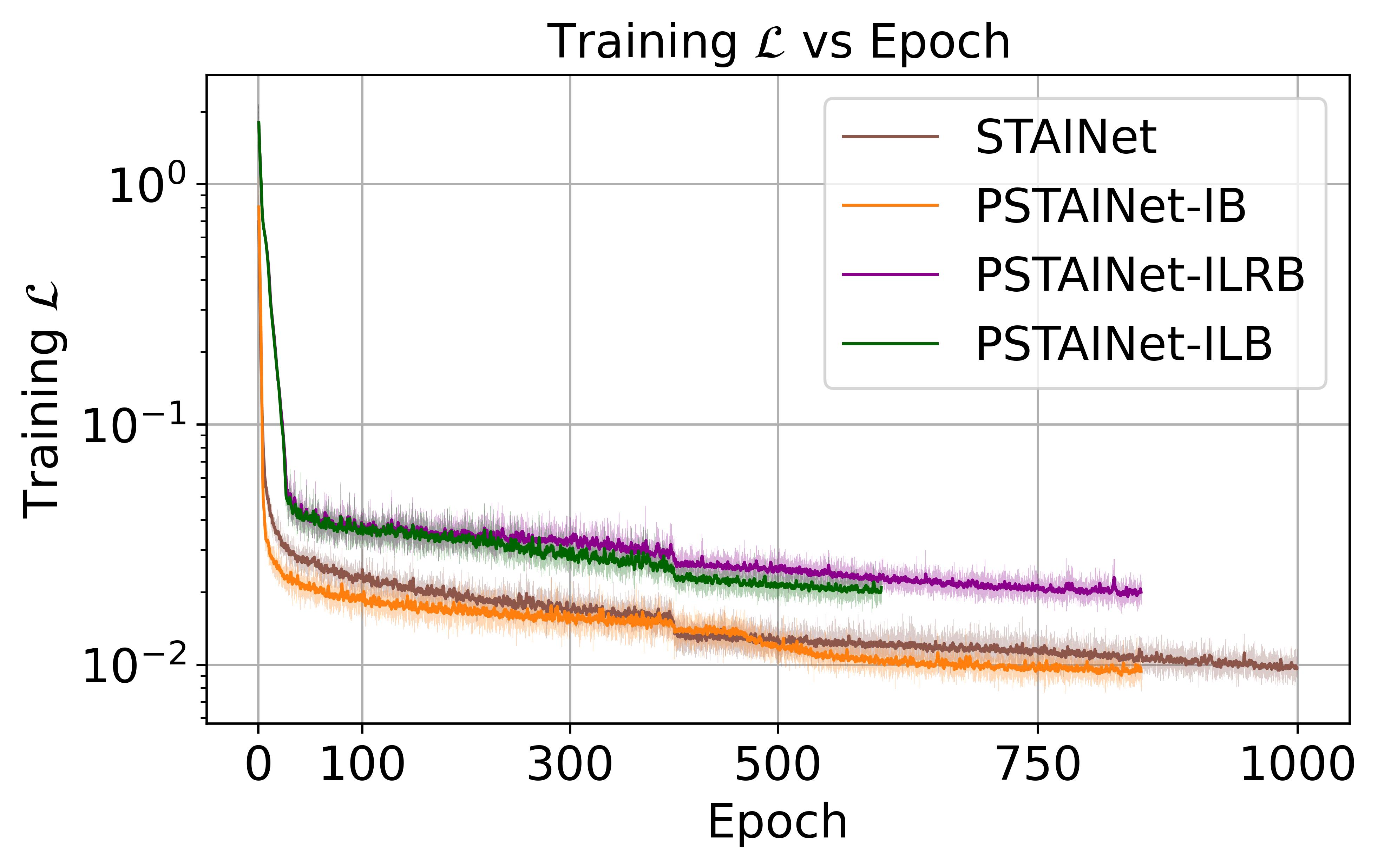}
        \caption{}
        \label{fig:train_loss}
    \end{subfigure}
    \begin{subfigure}[b]{0.49\textwidth}
        \centering
        \includegraphics[width=\textwidth]{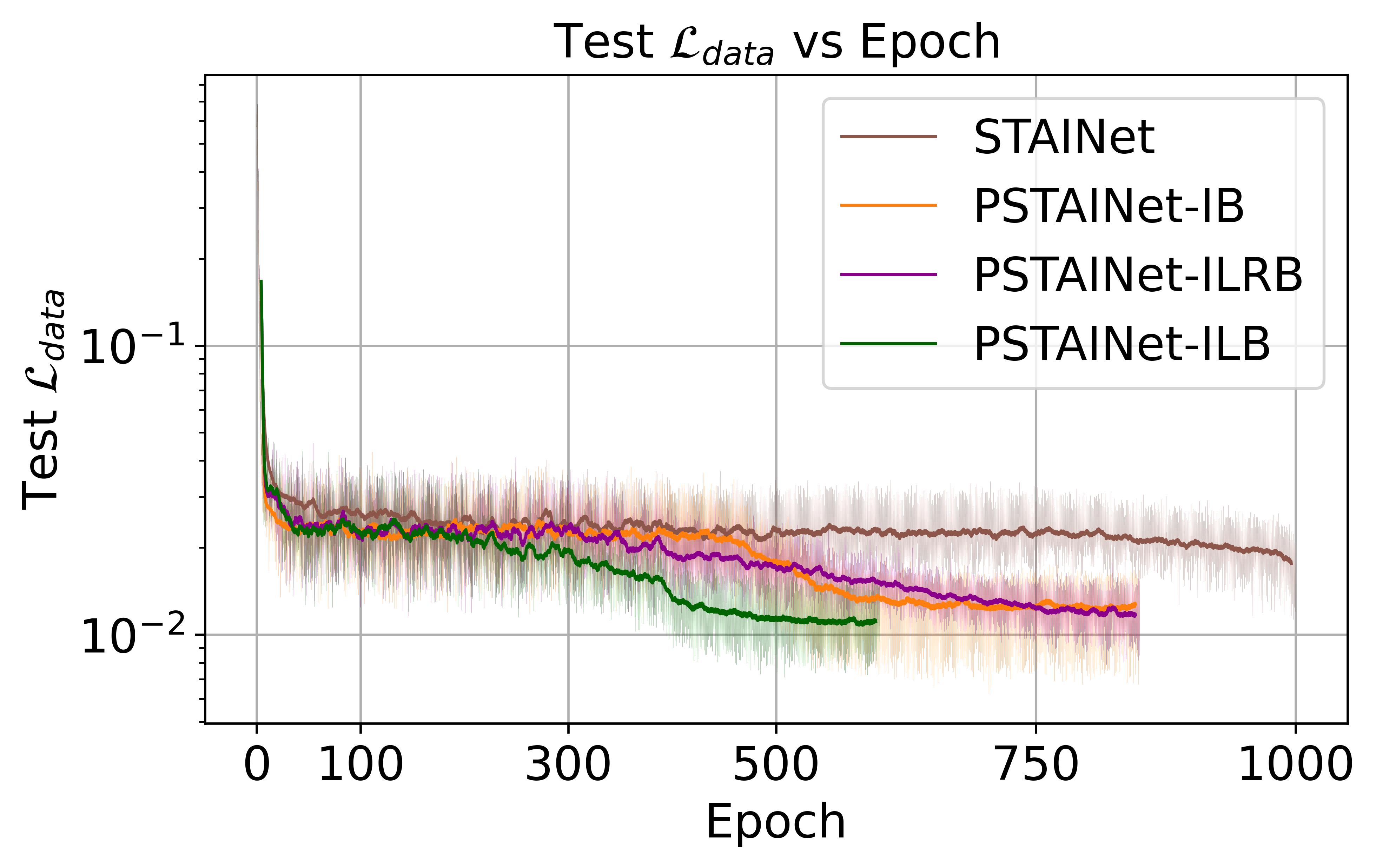}
        \caption{}
        \label{fig:test_loss}
    \end{subfigure}
    \begin{subfigure}[b]{0.49\textwidth}
        \centering
        \includegraphics[width=\textwidth]{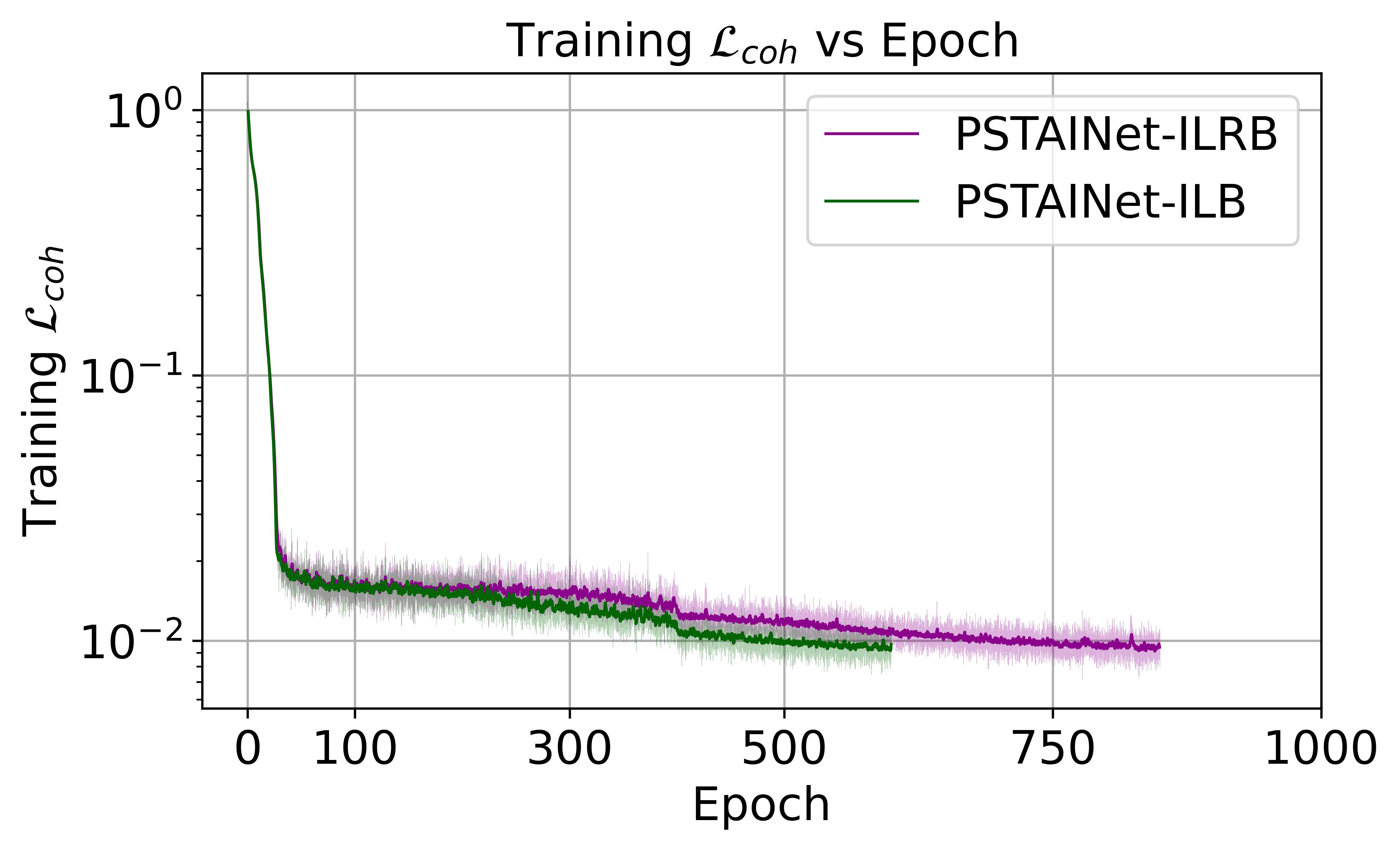}
        \caption{}
        \label{fig:coh_loss}
    \end{subfigure}
    \begin{subfigure}[b]{0.49\textwidth}
        \centering
        \includegraphics[width=\textwidth]{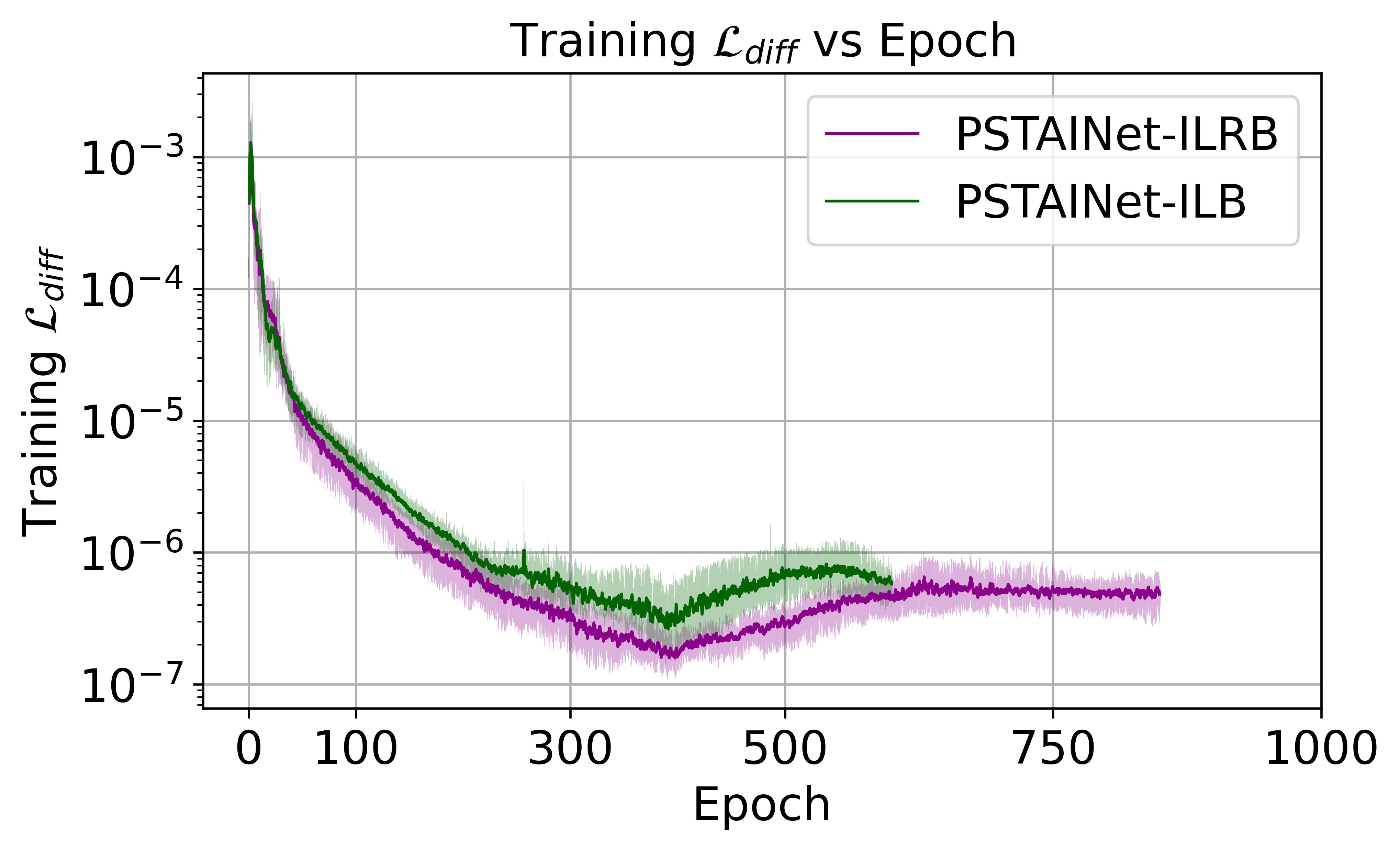}
        \caption{}
        \label{fig:diff_loss}
    \end{subfigure}
    \begin{subfigure}[b]{0.49\textwidth}
        \centering
        \includegraphics[width=\textwidth]{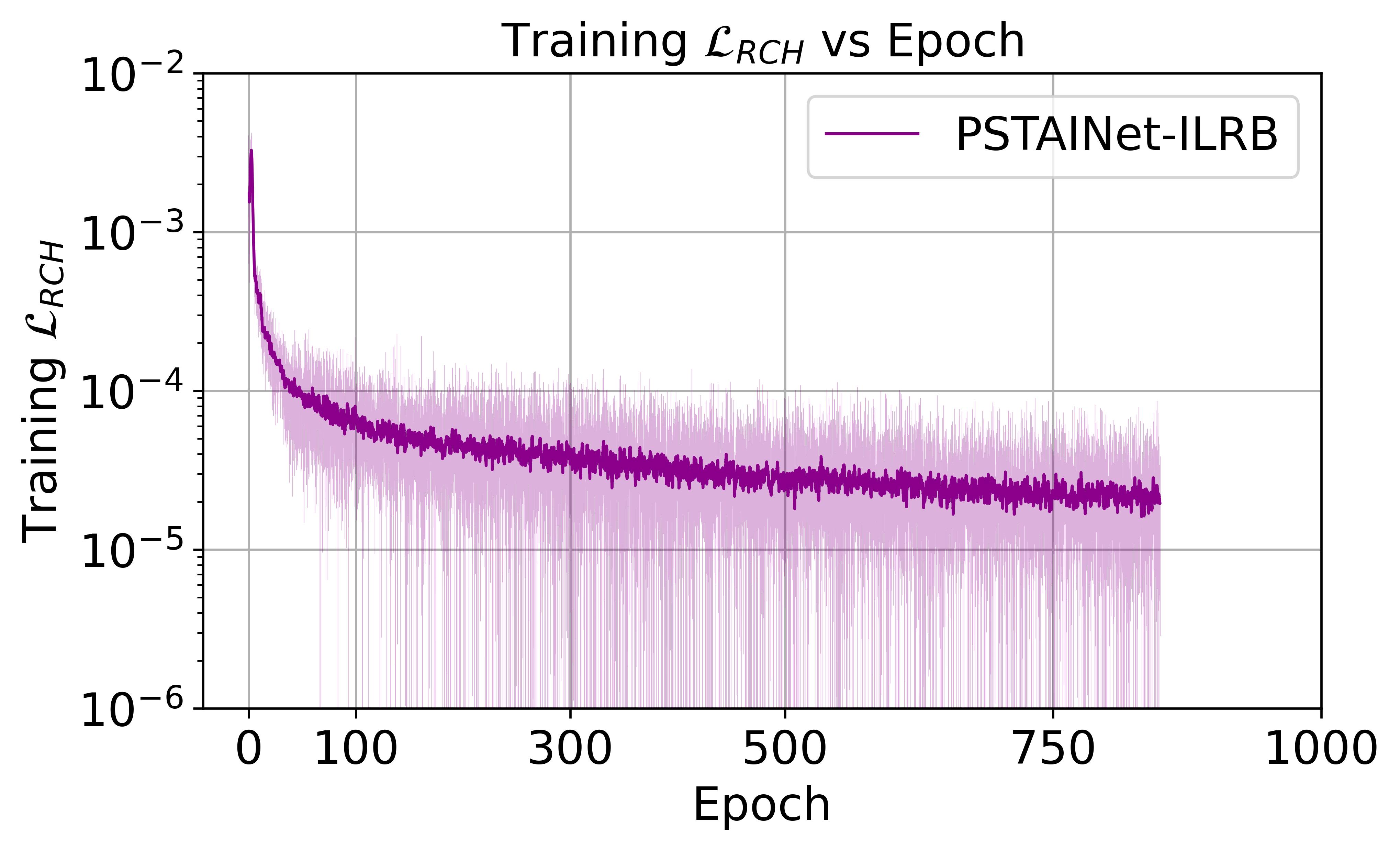}
        \caption{}
        \label{fig:rch_loss}
    \end{subfigure}
    \caption{Loss plots \textbf{a)} training $\mathcal{L}$, \textbf{b)} test $\mathcal{L}_{\text{\textit{data}}}$, \textbf{c)} training $\mathcal{L}_{\text{\textit{coh}}}$, \textbf{d)} training $\mathcal{L}_{\text{\textit{diff}}}$, \textbf{e)} training $\mathcal{L}_{RCH}$.}
    \label{fig:loss_plots}
\end{figure*}

Figures \ref{fig:train_loss} and \ref{fig:test_loss}, which report the values of the training total loss $\mathcal{L}$ (from 2001-01-01 to 2021-12-31) and test data loss (from 2022-01-01 to 2023-12-31), indicate that PSTAINet-ILB and PSTAINet-ILRB exhibit superior generalization ability.
Specifically, although both models obtained higher training $\mathcal{L}$ values, they achieved lower test errors, demonstrating better out-of-sample performance.
Furthermore, even if PSTAINet-ILB has been trained for fewer epochs (600), it achieved the best performance, providing additional evidence of the effectiveness of the adoption of both inductive and learning bias strategies.
Nevertheless, the introduction of learning biases substantially increased the time required for completing a training epoch, especially due to gradient computations, which were on average $4.616s$ for STAINet, $4.157s$ for PSTAINet-IB, $18.286s$ for PSTAINet-ILB, and $18.697s$ for PSTAINet-ILRB.

Figures \ref{fig:coh_loss}, \ref{fig:diff_loss}, and \ref{fig:rch_loss} report the values of $\mathcal{L}_{\text{\textit{coh}}}$, $\mathcal{L}_{\text{\textit{diff}}}$, and $\mathcal{L}_{\text{RCH}}$ on the training set, respectively.
Both $\mathcal{L}_{\text{\textit{coh}}}$ and $\mathcal{L}_{\text{RCH}}$ exhibit a monotonic decrease over the course of training, whereas $\mathcal{L}_{\text{\textit{diff}}}$ shows an initial decline followed by a subsequent rebound for both models.
This behaviour may stem from interactions or conflicts among the different loss terms, but it may also be attributed to noise, given the very small magnitudes reached by $\mathcal{L}_{\text{\textit{diff}}}$ (see the y-axis in Figure~\ref{fig:diff_loss}).
A more in-depth analysis of these loss dynamics is left for future work.

\subsection{Error Analyses}
\subsubsection{Missing Data Analysis}
\begin{figure}[H]
    \centering
    \includegraphics[width=0.85\linewidth]{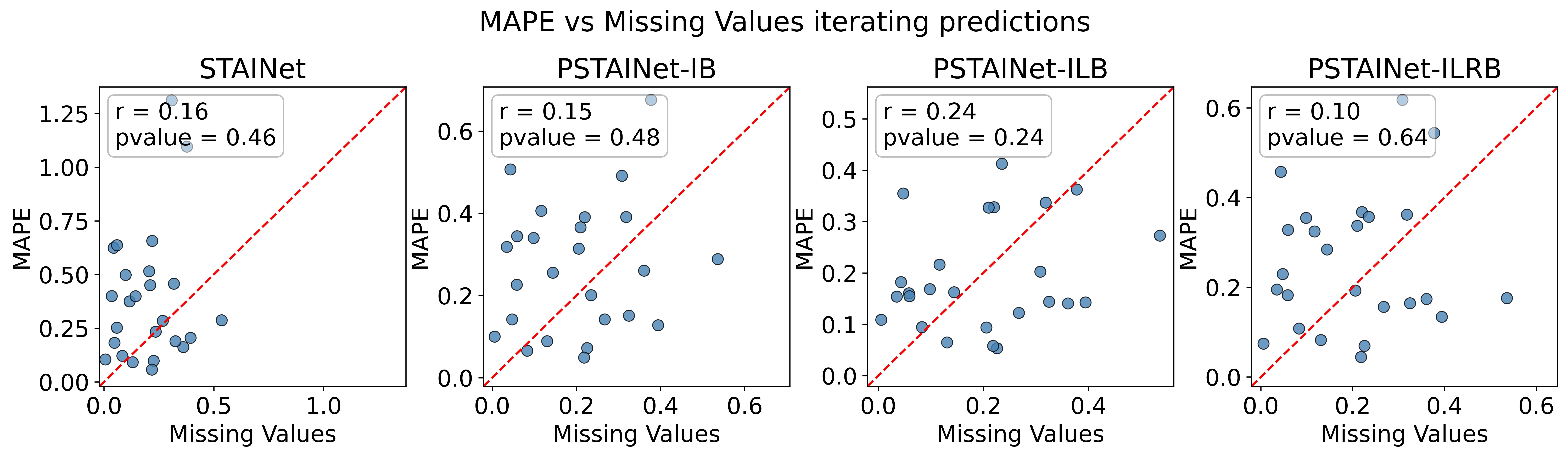}
    \caption{Correlation plots between MAPE and missing values percentage for STAINet, PSTAINet-IB, PSTAINet-ILB, and PSTAINet-ILRB; in all plots, the red line is the bisector, r is Pearson's correlation coefficient, and pvalue is the corresponding p-value.}
    \label{fig:mape_vs_nan}
\end{figure}

To analyze the effect of missing data on model performance, we studied the correlation between MAPE and missing values.
Figure~\ref{fig:mape_vs_nan} depicts the scatterplots and the Pearson correlation between MAPE and missing values for all sensors in the rollout setting \rev{and the corresponding p-value}.
A slight positive correlation is observed for all the models, with values ranging from $0.1$ for the PSTAINet-ILRB to $0.24$ for the PSTAINet-ILB. \rev{However, looking at the p-values, they are all not statistically different from $0$, suggesting that all models are relatively robust to missing data and can leverage the available information from the entire piezometer network.}
The points with the highest MAPE for the STAINet, PSTAINet-IB, and PSTAINet-ILRB correspond to the Buriasco and Orbassano sensors, which appear somewhat isolated in the scatterplot.
These high MAPEs may be related to the very large amount of missing data in recent years and to the pronounced decline observed in their measurements.

\subsubsection{Temporal Error Analysis}
\begin{figure}[H]
    \centering
    \includegraphics[width=0.85\linewidth]{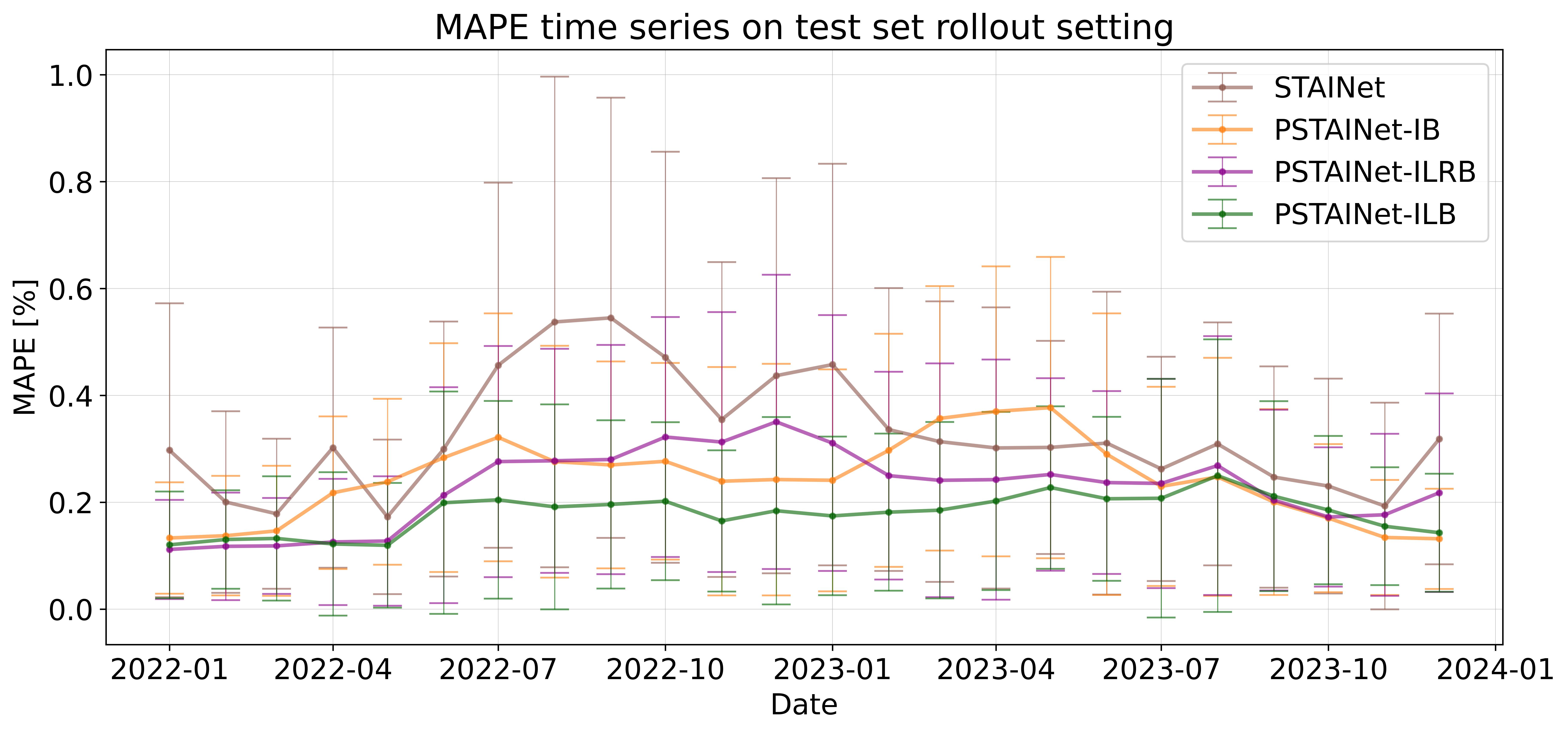}
    \caption{Monthly average MAPE over all sensors in the rollout setting on the test set. Vertical lines represent the corresponding monthly standard deviations.}
    \label{fig:ts_mape}
\end{figure}

In Figure \ref{fig:ts_mape} we reported the monthly MAPE averaged over all sensors for each date of the test set, along with the corresponding standard deviation. 
PSTAINet-ILB showed the lowest mean MAPE, with slightly higher errors in summer 2022 and summer 2023 -- the former affected by a severe drought. 
STAINet, and PSTAINet-ILRB produced larger errors in summer and autumn 2022 related to the drop, not well predicted, in groundwater levels in many locations (see Carmagnola in Figure~\ref{fig:carm1_pred_iter}, Cavour in Figure~\ref{fig:cavo_pred_iter}, Fossano in Figures~\ref{fig:foss1_pred_iter} and \ref{fig:foss2_pred_iter}, and Tarantasca in Figure~\ref{fig:tara_pred_iter}).

This behaviour is also observed, though less accentuated, for the PSTAINet-IB model, which produced slightly larger errors also during spring 2023.
Nevertheless, given the limited length of the test set, it is difficult to state seasonal error patterns for the models.
It would be necessary to have more observations from additional years as a test set -- a condition that will be met in the near future, once data from the 2024 and 2025 field measurement campaigns become available.

\subsubsection{Trend \& Seasonality Analysis}

\begin{figure}[H]
    \centering
    \includegraphics[width=0.85\linewidth]{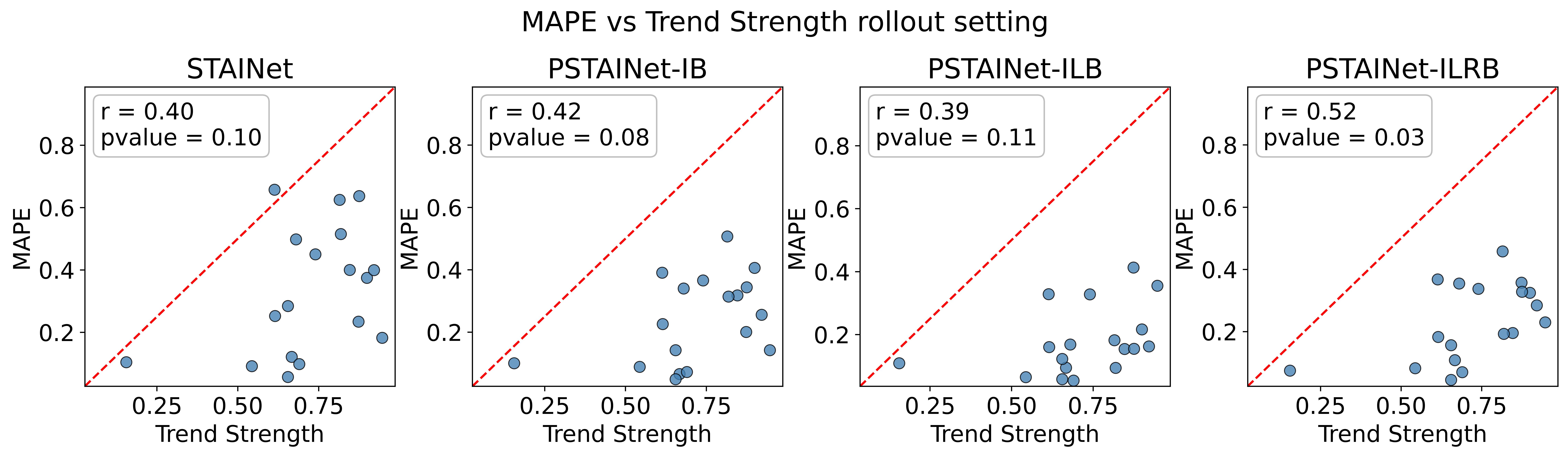}
    \caption{Correlation plots between MAPE and Trend Strength $\mathcal{S}_T$ for STAINet, PSTAINet-IB, PSTAINet-ILB, and PSTAINet-ILRB; in all plots, the red line is the bisector, r is Pearson's correlation coefficient, and pvalue is the corresponding p-value.}
    \label{fig:mape_trend}
\end{figure}

\begin{figure}[H]
    \centering
    \includegraphics[width=0.85\linewidth]{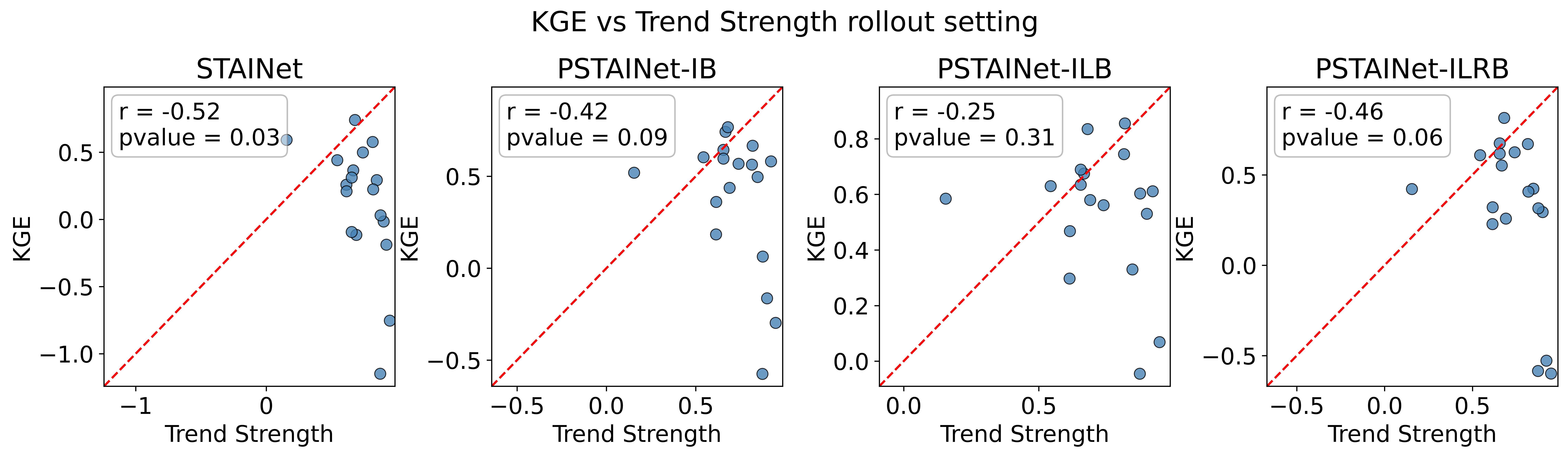}
    \caption{Correlation plots between KGE and Trend Strength $\mathcal{S}_T$ for STAINet, PSTAINet-IB, PSTAINet-ILB, and PSTAINet-ILRB; in all plots, the red line is the bisector, r is Pearson's correlation coefficient, and pvalue is the corresponding p-value.}
    \label{fig:kge_trend}
\end{figure}

\begin{figure}[H]
    \centering
    \includegraphics[width=0.85\linewidth]{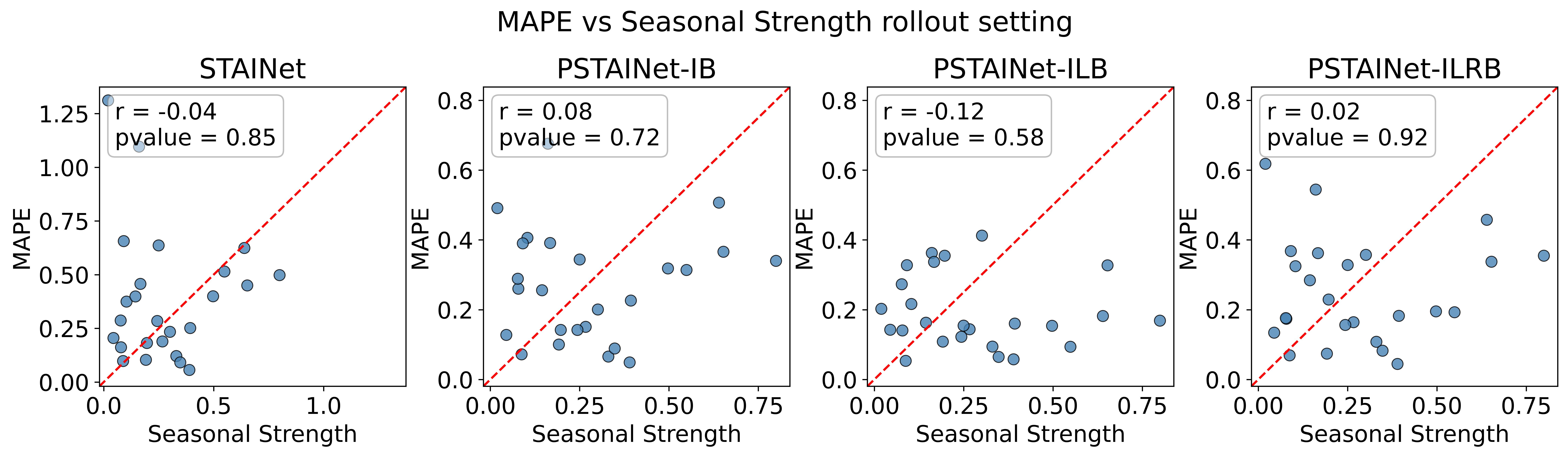}
    \caption{Correlation plots between MAPE and Seasonal Strength $\mathcal{S}_S$ for STAINet, PSTAINet-IB, PSTAINet-ILB, and PSTAINet-ILRB; in all plots, the red line is the bisector, r is Pearson's correlation coefficient, and pvalue is the corresponding p-value.}
    \label{fig:mape_seasonal}
\end{figure}

\begin{figure}[H]
    \centering
    \includegraphics[width=0.85\linewidth]{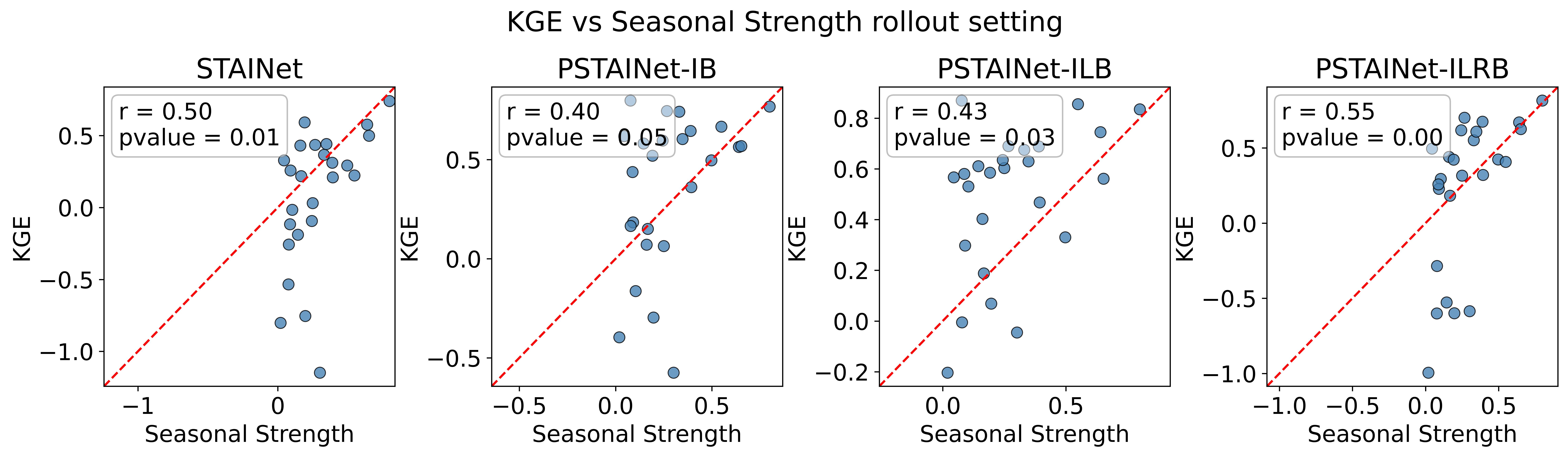}
    \caption{Correlation plots between KGE and Seasonal Strength $\mathcal{S}_S$ for STAINet, PSTAINet-IB, PSTAINet-ILB, and PSTAINet-ILRB; in all plots, the red line is the bisector, r is Pearson's correlation coefficient, and pvalue is the corresponding p-value.}
    \label{fig:kge_seasonal}
\end{figure}

Figures \ref{fig:mape_trend} and \ref{fig:kge_trend} report the scatterplots and the Pearson correlation between MAPE and $\mathcal{S}_T$, and KGE and $\mathcal{S}_T$, respectively. 
For all models, a positive correlation (ranging from $0.39$ to $0.52$) is present between MAPE and $\mathcal{S}_T$ while a negative one with KGE (ranging from $-0.52$ to $-0.25$), \rev{suggesting} that the stronger the trend of each time series, the lower the performance is.
Indeed, a stronger trend component might indicate a nonstationary time series, which is more difficult to model, especially if the nonstationarity is caused by external factors (e.g., drought and climate change).
\rev{However, looking at the p-values, correlations are statistically different from $0$ for a significance level of 0.10 for all models except PSTAINet-ILB, which delivers better performance disregarding the trend strength.
This may suggest that PSTAINet-ILB is actually able to capture exogenous phenomena and adapt its predictions accordingly, thereby maintaining good predictive performance.}
Conversely, as Figures \ref{fig:mape_seasonal} and \ref{fig:kge_seasonal} show, a positive correlation was observed between KGE and $\mathcal{S}_S$ (ranging from $0.40$ to $0.55$), while no correlation with MAPE (ranging from $-0.12$ to $0.08$).
This means that models do not show any definite relation between the absolute percentage errors and stronger seasonal patterns.
However, a stronger seasonality facilitates the prediction of the dynamical patterns of the time series (higher KGE) -- an expected association given that a stronger seasonality means a clearer repeated pattern through time.

\subsubsection{Spatial Error Analysis}
\begin{figure*}[!h]
    \centering
    \includegraphics[width=0.95\linewidth]{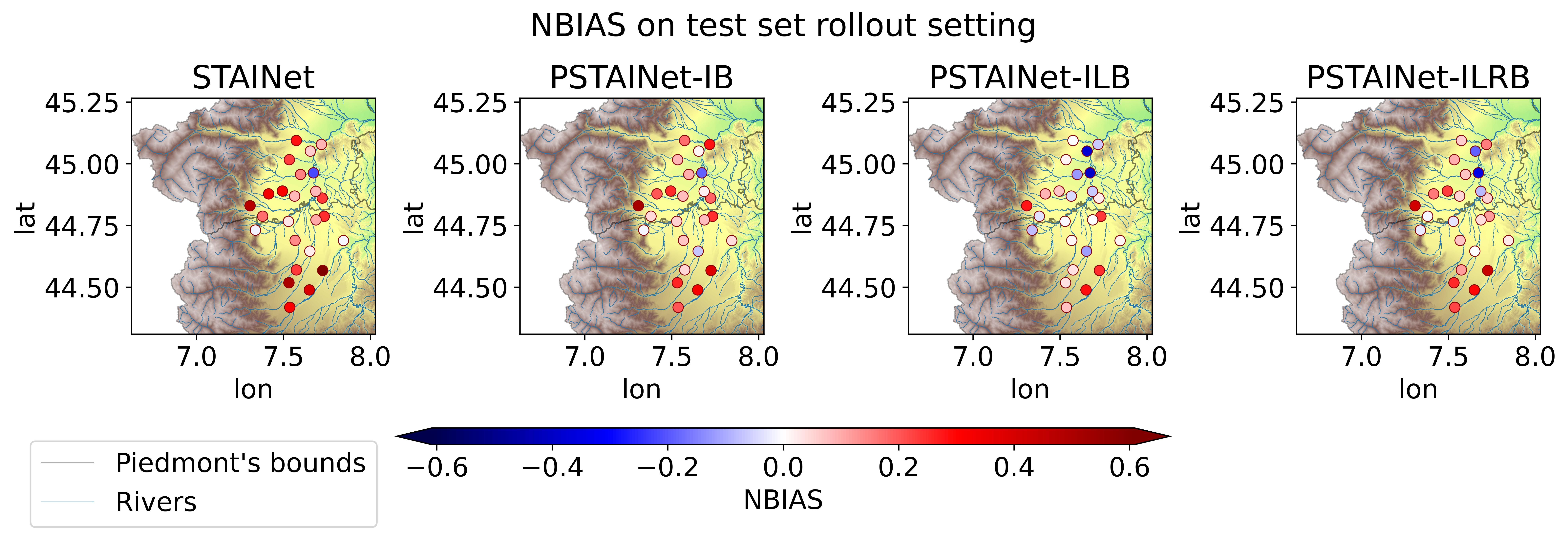}
    \caption{NBIAS for all sensors in the rollout setting on the test set.}
    \label{fig:map_nbias}
\end{figure*}

\begin{figure*}[!h]
    \centering
    \includegraphics[width=0.95\linewidth]{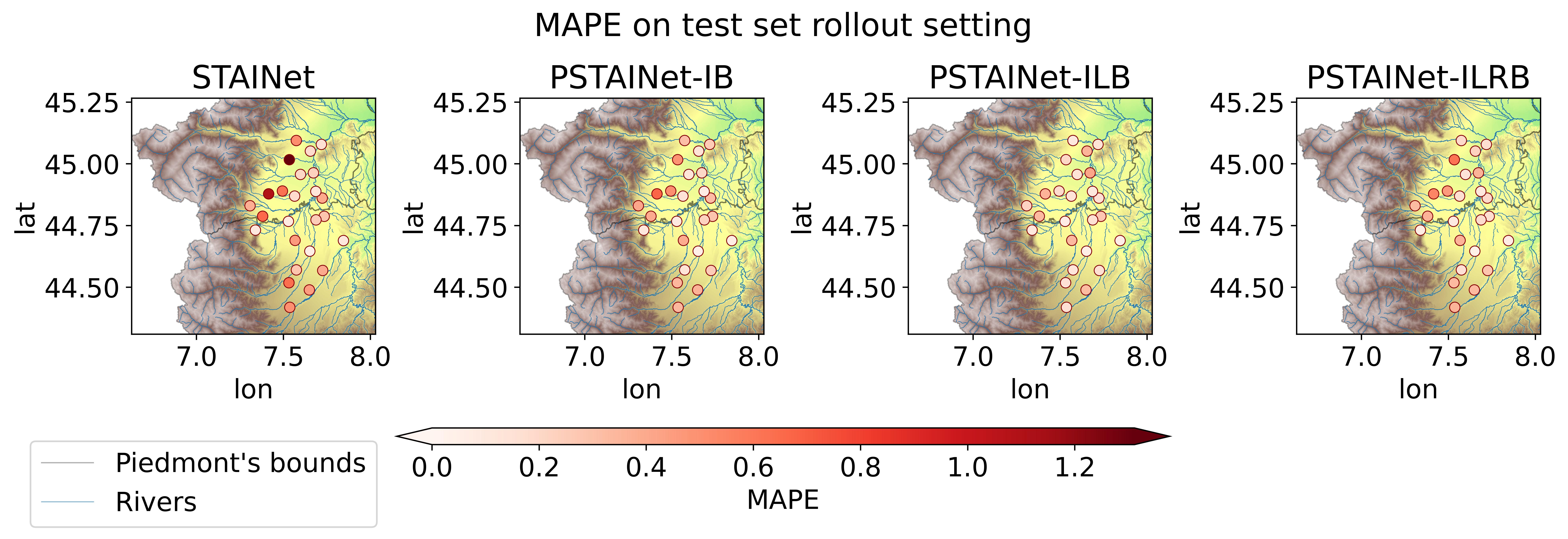}
    \caption{MAPE for all sensors in the rollout setting on the test set.}
    \label{fig:map_mape}
\end{figure*}

Figure~\ref{fig:map_nbias} and Figure~\ref{fig:map_mape} report the NBIAS and MAPE for all sensors in the rollout setting, respectively.
Overall, the models tended to overestimate groundwater levels, as indicated also by the positive median NBIAS values reported in Table~\ref{tab:PI_metrics}.
Among all models, the PSTAINet-ILB exhibited the smallest NBIAS, showing a more balanced spatial pattern -- with slight underestimations in the northern sensors and overestimations in the southern ones (Figure~\ref{fig:map_nbias}).
The sensor located in La Loggia represented a challenging case for all models, which was underestimated by all of them (as shown also in Figure~\ref{fig:lal_pred_iter}). 
This could be due to the high number of missing data in the last year, but also because this sensor is among the few that did not exhibit a pronounced decreasing trend in recent years (see Figure~\ref{fig:all_ts} and in particular Figure~\ref{fig:lal_ts}) -- most likely as a result of its proximity to the Po River.\\
Concerning the spatial distribution of the MAPE in Figure~\ref{fig:map_mape}, a mild tendency can be observed for higher MAPE values in the north-western and south-western sensors across all models (as for the NBIAS) apart from the PSTAINet-ILB, which displays a more even spatial pattern.

\rev{
\subsubsection{Incorporation of the Physics Knowledge}
As already reported in Section~\ref{sec:results}, the injection of physics knowledge, either through an inductive bias (PSTAINet-IB) or through learning biases (PSTAINet-ILB and PSTAINet-ILRB), consistently improved the predictive performance with respect to the baseline STAINet.}

\rev{
The inductive bias alone, obtained by restructuring the tail of the original STAINet architecture inspired by the groundwater flow equation, already produced a measurable improvement (reduction of MAPE by 18\% and increase of KGE by 65\%).
It is important to note that this gain should not be attributed to an increase in the number of layers or the size of the architecture, as reported in other studies~\cite{chen2020}.
In our case, the backbone of STAINet remained the same in all the models, and only the final layers were redesigned so that the model output is expressed as the sum of the theoretical components.
Moreover, this modification reduced the number of trainable parameters from 62081 to 49666.}

\begin{figure}[h]
    \centering
    
    \begin{subfigure}{\linewidth}
        \centering
        \includegraphics[width=0.75\textwidth]{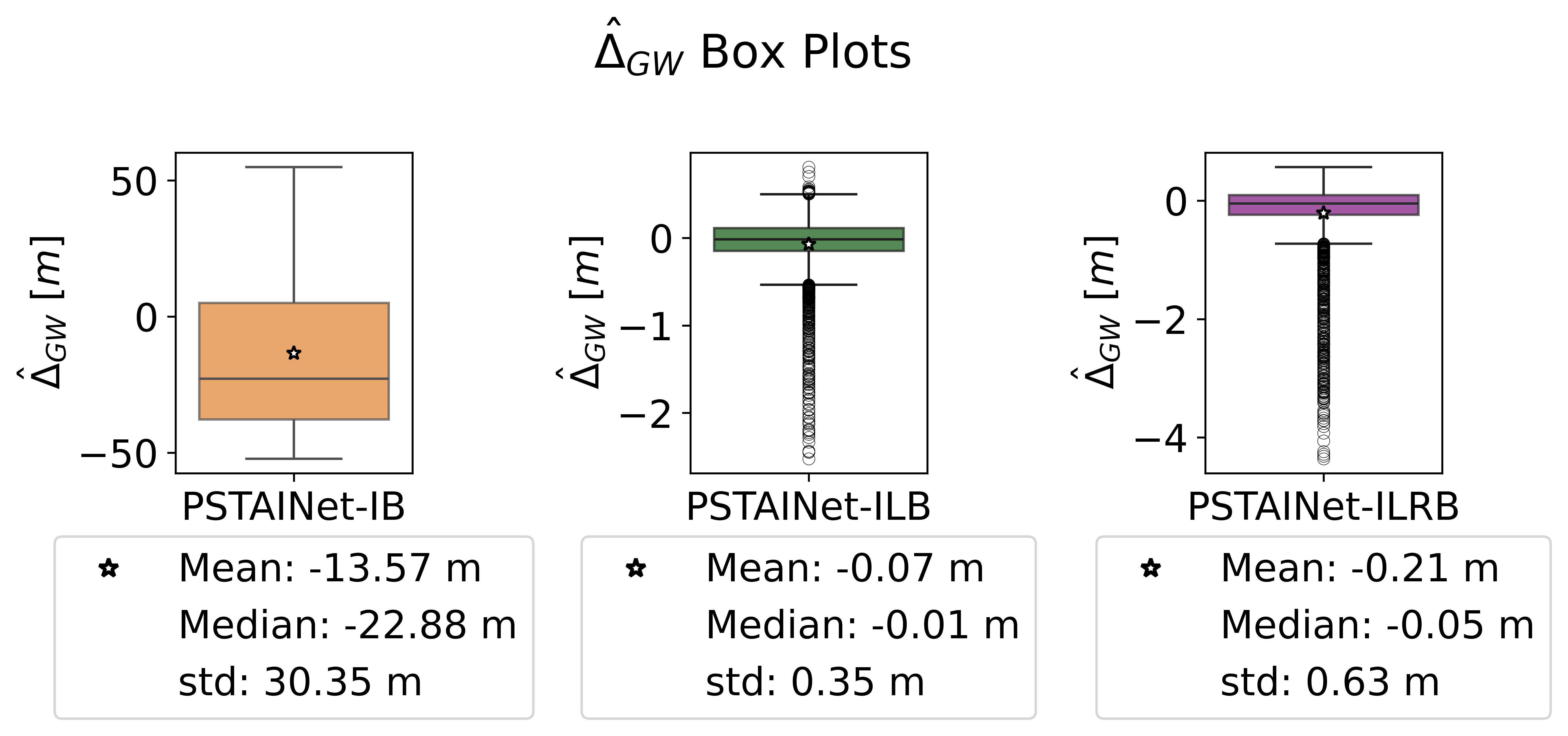}
        \caption{}
        \label{fig:deltawg_boxp}
    \end{subfigure}
    \begin{subfigure}{\linewidth}
        \centering
        \includegraphics[width=0.75\textwidth]{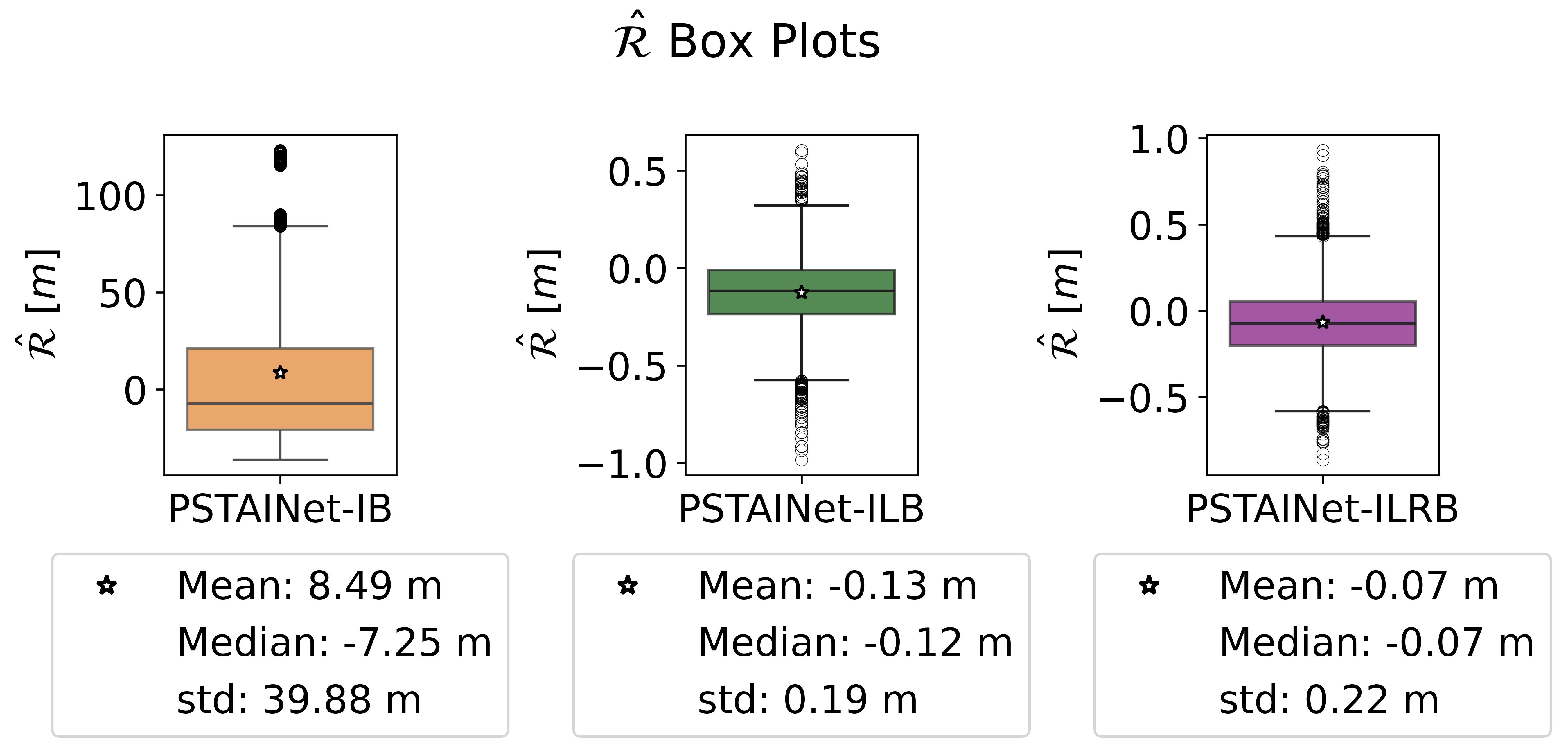}
        \caption{}
        \label{fig:R_boxp}
    \end{subfigure}
    \caption{Box plots representing predictions for $\hat{\Delta}_{GW_{t^*}}$ (\textbf{a)}) and $\hat{\mathcal{R}}_{t^*}$ (\textbf{b}) generated by PSTAINet-IB, PSTAINet-ILB, and PSTAINet-ILRB.
    The star in each box plot represents the mean, while the horizontal line inside the box is the median.}
    \label{fig:eqboxp}
\end{figure}

\rev{
However, the analysis of the estimated physics components revealed that the inductive bias alone (PSTAINet-IB) does not guarantee physically meaningful representations.
As shown in Figure~\ref{fig:eqboxp}, PSTAINet-IB produced $\hat{\Delta}_{GW_{t^*}}$ values spanning unrealistically large ranges, corresponding to implausible weekly groundwater variations at the considered monitoring stations‌.}

\rev{
Introducing learning biases (PSTAINet-ILB and PSTAINet-ILRB) substantially improved the plausibility of the estimated components while further increasing the predictive accuracy on the test set (MAPE improved by 48\% and 33\%, respectively).
Among the two approaches, PSTAINet-ILB outperformed PSTAINet-ILRB, likely because the constraint imposed on $\hat{\mathcal{R}}_{t^*}$ through the recharge zone prior was overly restrictive.
In addition, the recharge zone estimates are themselves affected by uncertainty, which may have limited the effectiveness of this constraint.}

\rev{
Notwithstanding, both PSTAINet-ILB and PSTAINet-ILRB generated $\hat{\Delta}_{GW_{t^*}}$ and $\hat{\mathcal{R}}_{t^*}$  with ranges of values more consistent with domain expectations.
In particular, the estimated $\hat{\Delta}_{GW_{t^*}}$ values are approximately centered around zero and exhibit relatively small magnitudes (on the order of centimeters), with a pronounced left tail corresponding to occasional large groundwater displacements which may occur during intense precipitation events.}

\rev{
Although the absence of ground-truth measurements for the individual equation components precludes a formal quantitative evaluation, the obtained results suggest that the learning biases improve the physical plausibility, in addition to the predictive performance, compared with STAINet and PSTAINet-IB.
Considering both predictive performance and physics consistency, PSTAINet-ILB was then selected as the definitive model.
Overall, these findings indicate that the most effective approach is to combine inductive and learning biases while avoiding additional constraints on the residual component.}

\rev{
It is worth emphasizing that the proposed approaches incorporate physics knowledge into a deep learning model without imposing any hard physics constraints on the generated output.
Such hybrid approaches can be particularly advantageous when dealing with complex environmental processes, where purely theory-based models may not fully capture complex, latent, and context-dependent relationships due to simplifying assumptions, but still provide an informative prior for the data-driven methods.
}

\vspace{1cm}

\section{Conclusion}
\label{sec:conclusion_fw}

In this work, we aimed to develop deep learning models \rev{to predict the weekly groundwater level, evaluating pure data-driven and physics-guided deep learning approaches.}
\rev{Specifically, }we investigated different bias strategies, namely inductive and learning biases, to incorporate physics knowledge derived from the groundwater flow equation into the models.

To this end, we first developed a purely deep learning model, STAINet, based on the Multi-Head Attention mechanism. 
We then proposed a restructured variant, PSTAINet-IB, using inductive bias by explicitly predicting the autoregressive ($h_{\mathcal{P},t^*-1}$), diffusion ($\Delta_{GW_{t^*}}$), and sink/source ($\mathcal{R}_{t^*}$) components of the governing equation.
Furthermore, we derived the PSTAINet-ILB model by introducing regularization terms into the loss function to explicitly supervise these components (learning bias), and the PSTAINet-ILRB model by further integrating prior information on recharge zones.

The PSTAINet-ILB model, trained adopting both inductive and learning bias strategies, achieved the best overall performance, \rev{decreasing the MAPE by 48\% and increasing the KGE by 129\% with respect to the pure data-driven baseline STAINet.} 
Although PSTAINet-ILRB also performed well, it achieved slightly lower metrics, probably because the recharge-zone prior was ‌too restrictive and uncertain‌.

The predicted diffusion and residual components provide meaningful insights for assessing the physics soundness of the models. 
In particular, while PSTAINet-IB produced implausible $\hat{\Delta}_{GW_{t^*}}$ and $\hat{\mathcal{R}}_{t^*}$ predictions, both PSTAINet-ILB and PSTAINet-ILRB produced \rev{more} physically consistent predictions, with PSTAINet-ILB especially capturing dynamics evolving according to the recharge seasons.

The PSTAINet-ILB model demonstrated similar performance whether leveraging true data as lag inputs or iterating its own predictions over the entire test set (years 2022–2023), thus enabling long-term forecasts all over the ROI.
This was further confirmed in the time series reconstruction task, where the complete time series for all sensors were predicted using a six-month forecast horizon.
In this task as well, the PSTAINet-ILB model achieved remarkable results.

In conclusion, the adoption of inductive and learning biases has led to improved performance and enhanced the interpretability of deep learning models, despite several limitations remaining.
With respect to the inductive bias, a limitation is the requirement of a closed-form equation that has to be explicitly hard-coded into the model -- a condition that cannot always be satisfied in practice, especially in cases of partial or incomplete knowledge.
Differently, under the learning-bias strategy, the governing equation constraint must be formulated as a loss term, and this can substantially increase the computational cost (particularly due to gradient evaluation) when dealing with deep and complex models.
Furthermore, the number of control points, namely the locations at which the learning-bias loss terms are evaluated, has a significant impact on both computational cost and the effectiveness of prior injection.
On the one hand, increasing the number of control points strengthens adherence to the governing equation; on the other hand, it raises computational demands.
This trade-off must therefore be carefully considered and evaluated on a case-by-case basis.
Ultimately, even if the weight of the learning bias loss terms could be tuned, a strict compliance of the model's output with the governing equation cannot be guaranteed.
For this reason, adopting both inductive and learning bias strategies jointly may be beneficial to inject prior domain knowledge more rigorously and robustly.

The proposed models define a general pipeline applicable to a wide range of environmental applications, where in situ measurements and spatially distributed data (like sequence images) are jointly leveraged to predict a target variable, and where a governing equation, even if with some unknown parameters, is desirable to be incorporated. 
Physics-guided machine learning thus enables the development of hybrid hydrological and, more generally, Earth system models that combine the strengths of theory-driven and data-driven approaches, improving computational efficiency while retaining the flexibility to adapt to observed data. In this way, such models provide more effective tools for representing and predicting complex Earth system dynamics.

\section{Future Works}
Even though we obtained remarkable results, several research directions can be undertaken.
First, given the widespread presence of missing and irregular data across many disciplines, a more extensive evaluation of alternative strategies for handling missing data would be valuable.
For instance, attention-based mechanisms could be exploited to design masking strategies that avoid explicit imputation.
In addition, alternative neural architectures could be explored to capture better complex and context-dependent relationships, such as Mixture-of-Experts~\cite{fedus_2022}.

In developing the proposed physics-guided strategies, a simple yet effective Euler integration scheme was adopted; however, future work could investigate Neural Differential Equation frameworks to assess more advanced numerical integration methods~\cite{chen_2018,song_2024}.
While such approaches would increase computational demands, they could enable models to better learn the underlying dynamics and provide a more rigorous mathematical foundation for incorporating additional physics constraints (e.g., mass conservation).

Finally, future studies could benefit from the inclusion of additional data sources, including newly acquired groundwater level observations from the 2024 and 2025 field campaigns, as well as other spatially distributed datasets such as land-cover maps~\cite{buchhorn2020copernicus} or soil moisture products~\cite{smap}. These data could offer more informative exogenous forcings, particularly for those related to anthropogenic activities (e.g., irrigation).

\roles{\textbf{Matteo Salis}: Writing - original draft, review \& editing, Investigation, Conceptualization, Visualization, Validation, Software, Methodology, Data curation, Resources. 
\textbf{Gabriele Sartor}: Supervision, Software, Investigation, Conceptualization, Resources.
\textbf{Rosa Meo}: Supervision, Writing - review.
\textbf{Stefano Ferraris}: Supervision, Conceptualization, , Writing - review.
\textbf{Abdourrahmane M. Atto}: Supervision, Investigation, Conceptualization, Writing - review.}

\ack{We acknowledge ISCRA for awarding our \textit{Deep Learning for Spatio Temporal Phenomena, application to environmental data} (DL4STP) project access to the LEONARDO supercomputer, owned by the EuroHPC Joint Undertaking, hosted by CINECA (Italy).\\
The authors declare that they have no conflict of interest to disclose.\\
The authors disclose the use of LLM-based tools for spell-checking and improving fluency.
}

\data{All adopted data are freely accessible, in detail, the groundwater data at the ARPA website \url{https://shorturl.at/F3KzQ}, and ERA5-land at \hyperlink{https://cds.climate.copernicus.eu/datasets/reanalysis-era5-land?tab=overview}{10.24381/cds.e2161bac}.
The code is released in a GitHub repository at \url{https://github.com/Matteo-Salis/physics-guided-gwl} with additional animations (GIFs) of the predictions over the test set.
}

\bibliographystyle{elsarticle-num} 
\bibliography{bibliography}

\setcounter{page}{1}

\appendix
\title{Supplementary Materials}

\renewcommand{\thefigure}{S\arabic{figure}}
\setcounter{figure}{0}
\renewcommand{\thetable}{S\arabic{table}}
\setcounter{table}{0}

\section{Sensors Statistics and Time Series}
\label{apx:sensor_stat_ts}

In Table~\ref{tab:all_ts_stats}, we reported summary statistics for every sensor considered in the study.
Figure~\ref{fig:miss_perc} and Figure~\ref{fig:num_obs} represent the percentage of missing data and the number of observations in the training, validation, and test sets, respectively.
Figure~\ref{fig:all_ts} shows the time series plots of the other twenty-four sensors in addition to the ones shown in Figure~\ref{fig:focus_ts}.

\section{Test Set Predictions}
\label{sec:appx_test_set_metr_pred}

In Tables~\ref{tab:PI_sens_metrics_STNet}, \ref{tab:PI_sens_metrics_STDisNet}, \ref{tab:PI_sens_metrics_STDisNetPI}, and \ref{tab:PI_sens_metrics_STDisNetPI-RCH}, we reported the evaluation metrics computed on every sensor and every model in the iterative prediction (rollout) scenario. Furthermore, Figure~\ref{fig:all_pred_iter} shows models' rollout predictions in the test set for the other sensors in addition to the ones shown in Figure~\ref{fig:focus_pred}.

\section{Time Series Reconstruction}
\label{sec:appx_ts_rec}

In Table~\ref{tab:rec_metrics_all}, we reported the evaluation metrics for the reconstruction task of the PSTAINet-ILB model described in Section~\ref{sec:rec_task}. Figure~\ref{fig:all_rec} depicts the weekly prediction from 01-01-2001 for the other sensors in addition to the ones shown in Figure~\ref{fig:focus_rec}.

\vspace{2cm}
\begin{table*}[h!]
\centering
\caption{Summary statistics and percentage of missing data for each sensor.}
\label{tab:sensor_stats}
\begin{adjustbox}{width=0.9\textwidth}
\begin{tabular}{l|ccc}
\toprule
\textbf{Sensor (Municipality - Sensor ID)} & \textbf{Mean [m]} & \textbf{Std. Dev. [m]} & \textbf{Missing (\%)} \\
\cmidrule(lr){1-4}
\morecmidrules
\cmidrule(lr){1-4}
\textbf{Bricherasio - 00103510001} & 348.223 & 0.746 & 11.667 \\
\textbf{Buriasco - 00104110001} & 274.183 & 1.795 & 37.750 \\
\textbf{Candiolo - 00105110001} & 232.812 & 0.513 & 39.417 \\
\textbf{Carmagnola - 00105910001} & 231.893 & 0.813 & 3.500 \\
\textbf{Carmagnola - 00105910002} & 223.768 & 0.403 & 8.333 \\
\textbf{Cavour - 00107010001} & 274.777 & 1.842 & 22.000 \\
\textbf{Collegno - 00109010001} & 265.223 & 1.114 & 20.583 \\
\textbf{La Loggia - 00112710001} & 219.358 & 0.441 & 23.500 \\
\textbf{Orbassano - 00117110001} & 257.017 & 1.788 & 30.833 \\
\textbf{Poirino - 00119710001} & 237.157 & 0.544 & 76.917 \\
\textbf{Scalenghe - 00126010001} & 256.419 & 1.061 & 4.333 \\
\textbf{Torino - 00127210001} & 227.730 & 0.564 & 4.750 \\
\textbf{Torino - 00127210003} & 208.770 & 0.359 & 36.083 \\
\textbf{Virle - Piemonte 00131010001} & 241.877 & 0.585 & 32.500 \\
\textbf{Barge - 00401210001} & 321.579 & 0.488 & 0.583 \\
\textbf{Bra - 00402910001} & 275.202 & 0.341 & 22.583 \\
\textbf{Busca - 00403410001} & 434.676 & 1.862 & 24.917 \\
\textbf{Caramagna - Piemonte 00404110001} & 251.498 & 0.488 & 53.583 \\
\textbf{Cavallermaggiore - 00405910001} & 282.774 & 0.501 & 31.167 \\
\textbf{Cuneo - 00407810001} & 476.663 & 1.856 & 9.833 \\
\textbf{Fossano - 00408910001} & 351.209 & 0.485 & 14.417 \\
\textbf{Fossano - 00408910002} & 399.367 & 1.212 & 21.000 \\
\textbf{Moretta - 00414310002} & 249.211 & 0.348 & 13.083 \\
\textbf{Racconigi - 00417910001} & 252.498 & 0.744 & 5.833 \\
\textbf{Savigliano - 00421510001} & 312.866 & 0.266 & 21.833 \\
\textbf{Scarnafigi - 00421710001} & 282.321 & 1.169 & 31.833 \\
\textbf{Tarantasca - 00422510001} & 427.368 & 1.441 & 5.917 \\
\textbf{Vottignasco - 00425010001} & 380.387 & 0.834 & 26.750 \\
\bottomrule
\end{tabular}
\end{adjustbox}
\label{tab:all_ts_stats}
\end{table*}

\begin{figure}[H]
    \centering
    \includegraphics[width=0.9\linewidth]{missing_perc.jpeg}
    \caption{Bar plot representing the percentage of missing data for each sensor considered in the study. In the x-axis, sensors are labeled using both the municipality where they are placed and the identification code (sensor ID).}
    \label{fig:miss_perc}
\end{figure}

\begin{figure}[H]
    \centering
    \includegraphics[width=0.9\linewidth]{number_observations.jpeg}
    \caption{Bar plot representing the number of observations for each sensor in the training, validation, and test sets (also reported by the vertical numbers for each bar, respectively). In the x-axis, sensors are labeled using both the municipality where they are placed and the identification code (sensor ID).}
    \label{fig:num_obs}
\end{figure}

\begin{landscape}
\begin{figure}[h]
\vspace{-2cm}
\centering
    \begin{subfigure}{0.49\linewidth}
        \centering
        \includegraphics[width=\linewidth]{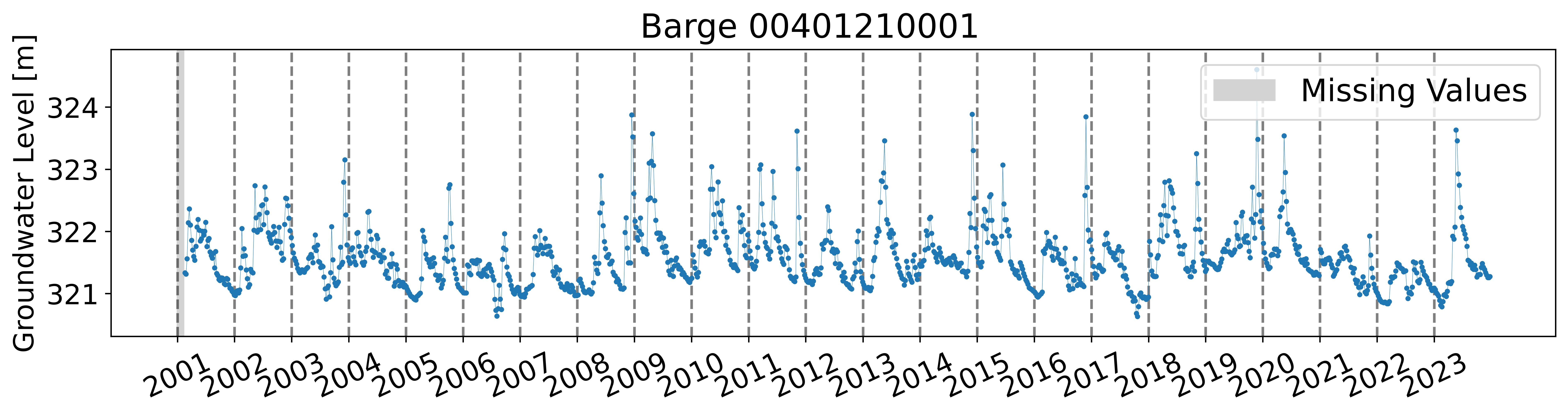}
        \caption{}
        \label{fig:bar_ts}
    \end{subfigure}
    \hfill
    \begin{subfigure}{0.49\linewidth}
        \centering
        \includegraphics[width=\linewidth]{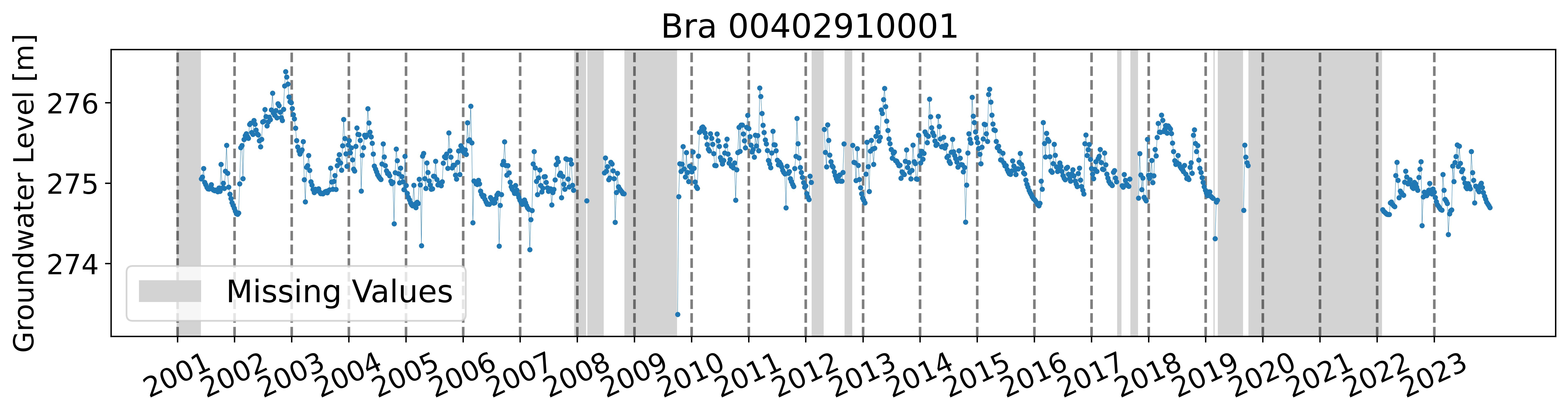}
        \caption{}
        \label{fig:bra_ts}
    \end{subfigure}
    \hfill
    \begin{subfigure}{0.49\linewidth}
        \centering
        \includegraphics[width=\linewidth]{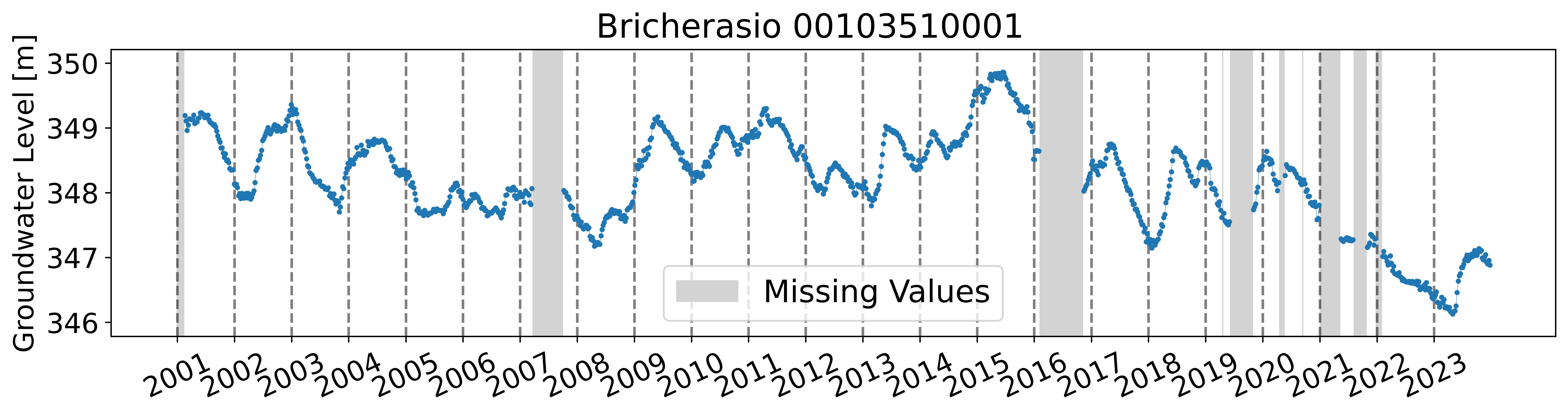}
        \caption{}
        \label{fig:bri_ts}
    \end{subfigure}
    \hfill
    \begin{subfigure}{0.49\linewidth}
        \centering
        \includegraphics[width=\linewidth]{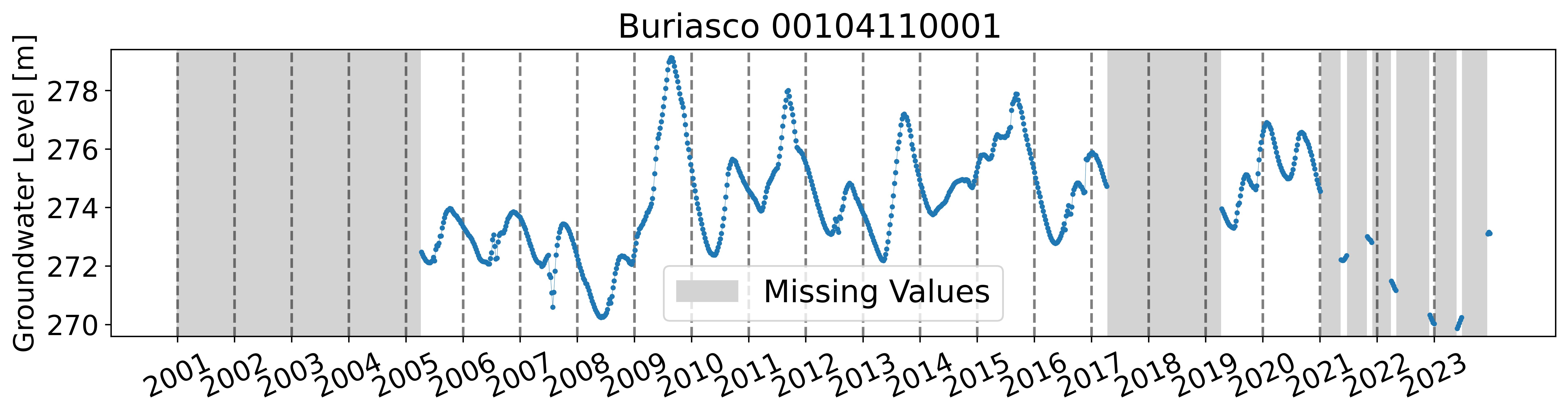}
        \caption{}
        \label{fig:bur_ts}
    \end{subfigure}
    \hfill
    \begin{subfigure}{0.49\linewidth}
        \centering
        \includegraphics[width=\linewidth]{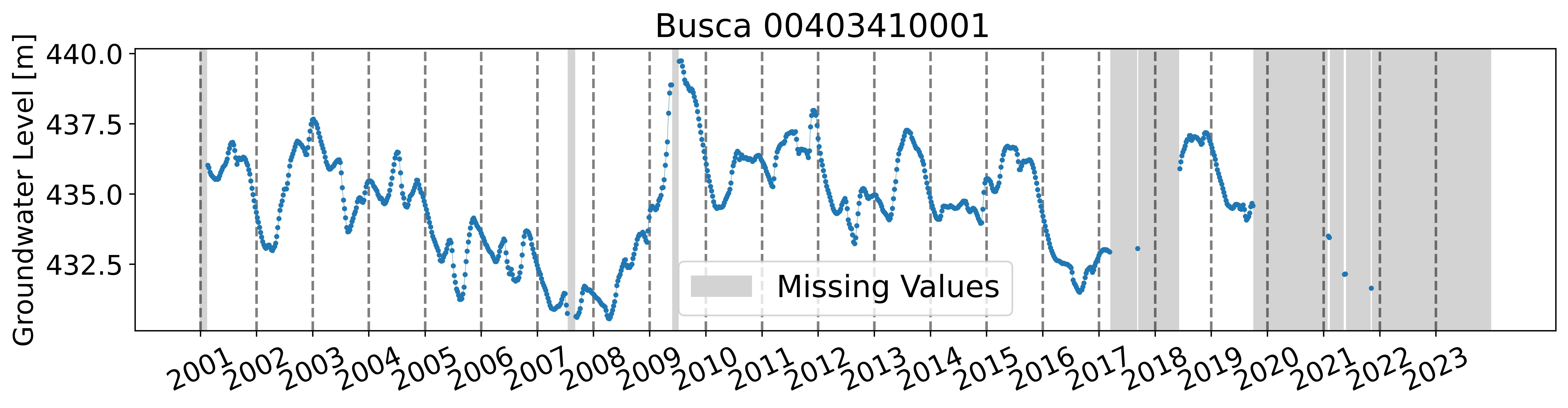}
        \caption{}
        \label{fig:busca_ts}
    \end{subfigure}
    \hfill
    \begin{subfigure}{0.49\linewidth}
        \centering
        \includegraphics[width=\linewidth]{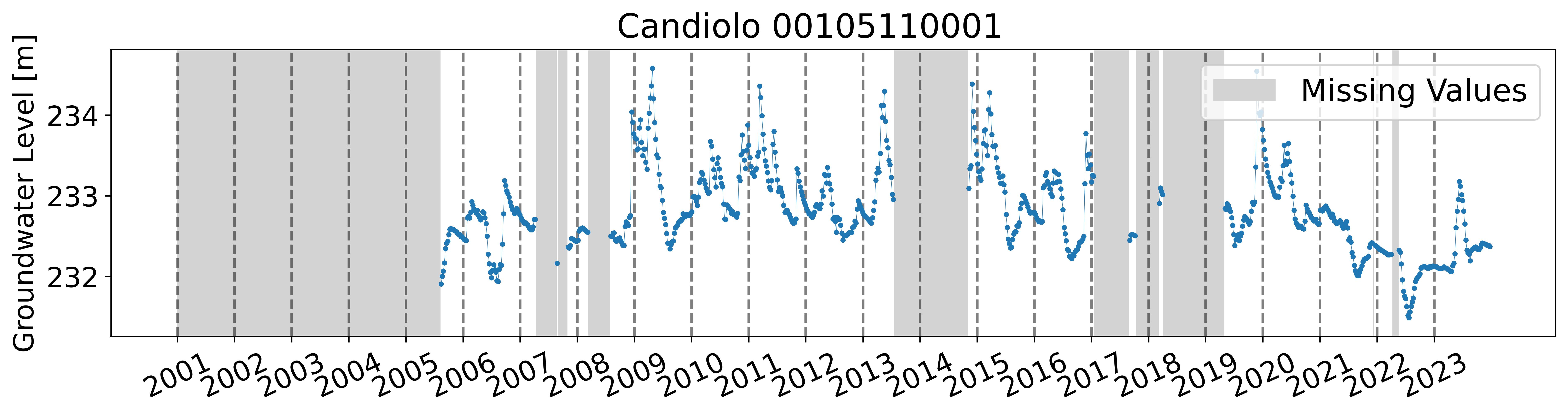}
        \caption{}
        \label{fig:cand_ts}
    \end{subfigure}
    \hfill
    \begin{subfigure}{0.49\linewidth}
        \centering
        \includegraphics[width=\linewidth]{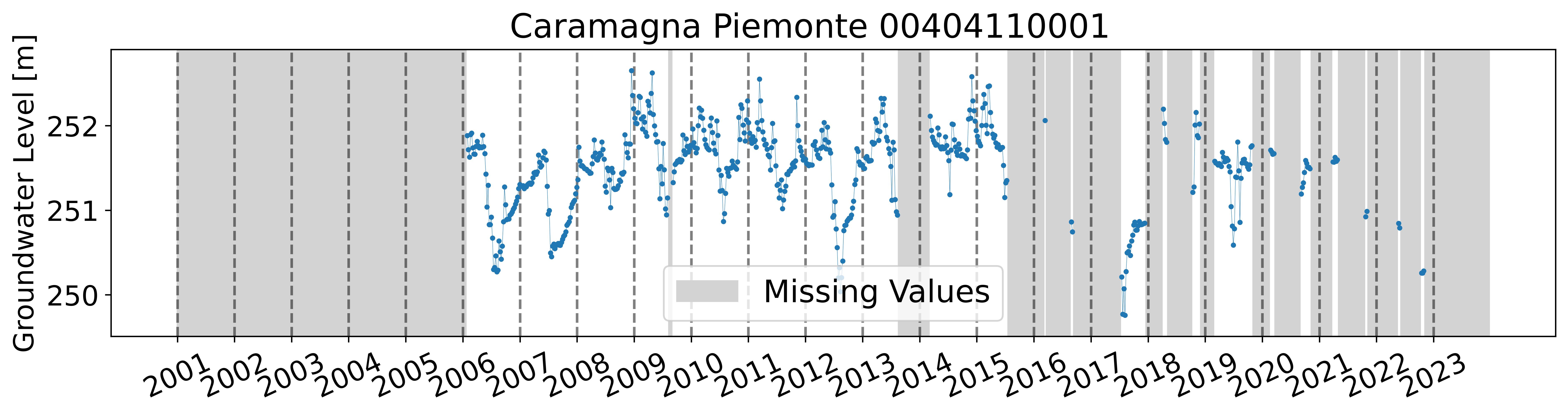}
        \caption{}
        \label{fig:cara_ts}
    \end{subfigure}
    \hfill
    \begin{subfigure}{0.49\linewidth}
        \centering
        \includegraphics[width=\linewidth]{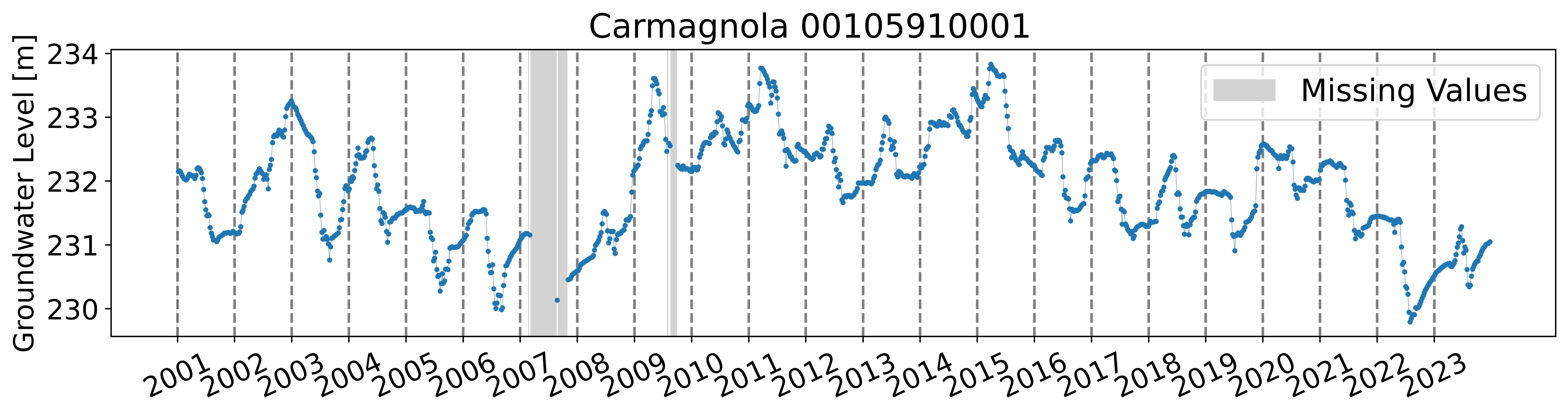}
        \caption{}
        \label{fig:carm1_ts}
    \end{subfigure}
\end{figure}
\end{landscape}

\begin{landscape}
\begin{figure}[ht]
\vspace{-2cm}
    \addtocounter{figure}{-1}  
    \begin{subfigure}{0.49\linewidth}
        \centering
        \addtocounter{subfigure}{8}
        \includegraphics[width=\linewidth]{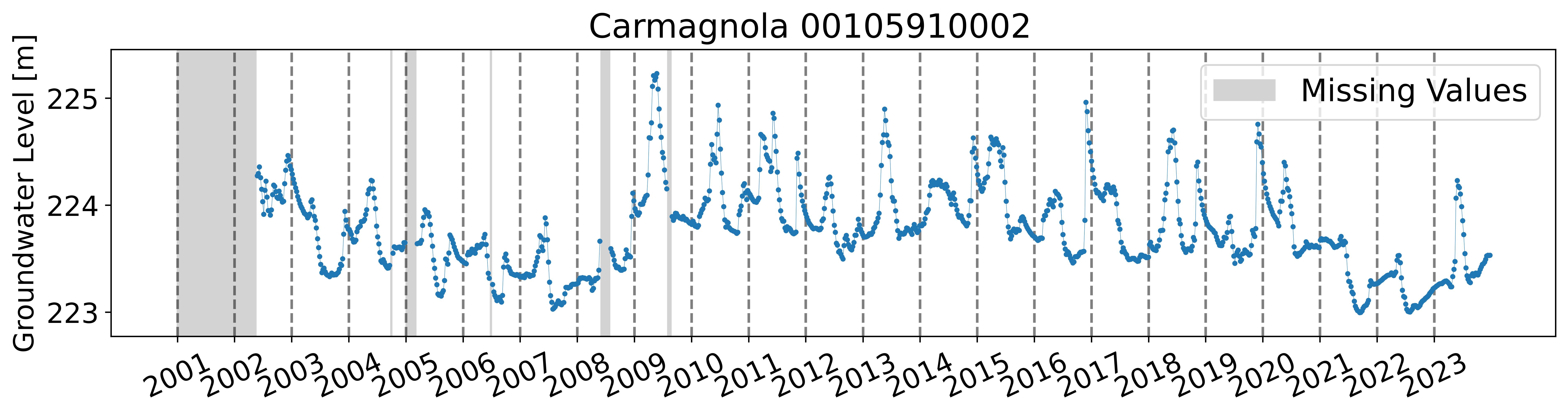}
        \caption{}
        \label{fig:carm2_ts}
    \end{subfigure}
    \hfill
    \begin{subfigure}{0.49\linewidth}
        \centering
        \includegraphics[width=\linewidth]{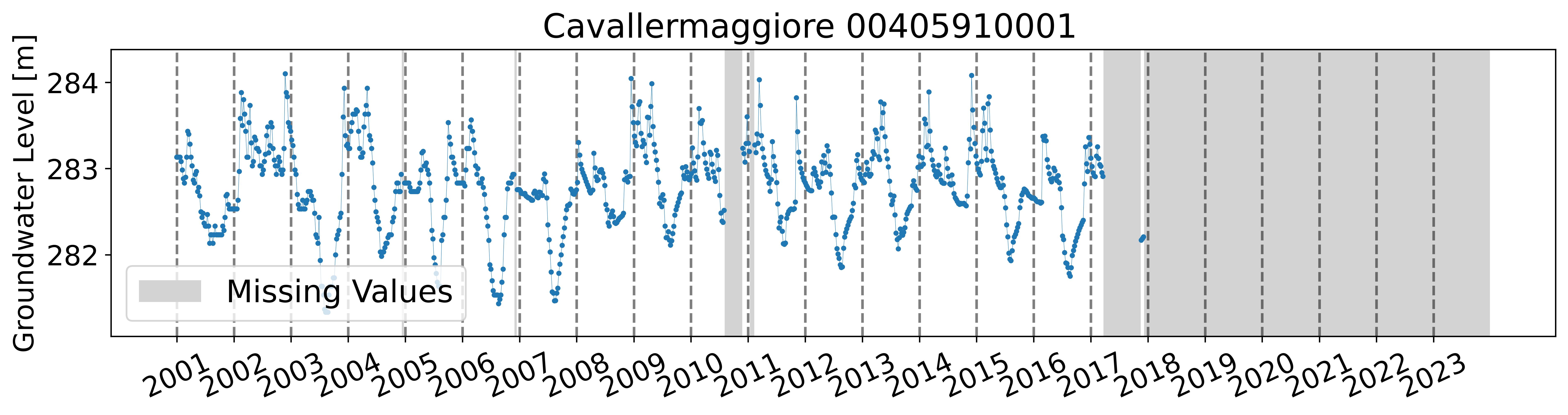}
        \caption{}
        \label{fig:cava_ts}
    \end{subfigure}
    \hfill
    \begin{subfigure}{0.49\linewidth}
        \centering
        \includegraphics[width=\linewidth]{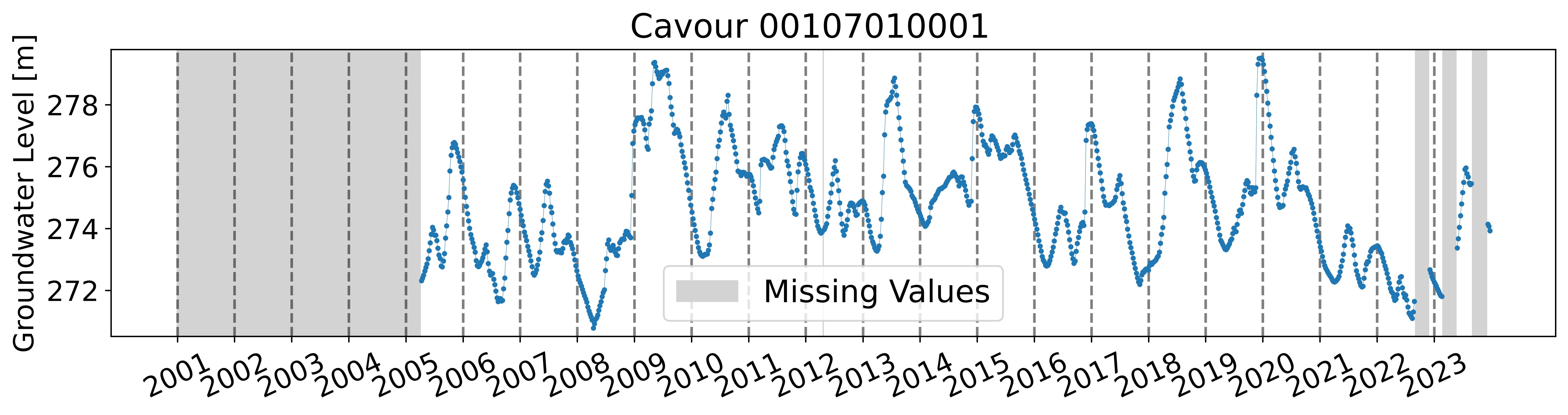}
        \caption{}
        \label{fig:cavo_ts}
    \end{subfigure}
    \hfill
    \begin{subfigure}{0.49\linewidth}
        \centering
        \includegraphics[width=\linewidth]{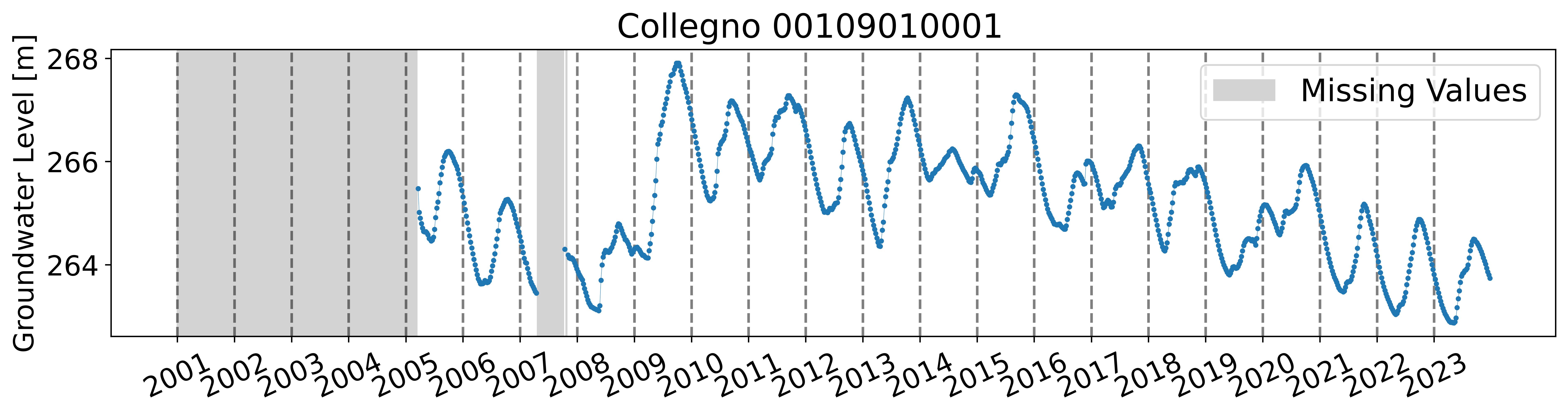}
        \caption{}
        \label{fig:coll_ts}
    \end{subfigure}
    \hfill
    \begin{subfigure}{0.49\linewidth}
        \centering
        \includegraphics[width=\linewidth]{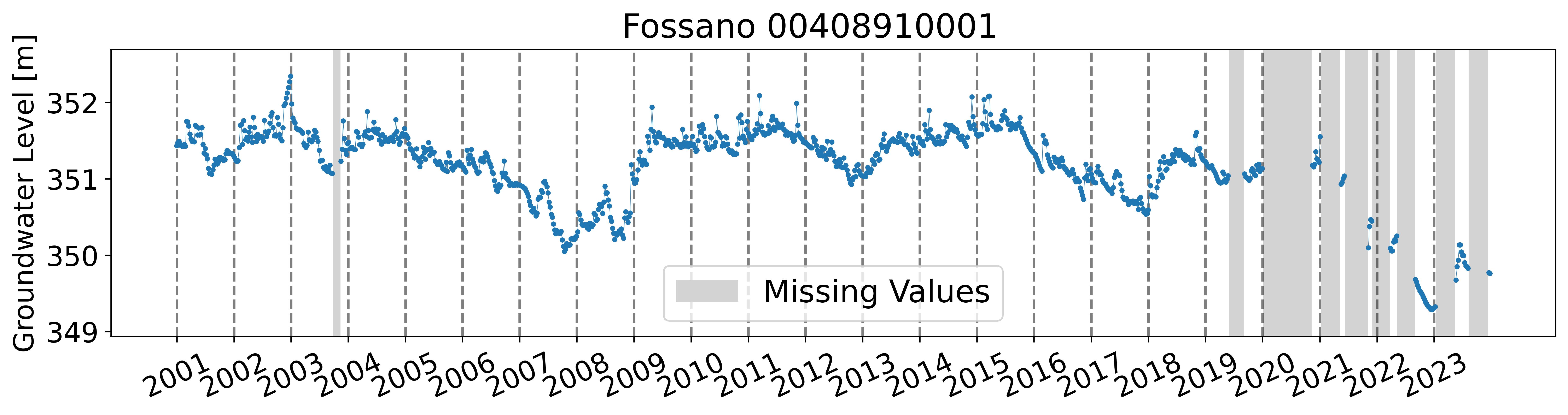}
        \caption{}
        \label{fig:foss1_ts}
    \end{subfigure}
    \hfill
    \begin{subfigure}{0.49\linewidth}
        \centering
        \includegraphics[width=\linewidth]{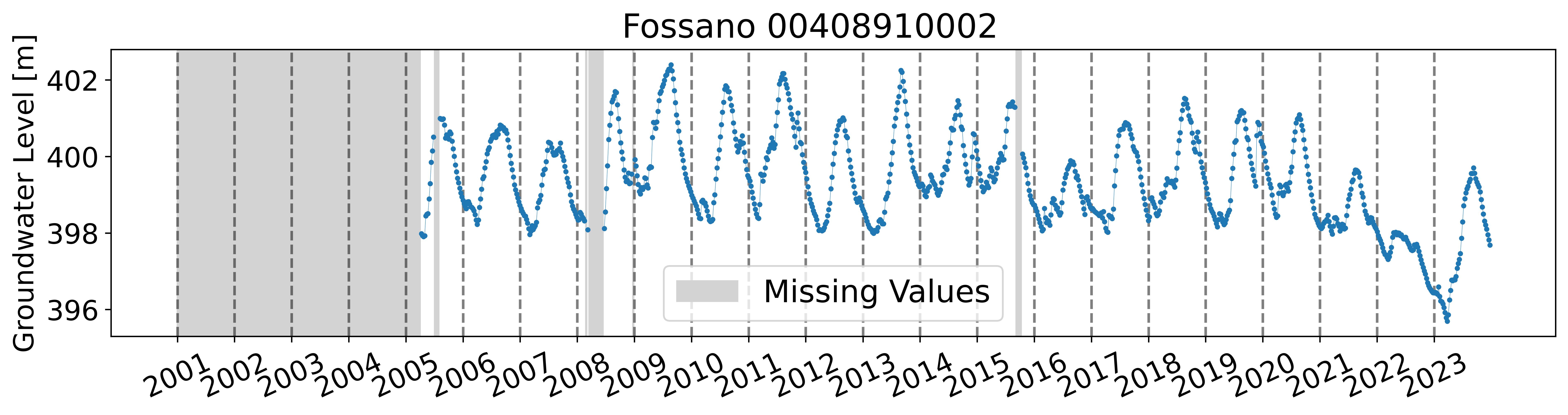}
        \caption{}
        \label{fig:foss2_ts}
    \end{subfigure}
    \hfill
    \begin{subfigure}{0.49\linewidth}
        \centering
        \includegraphics[width=\linewidth]{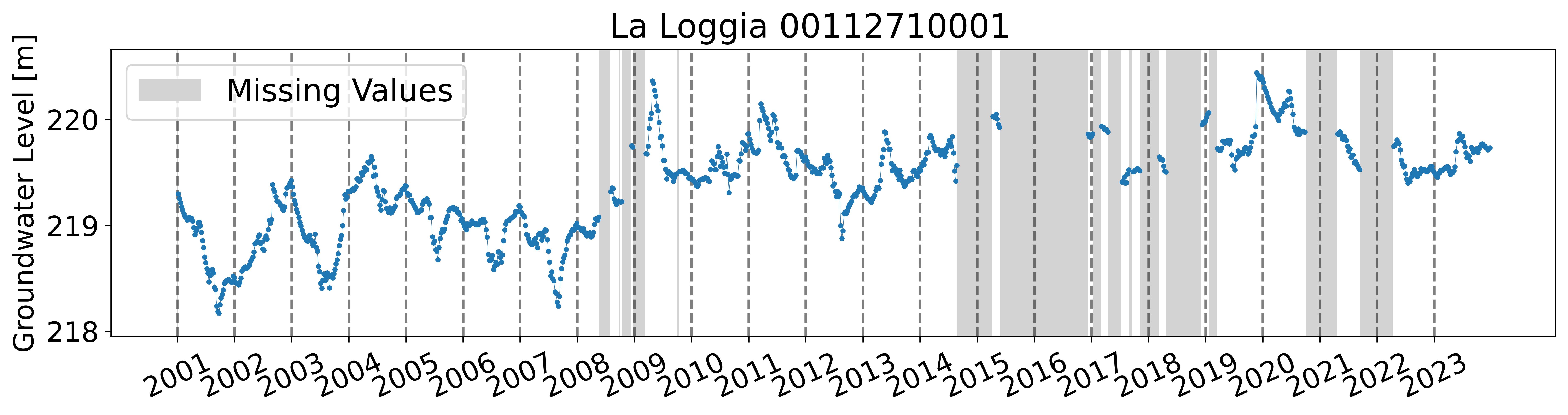}
        \caption{}
        \label{fig:lal_ts}
    \end{subfigure}
    \hfill
    \begin{subfigure}{0.49\linewidth}
        \centering
        \includegraphics[width=\linewidth]{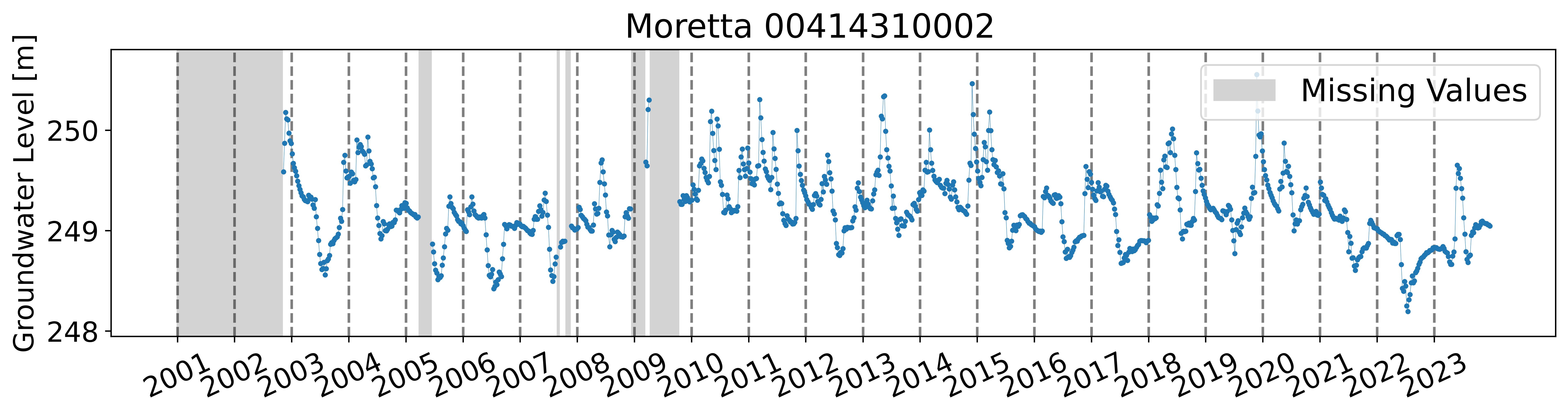}
        \caption{}
        \label{fig:mor_ts}
    \end{subfigure}
\end{figure}
\end{landscape}

\begin{landscape}
\begin{figure}[ht]
\vspace{-2cm}
    \addtocounter{figure}{-1}  
    \begin{subfigure}{0.49\linewidth}
        \centering
        \addtocounter{subfigure}{16} 
        \includegraphics[width=\linewidth]{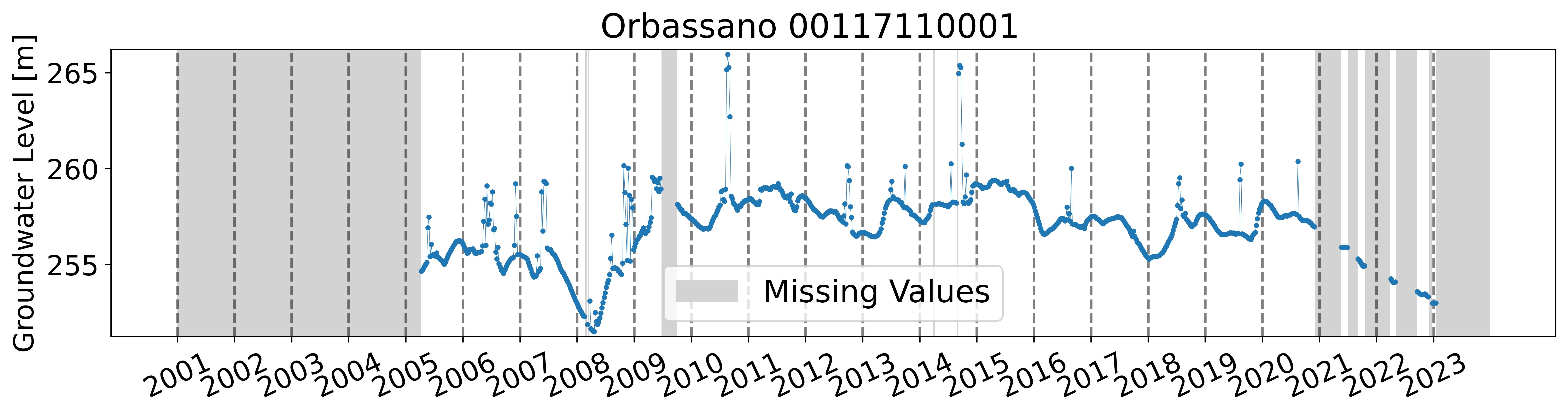}
        \caption{}
        \label{fig:orb_ts}
    \end{subfigure}
    \hfill
    \begin{subfigure}{0.49\linewidth}
        \centering
        \includegraphics[width=\linewidth]{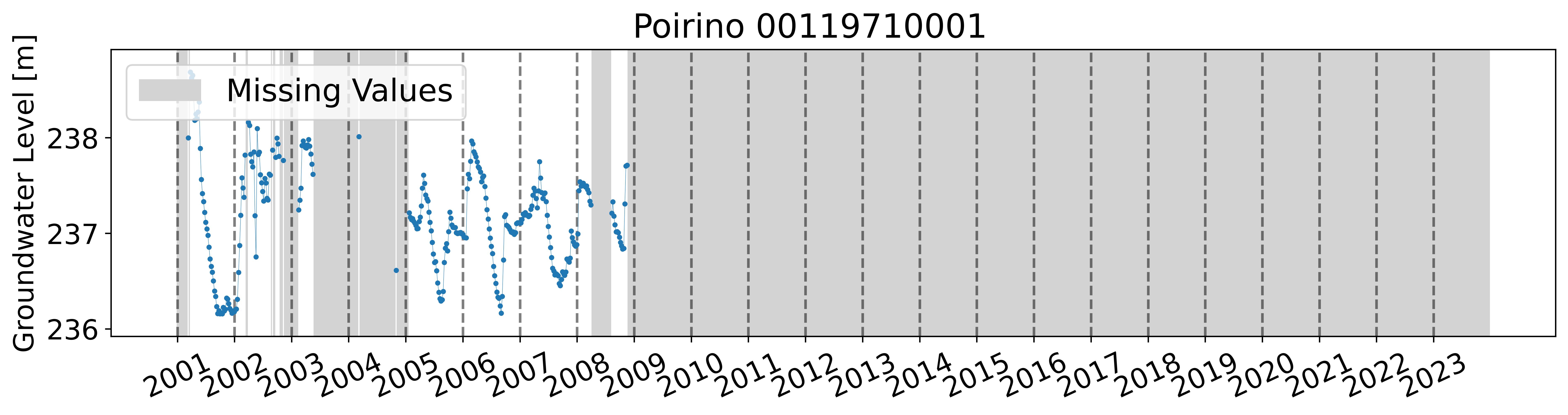}
        \caption{}
        \label{fig:poi_ts}
    \end{subfigure}
    \hfill
    \begin{subfigure}{0.49\linewidth}
        \centering
        \includegraphics[width=\linewidth]{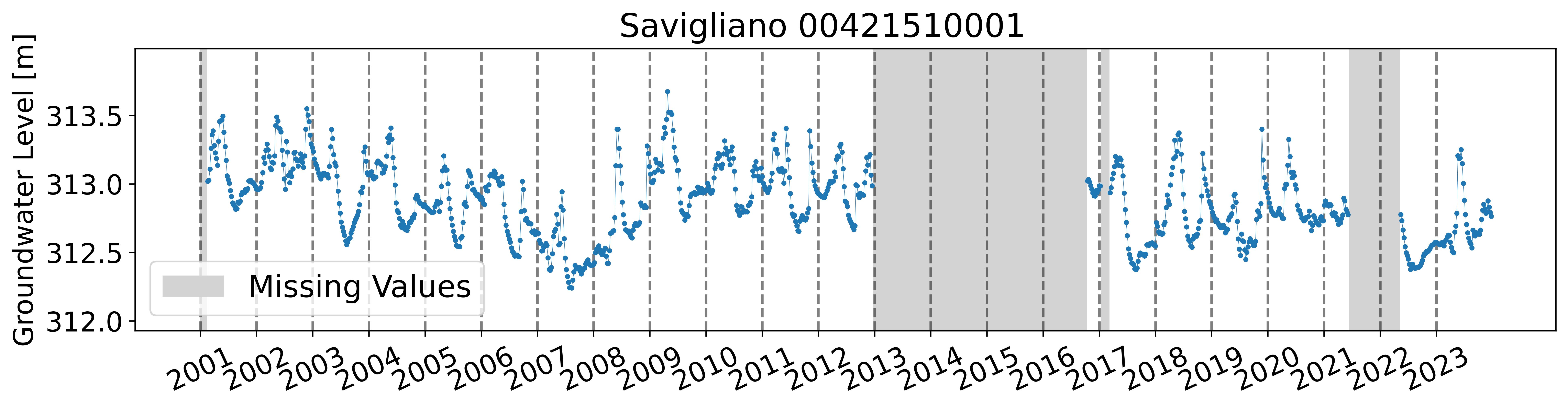}
        \caption{}
        \label{fig:sav_ts}
    \end{subfigure}
    \hfill
    \begin{subfigure}{0.49\linewidth}
        \centering
        \includegraphics[width=\linewidth]{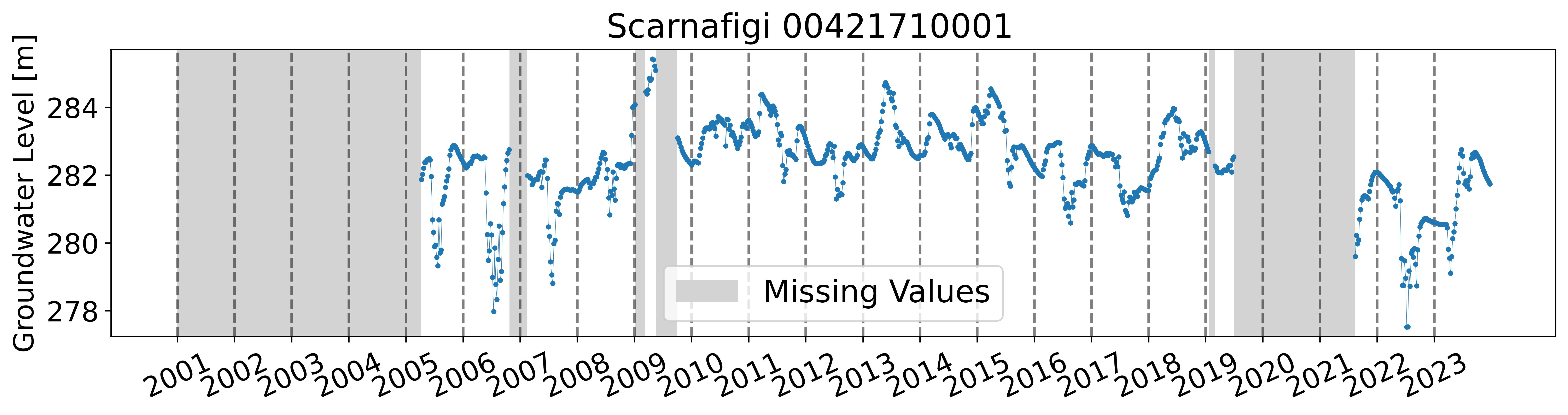}
        \caption{}
        \label{fig:scarn_ts}
    \end{subfigure}
    \hfill
    \begin{subfigure}{0.49\linewidth}
        \centering
        \includegraphics[width=\linewidth]{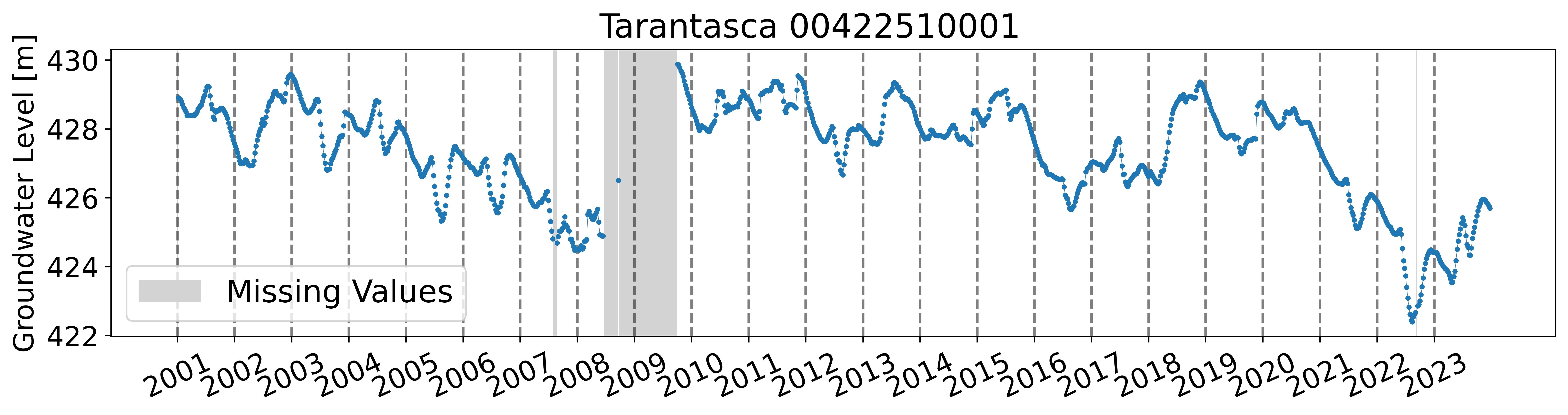}
        \caption{}
        \label{fig:tara_ts}
    \end{subfigure}
    \hfill
    \begin{subfigure}{0.49\linewidth}
        \centering
        \includegraphics[width=\linewidth]{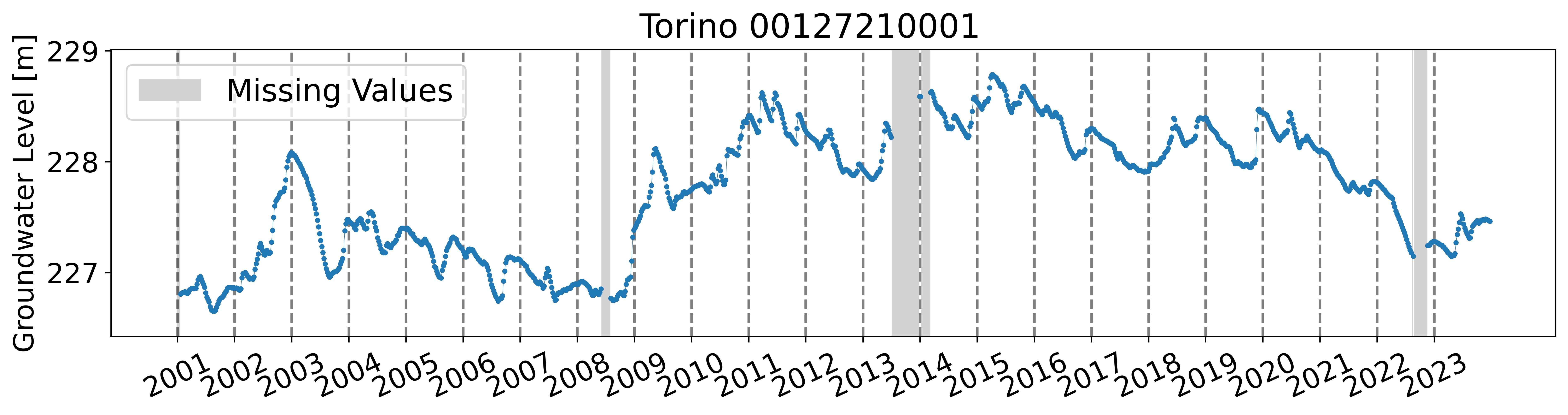}
        \caption{}
        \label{fig:tor1_ts}
    \end{subfigure}
    \hfill
    \begin{subfigure}{0.49\linewidth}
        \centering
        \includegraphics[width=\linewidth]{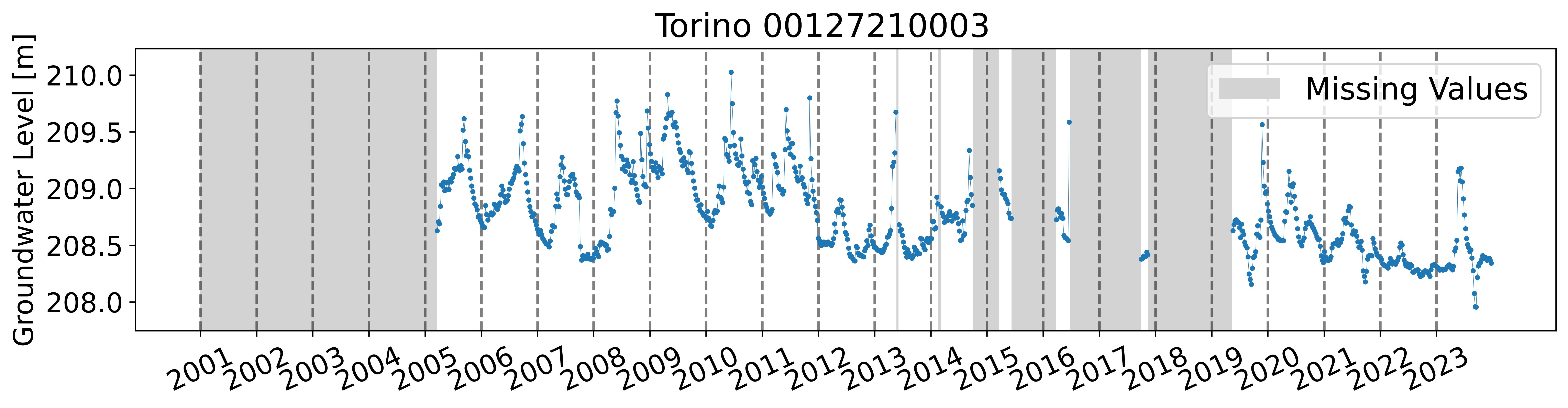}
        \caption{}
        \label{fig:tor3_ts}
    \end{subfigure}
    \hfill
    \begin{subfigure}{0.49\linewidth}
        \centering
        \includegraphics[width=\linewidth]{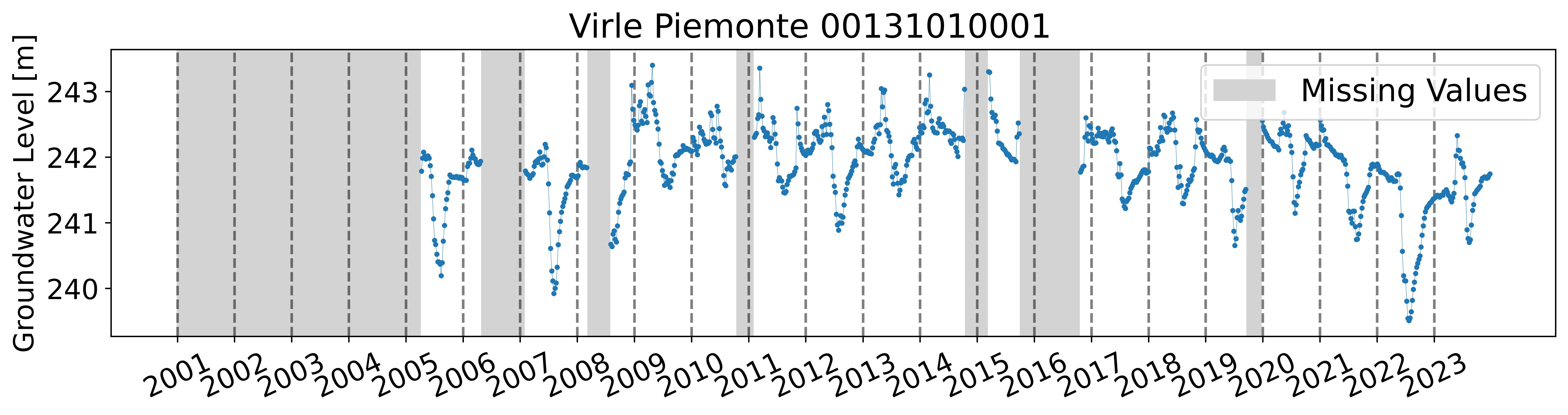}
        \caption{}
        \label{fig:vir_ts}
    \end{subfigure}
    \caption{Groundwater level time series.}
    \label{fig:all_ts}
\end{figure}
\end{landscape}


\begin{table*}[h]
\centering
\caption{STAINet metrics on the test set per sensor in the rollout setting.}
\begin{adjustbox}{width=0.9\textwidth}
\begin{tabular}{l|ccccc}
\hline
\textbf{Sensor (Municipality - Sensor ID)} & \textbf{NBIAS} & \textbf{RMSE[m]} & \textbf{MAPE[\%]} & \textbf{NSE} & \textbf{KGE} \\
\cmidrule(lr){1-6}
\morecmidrules
\cmidrule(lr){1-6}
\textbf{Bricherasio - 00103510001} & 0.479 & 1.389 & 0.375 & 0.354 & -0.016 \\
\textbf{Buriasco - 00104110001} & 0.334 & 3.132 & 1.097 & 0.199 & 0.431 \\
\textbf{Candiolo - 00105110001} & 0.149 & 0.564 & 0.205 & 0.440 & 0.328 \\
\textbf{Carmagnola - 00105910001} & 0.236 & 1.029 & 0.400 & 0.406 & 0.293 \\
\textbf{Carmagnola - 00105910002} & 0.085 & 0.330 & 0.122 & 0.625 & 0.366 \\
\textbf{Cavour - 00107010001} & 0.171 & 2.175 & 0.657 & 0.210 & 0.258 \\
\textbf{Collegno - 00109010001} & 0.283 & 1.472 & 0.515 & 0.268 & 0.224 \\
\textbf{La Loggia - 00112710001} & -0.214 & 0.624 & 0.234 & -3.188 & -1.148 \\
\textbf{Orbassano - 00117110001} & 0.230 & 3.471 & 1.311 & 0.067 & -0.801 \\
\textbf{Poirino - 00119710001} & - & - & - & - & - \\
\textbf{Scalenghe - 00126010001} & 0.315 & 1.708 & 0.625 & 0.288 & 0.577 \\
\textbf{Torino - 00127210001} & 0.059 & 0.501 & 0.182 & -0.640 & -0.753 \\
\textbf{Torino - 00127210003} & 0.089 & 0.450 & 0.162 & 0.181 & -0.257 \\
\textbf{Virle Piemonte - 00131010001} & 0.084 & 0.622 & 0.189 & 0.562 & 0.435 \\
\textbf{Barge - 00401210001} & -0.015 & 0.427 & 0.104 & 0.401 & 0.592 \\
\textbf{Bra - 00402910001} & -0.011 & 0.360 & 0.098 & 0.136 & -0.116 \\
\textbf{Busca - 00403410001} & - & - & - & - & - \\
\textbf{Caramagna Piemonte - 00404110001} & 0.248 & 0.835 & 0.287 & 0.374 & -0.534 \\
\textbf{Cavallermaggiore - 00405910001} & - & - & - & - & - \\
\textbf{Cuneo - 00407810001} & 0.314 & 2.525 & 0.498 & 0.330 & 0.740 \\
\textbf{Fossano - 00408910001} & 0.608 & 1.491 & 0.399 & 0.106 & -0.188 \\
\textbf{Fossano - 00408910002} & 0.392 & 2.034 & 0.450 & 0.144 & 0.499 \\
\textbf{Moretta - 00414310002} & 0.030 & 0.279 & 0.092 & 0.644 & 0.442 \\
\textbf{Racconigi - 00417910001} & 0.102 & 0.789 & 0.252 & 0.424 & 0.210 \\
\textbf{Savigliano - 00421510001} & 0.023 & 0.225 & 0.057 & 0.523 & 0.312 \\
\textbf{Scarnafigi - 00421710001} & 0.141 & 1.561 & 0.458 & 0.323 & 0.218 \\
\textbf{Tarantasca - 00422510001} & 0.499 & 2.974 & 0.637 & 0.162 & 0.031 \\
\textbf{Vottignasco - 00425010001} & 0.234 & 1.353 & 0.285 & 0.058 & -0.094 \\
\hline
\end{tabular}
\end{adjustbox}
\label{tab:PI_sens_metrics_STNet}
\end{table*}
\begin{table*}[h!]
\centering
\caption{PSTAINet-IB metrics on the test set per sensor in the rollout setting.}
\begin{adjustbox}{width=0.9\textwidth}
\begin{tabular}{l|ccccc}
\hline
\textbf{Sensor (Municipality - Sensor ID)} & \textbf{NBIAS} & \textbf{RMSE[m]} & \textbf{MAPE[\%]} & \textbf{NSE} & \textbf{KGE} \\
\cmidrule(lr){1-6}
\morecmidrules
\cmidrule(lr){1-6}
\textbf{Bricherasio - 00103510001} & 0.517 & 1.482 & 0.406 & 0.265 & -0.165 \\
\textbf{Buriasco - 00104110001} & 0.200 & 2.078 & 0.675 & 0.647 & 0.074 \\
\textbf{Candiolo - 00105110001} & 0.098 & 0.380 & 0.128 & 0.745 & 0.618 \\
\textbf{Carmagnola - 00105910001} & 0.183 & 0.829 & 0.318 & 0.614 & 0.497 \\
\textbf{Carmagnola - 00105910002} & 0.017 & 0.183 & 0.066 & 0.885 & 0.742 \\
\textbf{Cavour - 00107010001} & 0.047 & 1.269 & 0.391 & 0.731 & 0.183 \\
\textbf{Collegno - 00109010001} & 0.172 & 0.934 & 0.314 & 0.705 & 0.667 \\
\textbf{La Loggia - 00112710001} & -0.188 & 0.506 & 0.201 & -1.755 & -0.575 \\
\textbf{Orbassano - 00117110001} & 0.086 & 1.372 & 0.491 & 0.854 & -0.392 \\
\textbf{Poirino - 00119710001} & - & - & - & - & - \\
\textbf{Scalenghe - 00126010001} & 0.256 & 1.448 & 0.507 & 0.489 & 0.564 \\
\textbf{Torino - 00127210001} & 0.014 & 0.397 & 0.142 & -0.031 & -0.296 \\
\textbf{Torino - 00127210003} & 0.285 & 0.613 & 0.261 & -0.519 & 0.167 \\
\textbf{Virle Piemonte - 00131010001} & 0.077 & 0.441 & 0.151 & 0.779 & 0.746 \\
\textbf{Barge - 00401210001} & -0.011 & 0.422 & 0.100 & 0.414 & 0.520 \\
\textbf{Bra - 00402910001} & 0.043 & 0.252 & 0.072 & 0.576 & 0.437 \\
\textbf{Busca - 00403410001} & - & - & - & - & - \\
\textbf{Caramagna Piemonte - 00404110001} & 0.250 & 0.734 & 0.289 & 0.517 & 0.798 \\
\textbf{Cavallermaggiore - 00405910001} & - & - & - & - & - \\
\textbf{Cuneo - 00407810001} & 0.202 & 1.809 & 0.340 & 0.656 & 0.767 \\
\textbf{Fossano - 00408910001} & 0.389 & 0.943 & 0.256 & 0.642 & 0.582 \\
\textbf{Fossano - 00408910002} & 0.324 & 1.658 & 0.366 & 0.432 & 0.568 \\
\textbf{Moretta - 00414310002} & 0.056 & 0.270 & 0.089 & 0.666 & 0.604 \\
\textbf{Racconigi - 00417910001} & 0.080 & 0.654 & 0.226 & 0.604 & 0.361 \\
\textbf{Savigliano - 00421510001} & -0.071 & 0.199 & 0.049 & 0.627 & 0.643 \\
\textbf{Scarnafigi - 00421710001} & 0.069 & 1.284 & 0.391 & 0.542 & 0.149 \\
\textbf{Tarantasca - 00422510001} & 0.261 & 1.715 & 0.344 & 0.721 & 0.063 \\
\textbf{Vottignasco - 00425010001} & 0.047 & 0.645 & 0.142 & 0.786 & 0.596 \\
\hline
\end{tabular}
\end{adjustbox}
\label{tab:PI_sens_metrics_STDisNet}
\end{table*}

\begin{table*}[h]
\centering
\caption{PSTAINet-ILB metrics on the test set per sensor in the rollout setting.}
\begin{adjustbox}{width=0.9\textwidth}
\begin{tabular}{l|ccccc}
\hline
\textbf{Sensor (Municipality - Sensor ID)} & \textbf{NBIAS} & \textbf{RMSE[m]} & \textbf{MAPE[\%]} & \textbf{NSE} & \textbf{KGE} \\
\cmidrule(lr){1-6}
\morecmidrules
\cmidrule(lr){1-6}
\textbf{Bricherasio - 00103510001} & 0.276 & 0.795 & 0.216 & 0.788 & 0.531 \\
\textbf{Buriasco - 00104110001} & 0.089 & 1.230 & 0.363 & 0.877 & 0.403 \\
\textbf{Candiolo - 00105110001} & -0.122 & 0.418 & 0.143 & 0.692 & 0.567 \\
\textbf{Carmagnola - 00105910001} & 0.021 & 0.444 & 0.154 & 0.889 & 0.331 \\
\textbf{Carmagnola - 00105910002} & -0.065 & 0.253 & 0.095 & 0.779 & 0.675 \\
\textbf{Cavour - 00107010001} & -0.037 & 1.234 & 0.328 & 0.746 & 0.298 \\
\textbf{Collegno - 00109010001} & -0.001 & 0.312 & 0.094 & 0.967 & 0.855 \\
\textbf{La Loggia - 00112710001} & -0.399 & 0.928 & 0.413 & -8.280 & -0.045 \\
\textbf{Orbassano - 00117110001} & 0.013 & 0.576 & 0.203 & 0.974 & -0.204 \\
\textbf{Poirino - 00119710001} & - & - & - & - & - \\
\textbf{Scalenghe - 00126010001} & 0.070 & 0.600 & 0.182 & 0.912 & 0.745 \\
\textbf{Torino - 00127210001} & -0.378 & 0.866 & 0.355 & -3.899 & 0.069 \\
\textbf{Torino - 00127210003} & -0.065 & 0.366 & 0.141 & 0.459 & -0.005 \\
\textbf{Virle Piemonte - 00131010001} & -0.043 & 0.433 & 0.144 & 0.788 & 0.690 \\
\textbf{Barge - 00401210001} & -0.079 & 0.454 & 0.109 & 0.324 & 0.585 \\
\textbf{Bra - 00402910001} & 0.014 & 0.201 & 0.054 & 0.731 & 0.580 \\
\textbf{Busca - 00403410001} & - & - & - & - & - \\
\textbf{Caramagna Piemonte - 00404110001} & 0.236 & 0.690 & 0.273 & 0.573 & 0.870 \\
\textbf{Cavallermaggiore - 00405910001} & - & - & - & - & - \\
\textbf{Cuneo - 00407810001} & 0.075 & 0.933 & 0.169 & 0.908 & 0.835 \\
\textbf{Fossano - 00408910001} & 0.248 & 0.603 & 0.163 & 0.854 & 0.611 \\
\textbf{Fossano - 00408910002} & 0.286 & 1.549 & 0.327 & 0.504 & 0.561 \\
\textbf{Moretta - 00414310002} & -0.005 & 0.199 & 0.065 & 0.819 & 0.630 \\
\textbf{Racconigi - 00417910001} & -0.001 & 0.447 & 0.160 & 0.815 & 0.468 \\
\textbf{Savigliano - 00421510001} & -0.118 & 0.226 & 0.058 & 0.519 & 0.689 \\
\textbf{Scarnafigi - 00421710001} & 0.012 & 1.155 & 0.337 & 0.629 & 0.188 \\
\textbf{Tarantasca - 00422510001} & 0.034 & 0.753 & 0.155 & 0.946 & 0.603 \\
\textbf{Vottignasco - 00425010001} & 0.036 & 0.585 & 0.122 & 0.824 & 0.635 \\
\hline
\end{tabular}
\end{adjustbox}
\label{tab:PI_sens_metrics_STDisNetPI}
\end{table*}

\begin{table*}[h]
\centering
\caption{PSTAINet-ILRB metrics on the test set per sensor in the rollout setting.}
\begin{adjustbox}{width=0.9\textwidth}
\begin{tabular}{l|ccccc}
\hline
\textbf{Sensor (Municipality - Sensor ID)} & \textbf{NBIAS} & \textbf{RMSE[m]} & \textbf{MAPE[\%]} & \textbf{NSE} & \textbf{KGE} \\
\cmidrule(lr){1-6}
\morecmidrules
\cmidrule(lr){1-6}
\textbf{Bricherasio - 00103510001} & 0.415 & 1.161 & 0.325 & 0.548 & 0.295 \\
\textbf{Buriasco - 00104110001} & 0.160 & 1.889 & 0.544 & 0.709 & 0.441 \\
\textbf{Candiolo - 00105110001} & 0.069 & 0.357 & 0.134 & 0.775 & 0.496 \\
\textbf{Carmagnola - 00105910001} & 0.053 & 0.524 & 0.195 & 0.846 & 0.424 \\
\textbf{Carmagnola - 00105910002} & -0.083 & 0.297 & 0.108 & 0.695 & 0.552 \\
\textbf{Cavour - 00107010001} & -0.014 & 1.317 & 0.368 & 0.710 & 0.228 \\
\textbf{Collegno - 00109010001} & 0.061 & 0.678 & 0.193 & 0.844 & 0.407 \\
\textbf{La Loggia - 00112710001} & -0.345 & 0.840 & 0.357 & -6.607 & -0.585 \\
\textbf{Orbassano - 00117110001} & 0.099 & 1.777 & 0.618 & 0.756 & -0.995 \\
\textbf{Poirino - 00119710001} & - & - & - & - & - \\
\textbf{Scalenghe - 00126010001} & 0.231 & 1.311 & 0.458 & 0.581 & 0.671 \\
\textbf{Torino - 00127210001} & -0.188 & 0.608 & 0.230 & -1.411 & -0.599 \\
\textbf{Torino - 00127210003} & 0.157 & 0.511 & 0.174 & -0.057 & -0.285 \\
\textbf{Virle Piemonte - 00131010001} & 0.050 & 0.481 & 0.165 & 0.738 & 0.702 \\
\textbf{Barge - 00401210001} & -0.028 & 0.378 & 0.074 & 0.530 & 0.422 \\
\textbf{Bra - 00402910001} & 0.025 & 0.243 & 0.069 & 0.606 & 0.258 \\
\textbf{Busca - 00403410001} & - & - & - & - & - \\
\textbf{Caramagna Piemonte - 00404110001} & 0.114 & 0.522 & 0.176 & 0.756 & -0.601 \\
\textbf{Cavallermaggiore - 00405910001} & - & - & - & - & - \\
\textbf{Cuneo - 00407810001} & 0.221 & 1.809 & 0.355 & 0.656 & 0.815 \\
\textbf{Fossano - 00408910001} & 0.433 & 1.085 & 0.285 & 0.527 & -0.528 \\
\textbf{Fossano - 00408910002} & 0.295 & 1.560 & 0.337 & 0.497 & 0.626 \\
\textbf{Moretta - 00414310002} & -0.031 & 0.250 & 0.083 & 0.714 & 0.609 \\
\textbf{Racconigi - 00417910001} & 0.039 & 0.546 & 0.183 & 0.724 & 0.321 \\
\textbf{Savigliano - 00421510001} & -0.003 & 0.169 & 0.045 & 0.732 & 0.675 \\
\textbf{Scarnafigi - 00421710001} & 0.068 & 1.225 & 0.362 & 0.583 & 0.182 \\
\textbf{Tarantasca - 00422510001} & 0.256 & 1.641 & 0.328 & 0.745 & 0.316 \\
\textbf{Vottignasco - 00425010001} & 0.116 & 0.760 & 0.156 & 0.702 & 0.618 \\
\hline
\end{tabular}
\end{adjustbox}
\label{tab:PI_sens_metrics_STDisNetPI-RCH}
\end{table*}

\begin{landscape}
\begin{figure}[h]
\vspace{-3.75cm}
\centering
    \begin{subfigure}{0.49\linewidth}
        \centering
        \includegraphics[width=\linewidth]{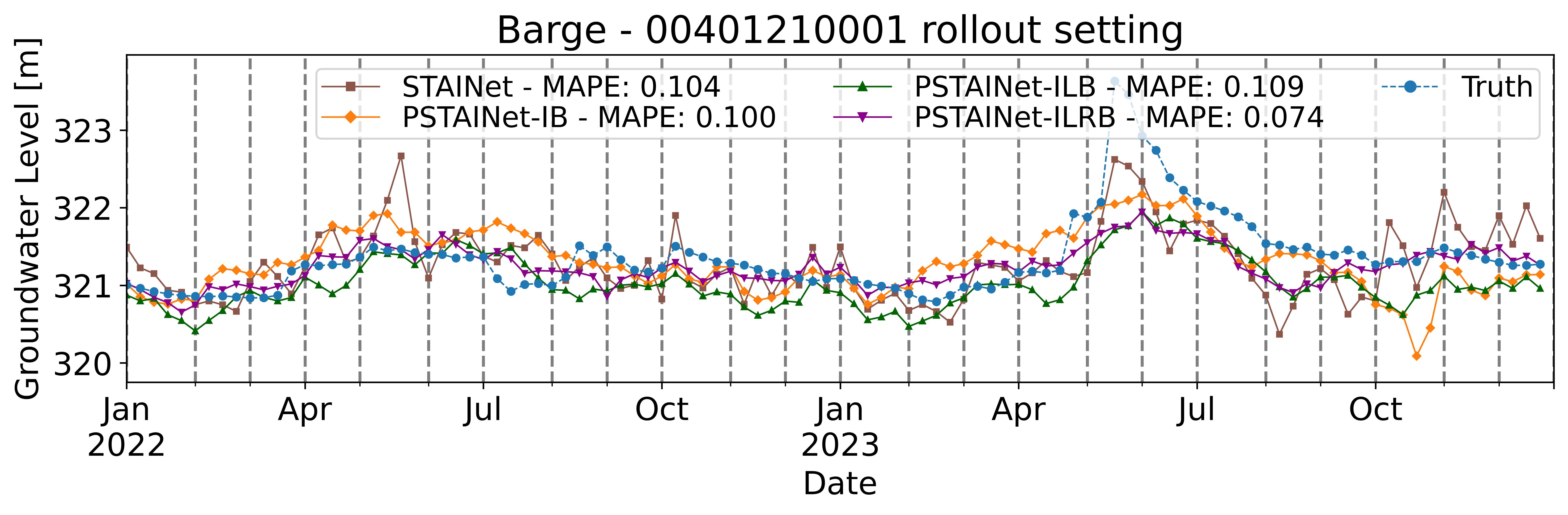}
        \caption{}
        \label{fig:bar_pred_iter}
    \end{subfigure}
    \hfill
    \begin{subfigure}{0.49\linewidth}
        \centering
        \includegraphics[width=\linewidth]{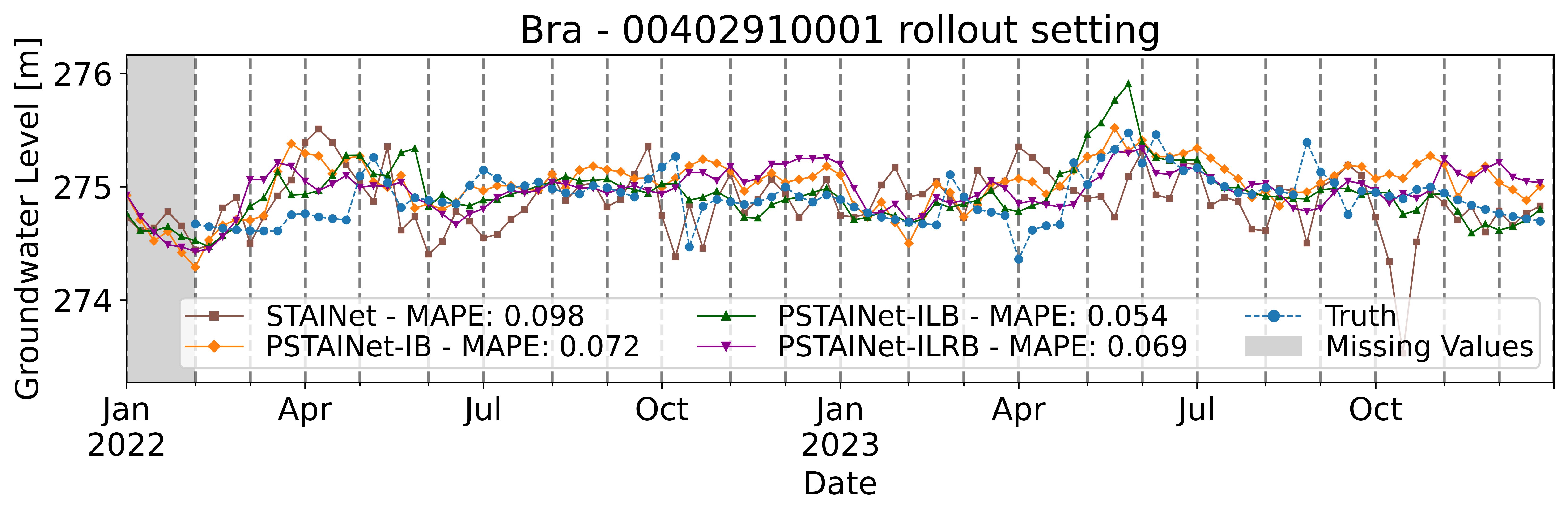}
        \caption{}
        \label{fig:bra_pred_iter}
    \end{subfigure}
    \hfill
    \begin{subfigure}{0.49\linewidth}
        \centering
        \includegraphics[width=\linewidth]{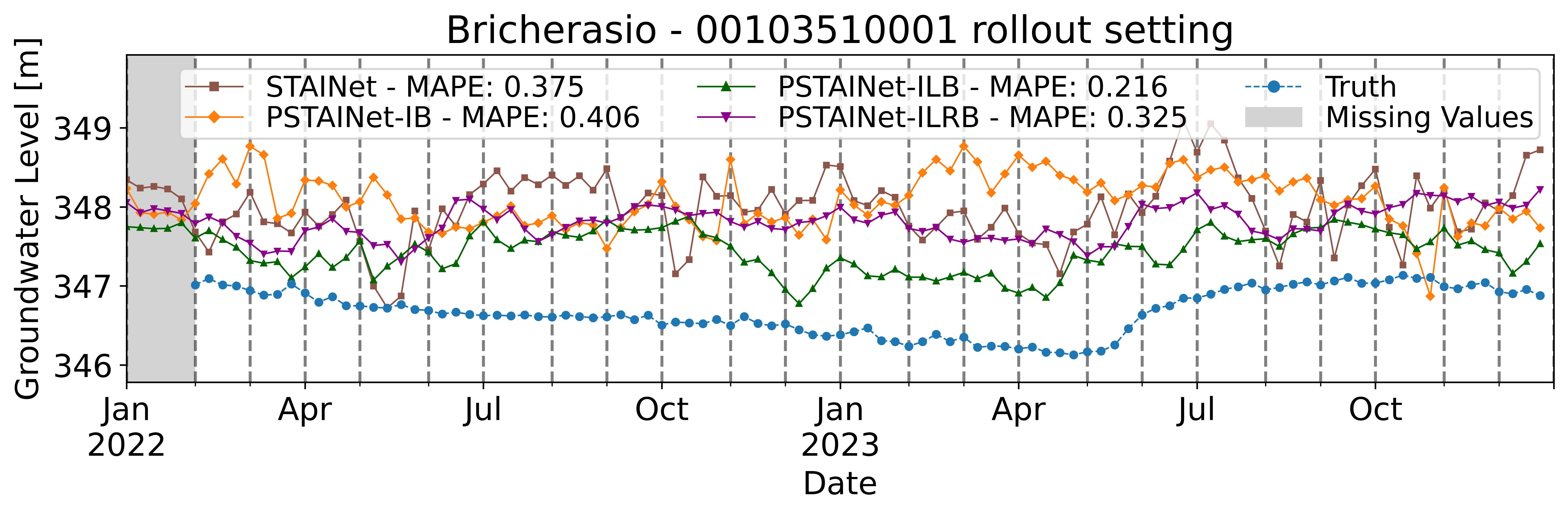}
        \caption{}
        \label{fig:bri_pred_iter}
    \end{subfigure}
    \hfill
    \begin{subfigure}{0.49\linewidth}
        \centering
        \includegraphics[width=\linewidth]{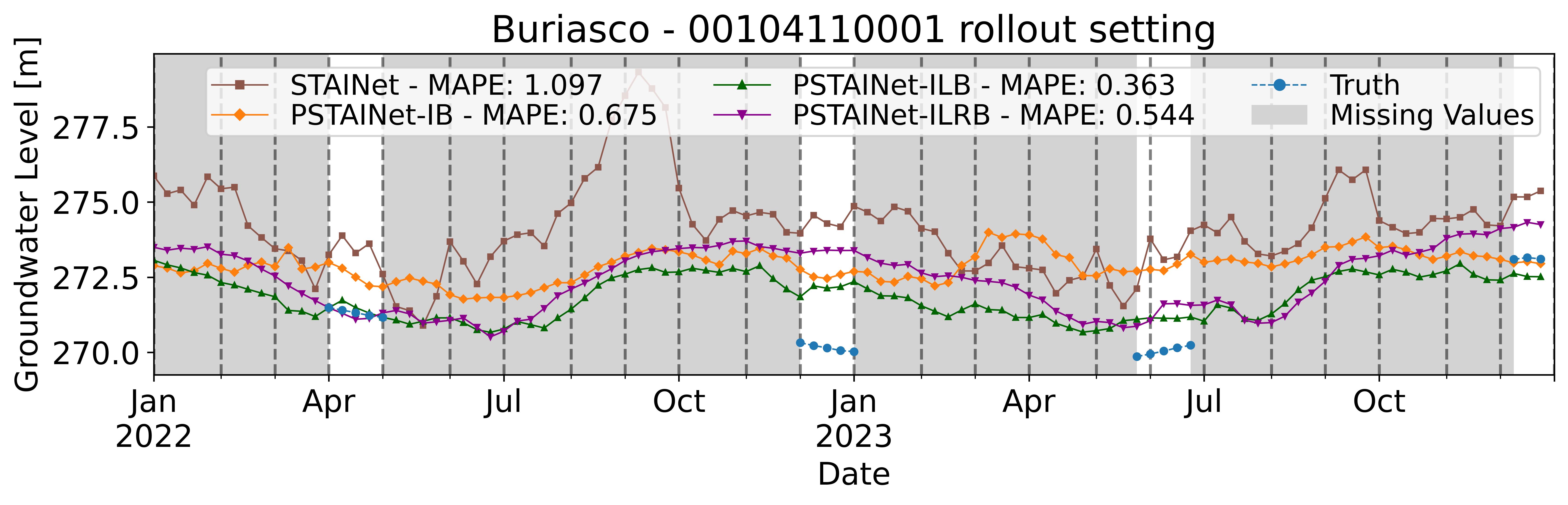}
        \caption{}
        \label{fig:bur_pred_iter}
    \end{subfigure}
    \hfill
    \begin{subfigure}{0.49\linewidth}
        \centering
        \includegraphics[width=\linewidth]{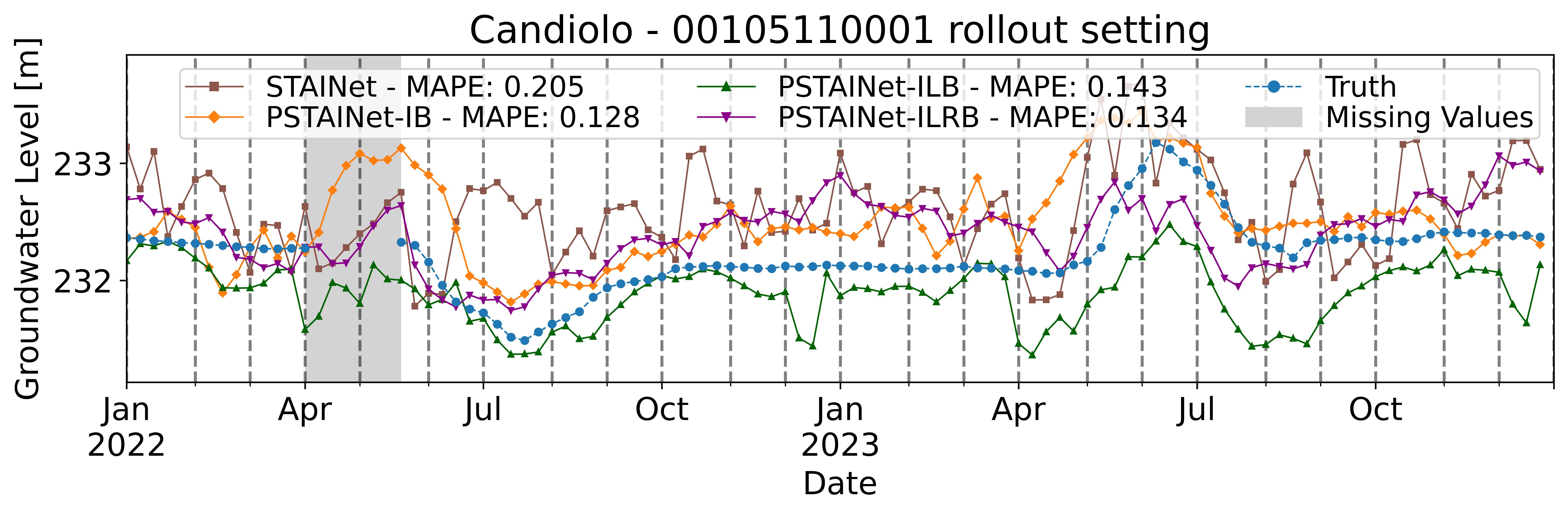}
        \caption{}
        \label{fig:cand_pred_iter}
    \end{subfigure}
    \hfill
    \begin{subfigure}{0.49\linewidth}
        \centering
        \includegraphics[width=\linewidth]{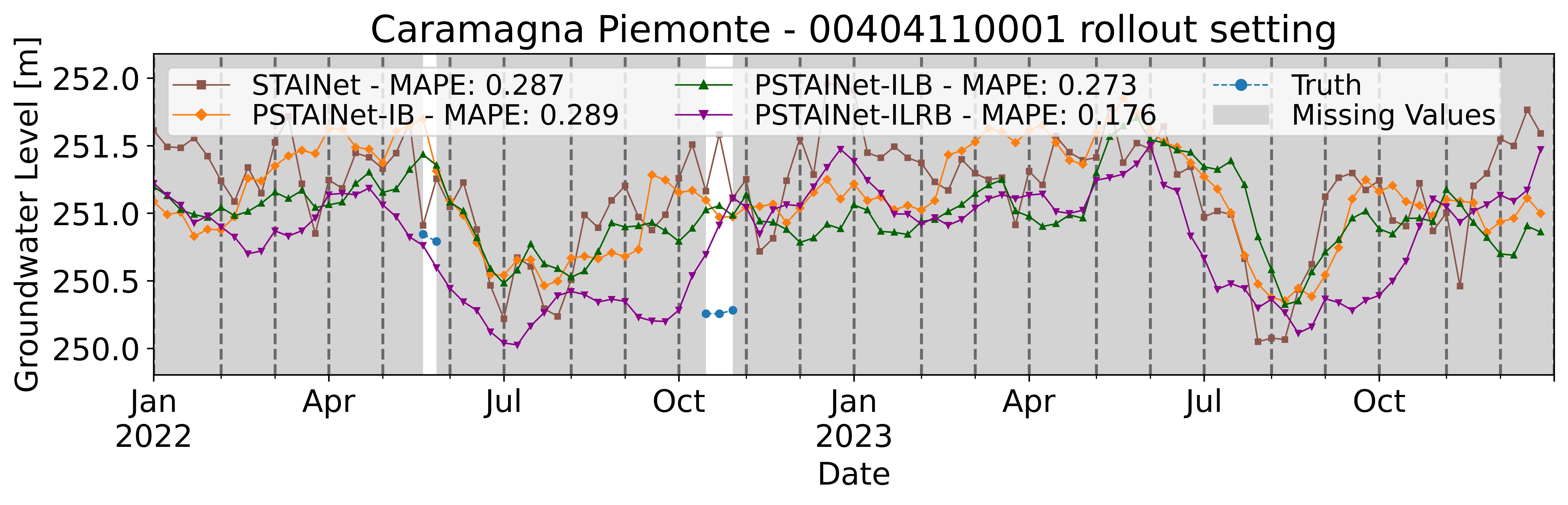}
        \caption{}
        \label{fig:cara_pred_iter}
    \end{subfigure}
    \hfill
    \begin{subfigure}{0.49\linewidth}
        \centering
        \includegraphics[width=\linewidth]{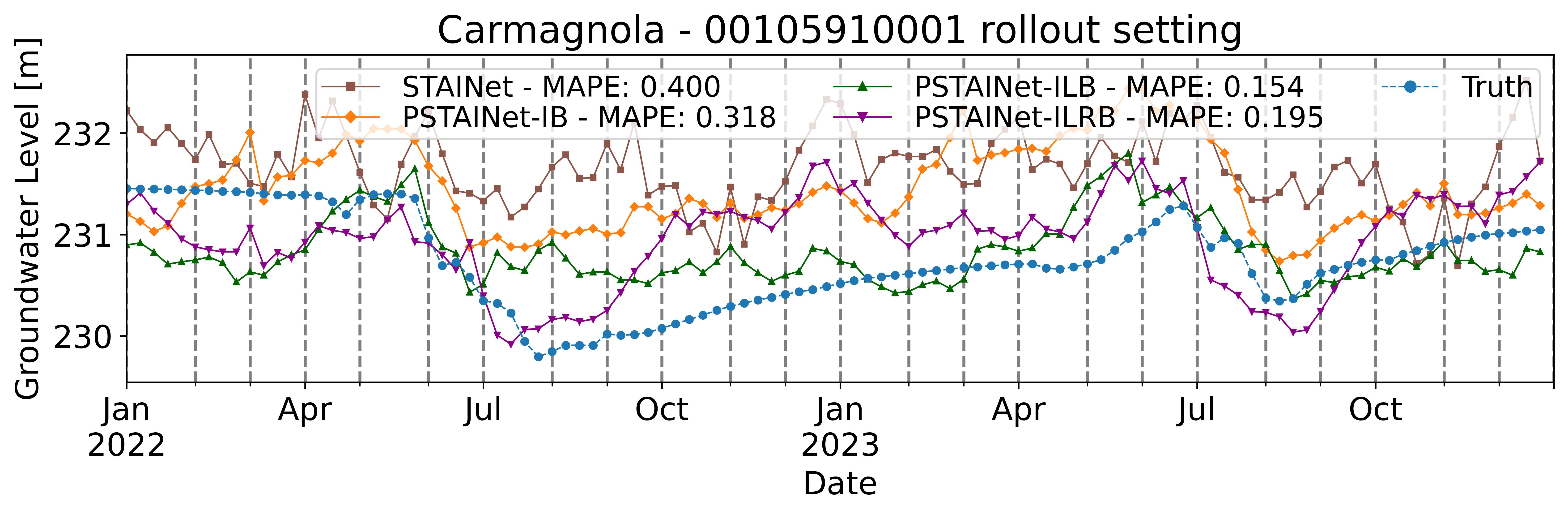}
        \caption{}
        \label{fig:carm1_pred_iter}
    \end{subfigure}
    \hfill
    \begin{subfigure}{0.49\linewidth}
        \centering
        \includegraphics[width=\linewidth]{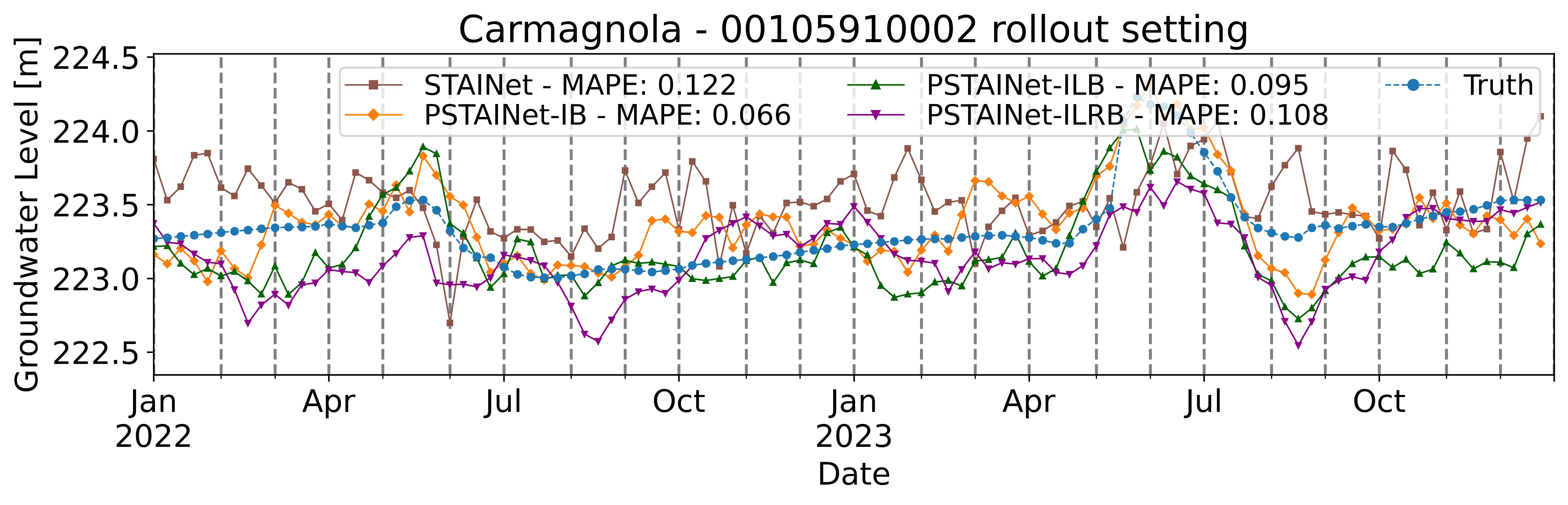}
        \caption{}
        \label{fig:carm2_pred_iter}
    \end{subfigure}
    \end{figure}
\end{landscape}

\begin{landscape}
\begin{figure}[ht]
\vspace{-3.75cm}
    \addtocounter{figure}{-1}  
    \begin{subfigure}{0.49\linewidth}
        \centering
        \addtocounter{subfigure}{8} 
        \includegraphics[width=\linewidth]{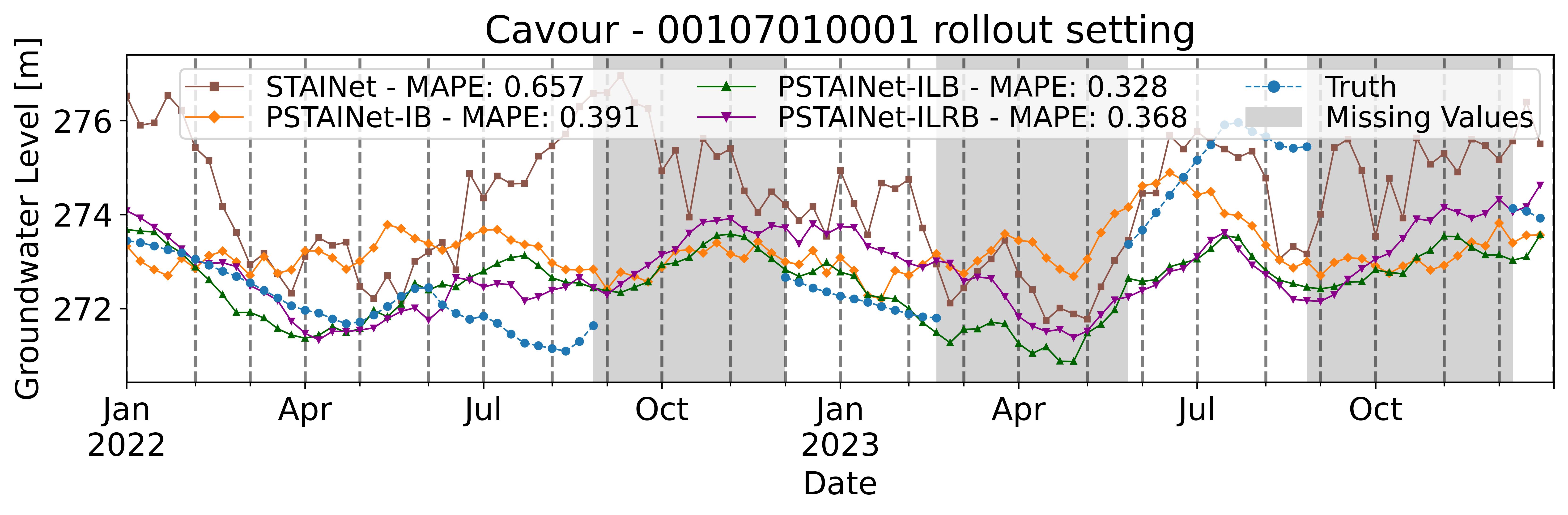}
        \caption{}
        \label{fig:cavo_pred_iter}
    \end{subfigure}
    \hfill
    \begin{subfigure}{0.49\linewidth}
        \centering
        \includegraphics[width=\linewidth]{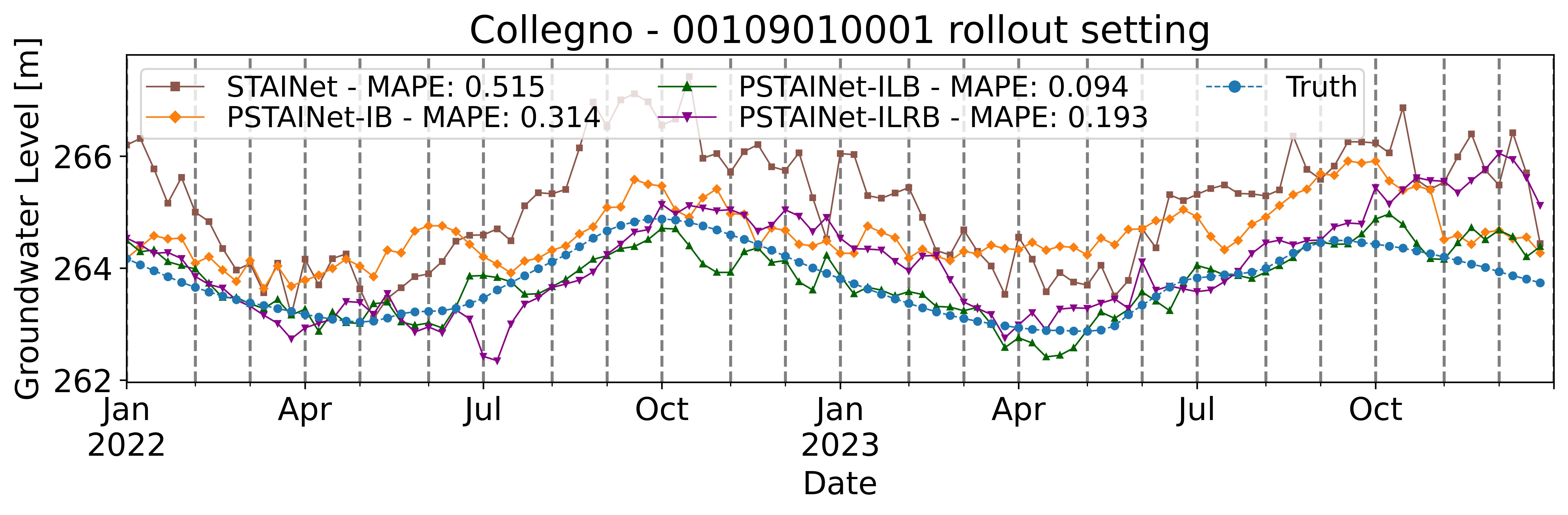}
        \caption{}
        \label{fig:coll_pred_iter}
    \end{subfigure}
    \hfill
    \begin{subfigure}{0.49\linewidth}
        \centering
        \includegraphics[width=\linewidth]{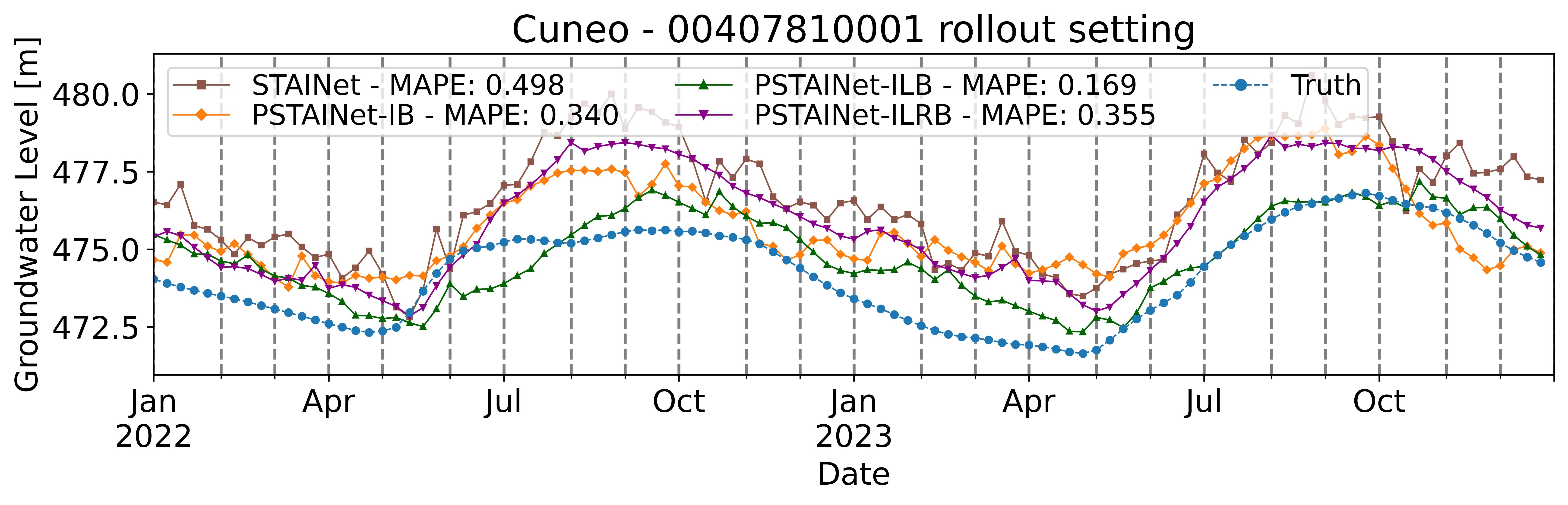}
        \caption{}
        \label{fig:cuneo_pred_iter}
    \end{subfigure}
    \hfill
    \begin{subfigure}{0.49\linewidth}
        \centering
        \includegraphics[width=\linewidth]{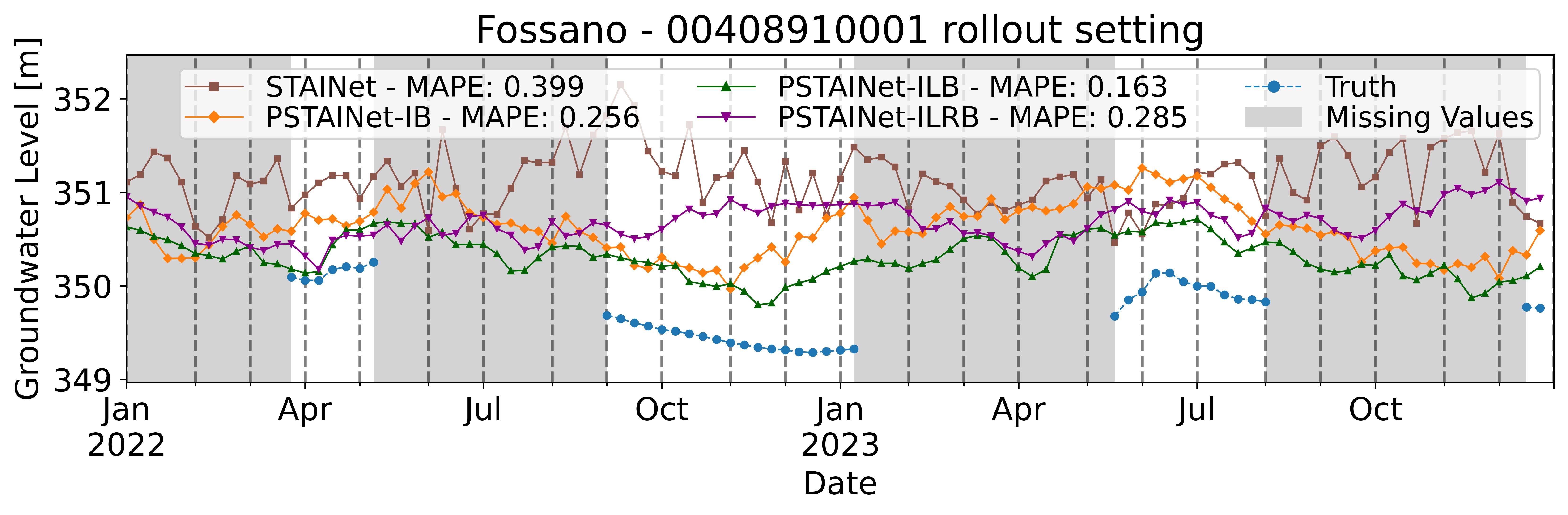}
        \caption{}
        \label{fig:foss1_pred_iter}
    \end{subfigure}
    \hfill
    \begin{subfigure}{0.49\linewidth}
        \centering
        \includegraphics[width=\linewidth]{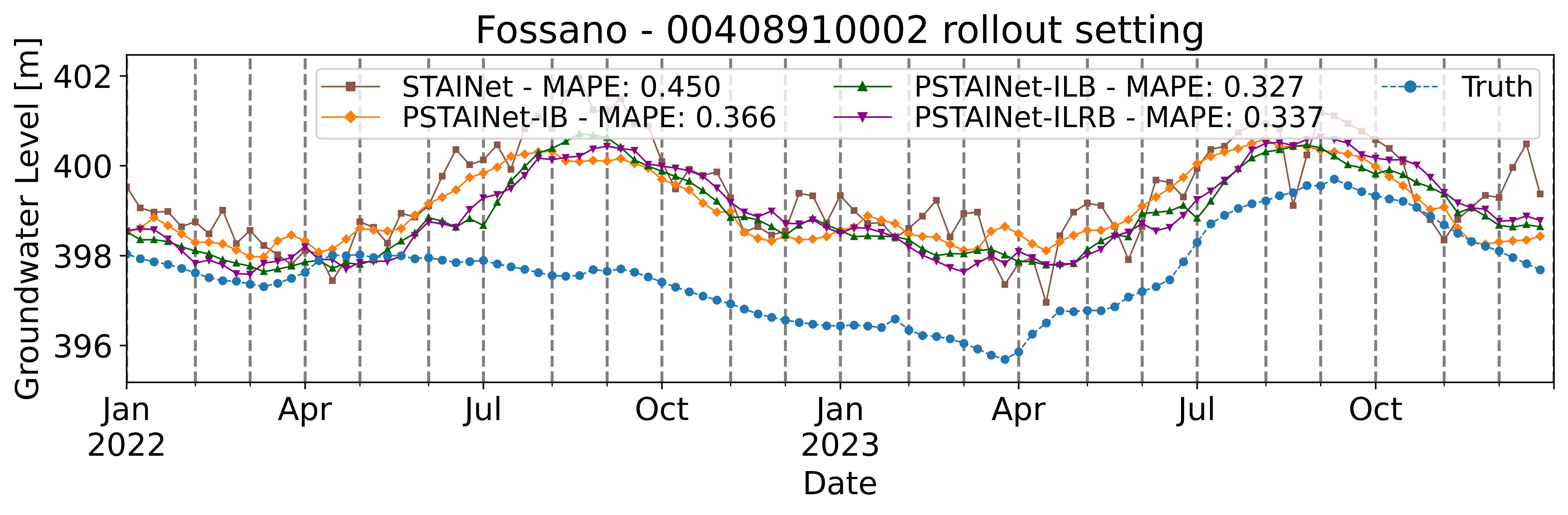}
        \caption{}
        \label{fig:foss2_pred_iter}
    \end{subfigure}
    \hfill    
    \begin{subfigure}{0.49\linewidth}
        \centering
        \includegraphics[width=\linewidth]{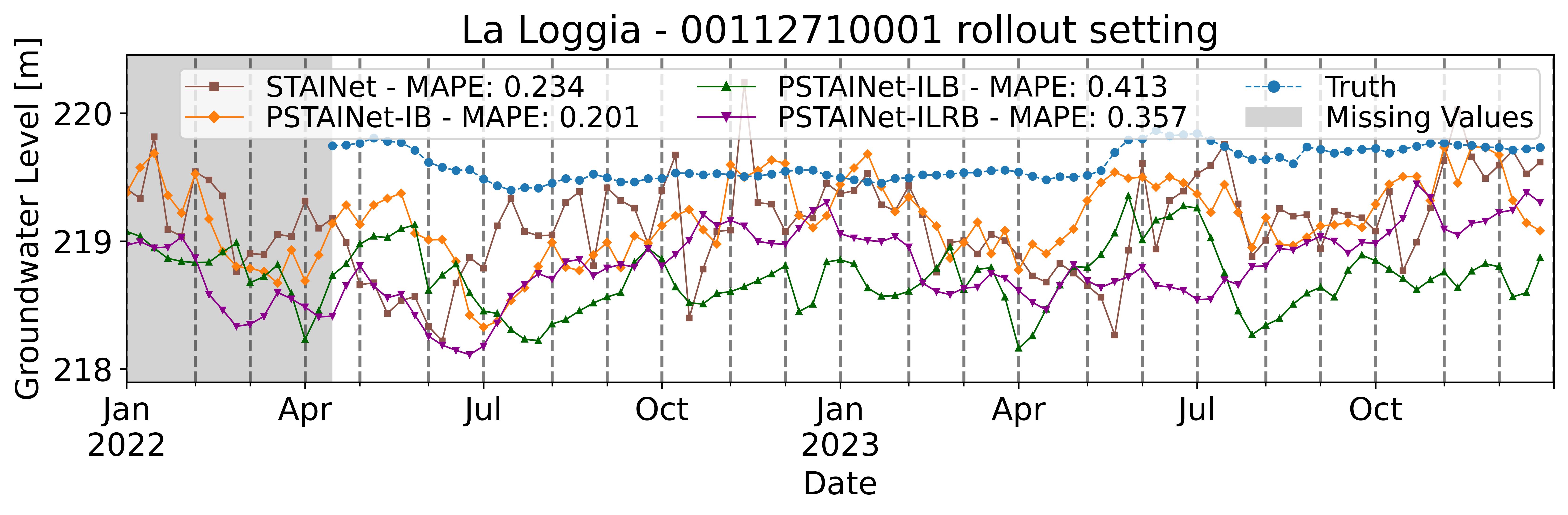}
        \caption{}
        \label{fig:lal_pred_iter}
    \end{subfigure}
    \hfill
    \begin{subfigure}{0.49\linewidth}
        \centering
        \includegraphics[width=\linewidth]{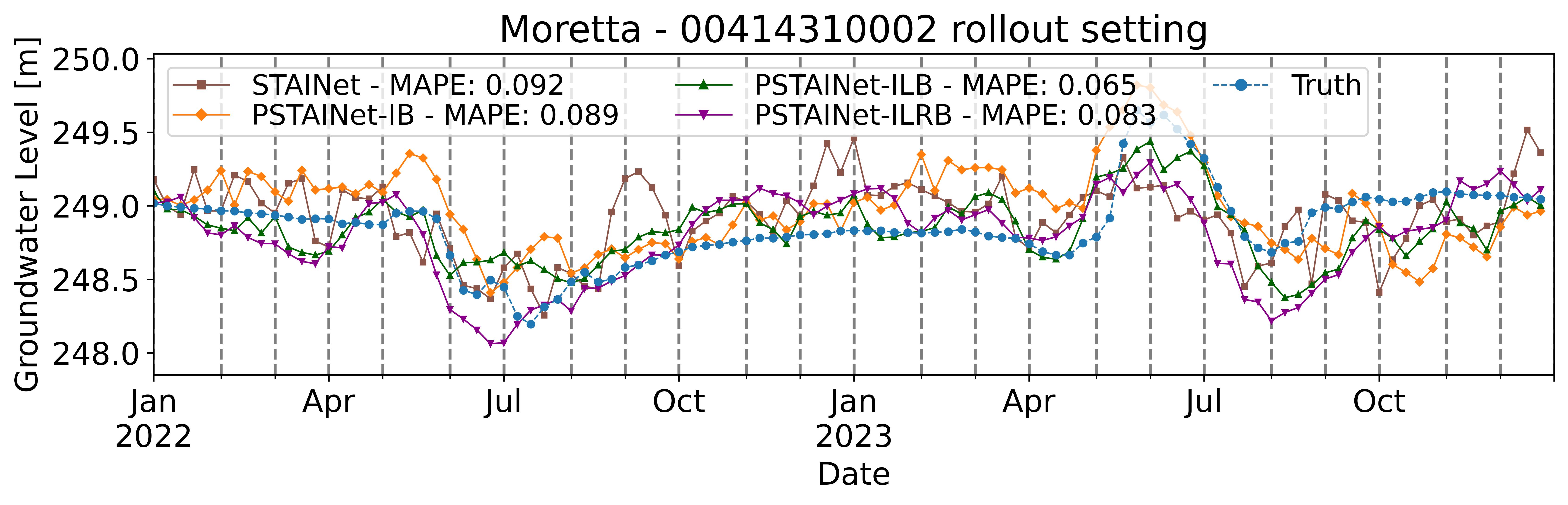}
        \caption{}
        \label{fig:mor_pred_iter}
    \end{subfigure}
    \hfill
    \begin{subfigure}{0.49\linewidth}
        \centering
        \includegraphics[width=\linewidth]{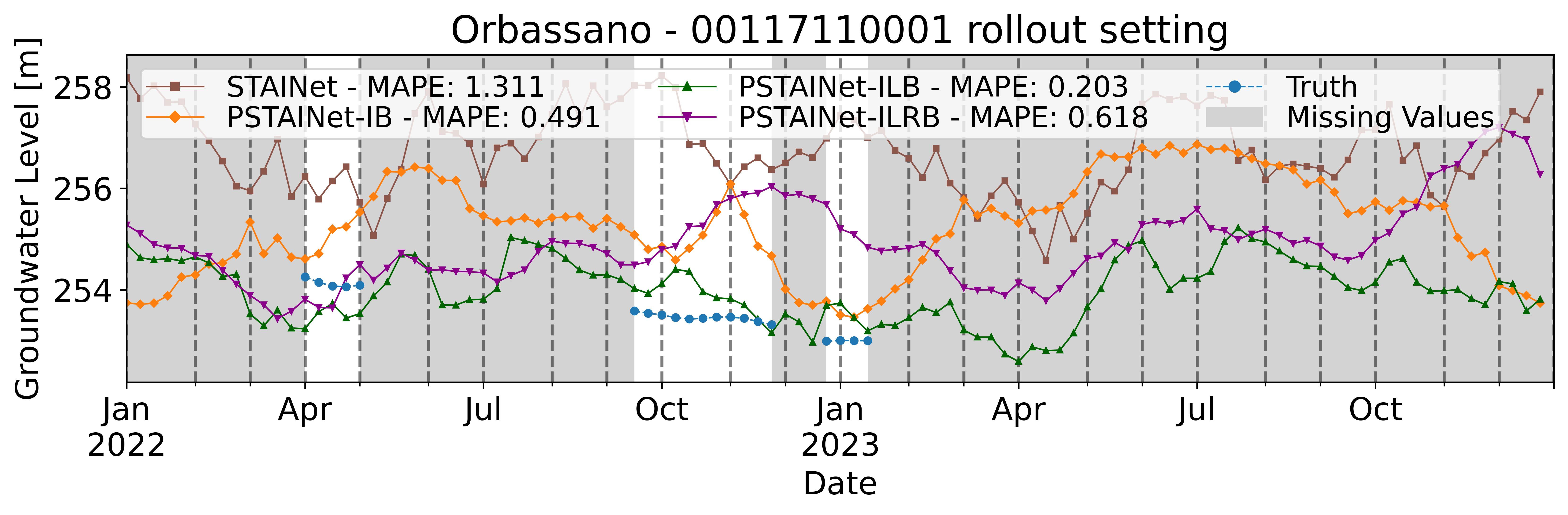}
        \caption{}
        \label{fig:orb_pred_iter}
    \end{subfigure}
    \hfill
    \end{figure}
\end{landscape}

\begin{landscape}
\begin{figure}[ht]
\vspace{-3.75cm}
    \addtocounter{figure}{-1}  
    \begin{subfigure}{0.49\linewidth}
        \centering
        \addtocounter{subfigure}{16} 
        \includegraphics[width=\linewidth]{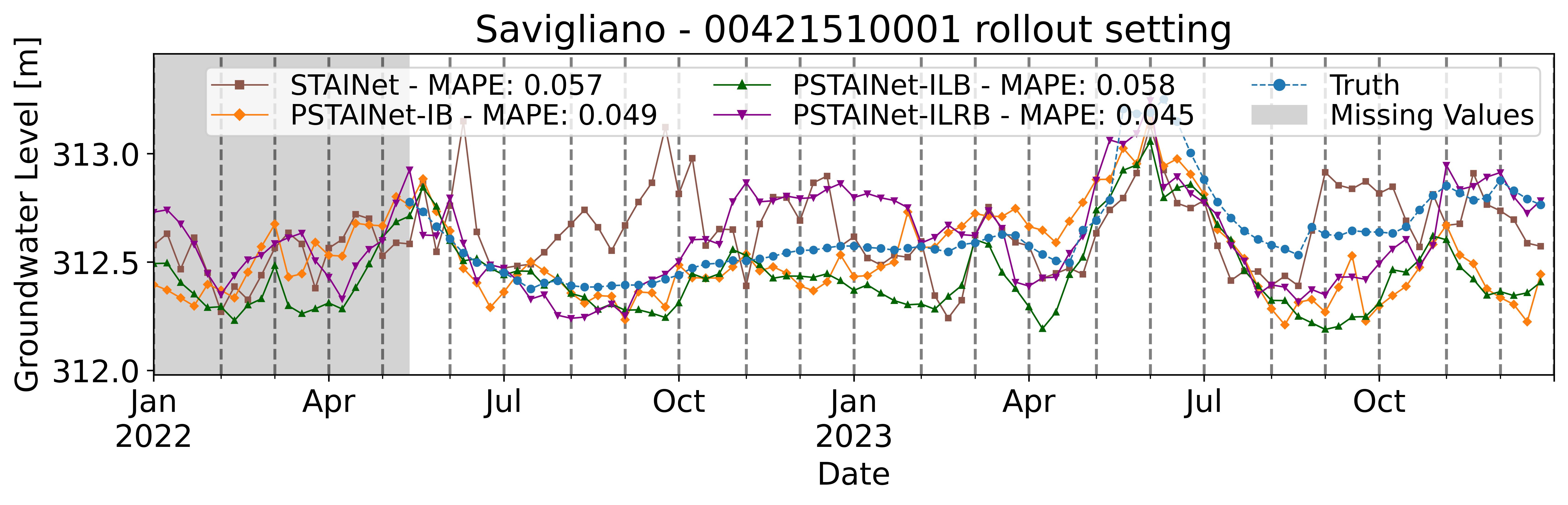}
        \caption{}
        \label{fig:sav_pred_iter}
    \end{subfigure}
    \hfill
    \begin{subfigure}{0.49\linewidth}
        \centering
        \includegraphics[width=\linewidth]{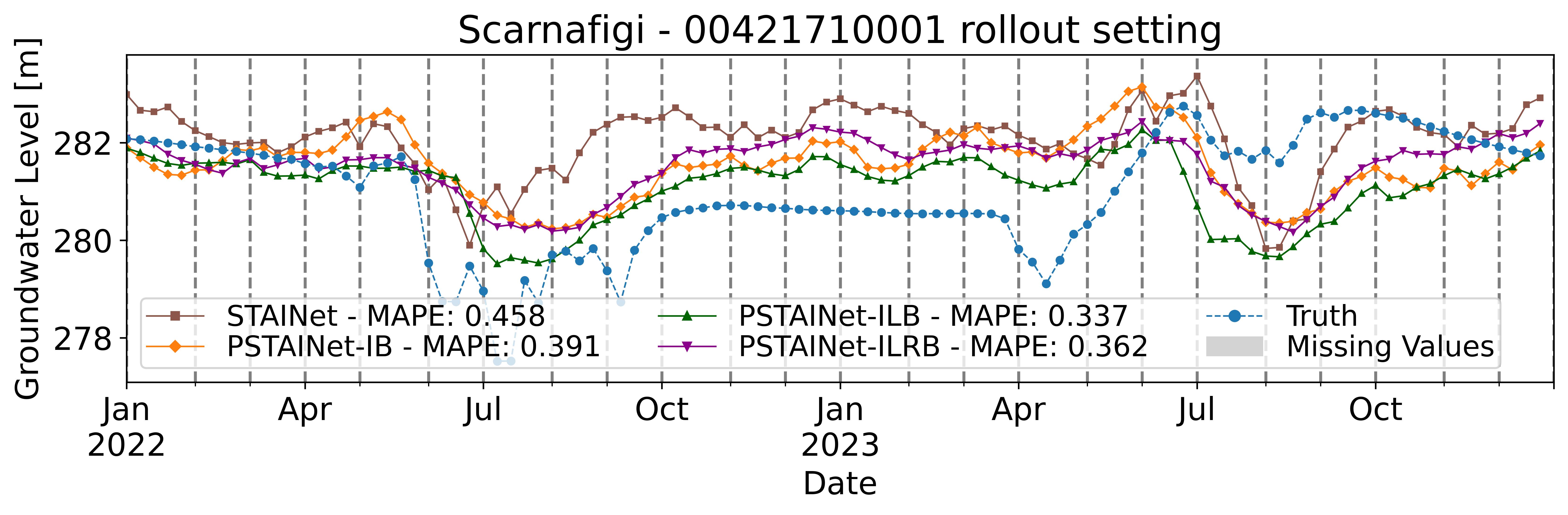}
        \caption{}
        \label{fig:scarn_pred_iter}
    \end{subfigure}
    \hfill
    \begin{subfigure}{0.49\linewidth}
        \centering
        \includegraphics[width=\linewidth]{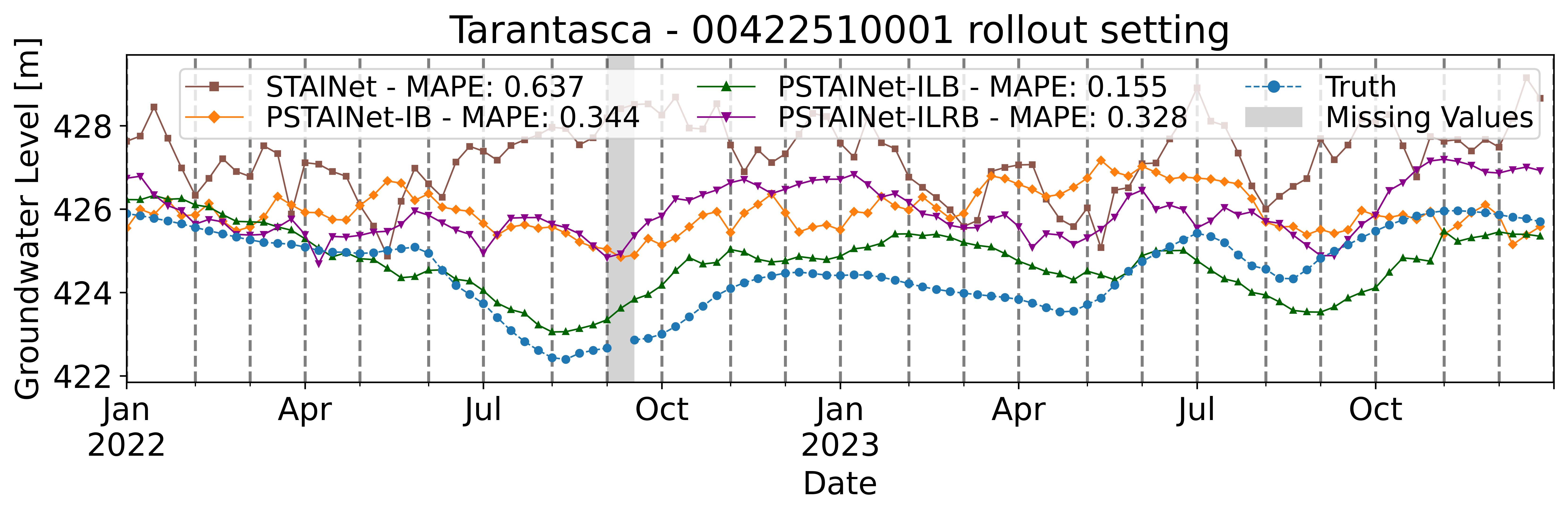}
        \caption{}
        \label{fig:tara_pred_iter}
    \end{subfigure}
    \hfill
    \begin{subfigure}{0.49\linewidth}
        \centering
        \includegraphics[width=\linewidth]{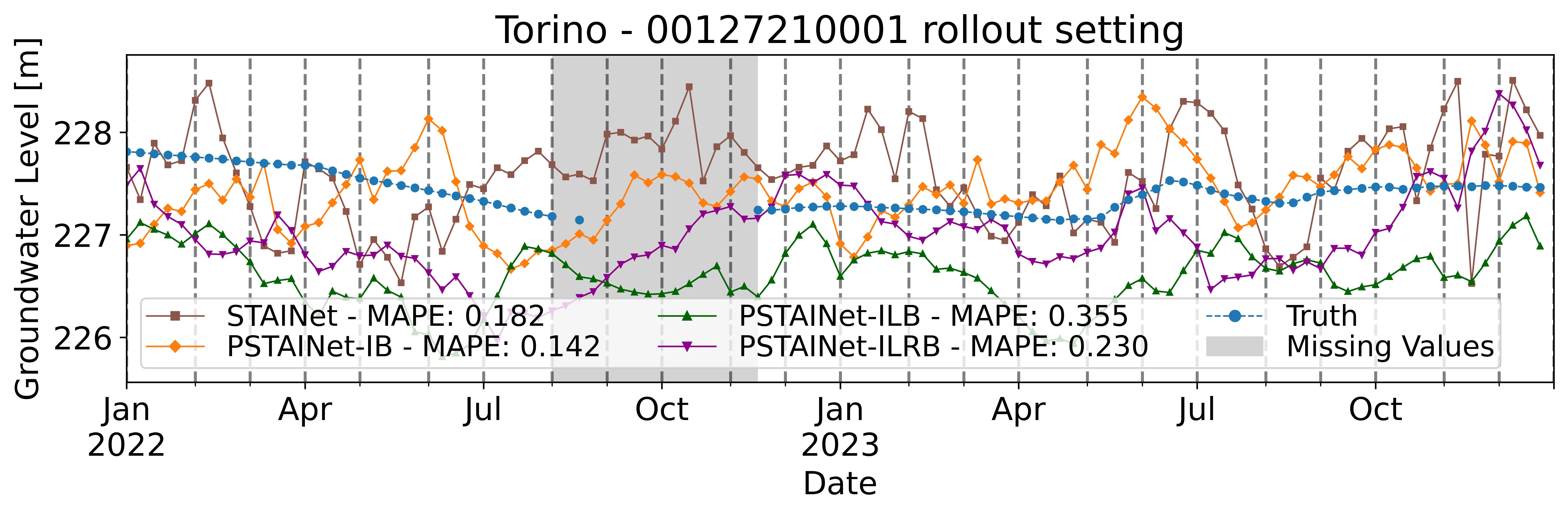}
        \caption{}
        \label{fig:tor1_pred_iter}
    \end{subfigure}
    \hfill
    \begin{subfigure}{0.49\linewidth}
        \centering
        \includegraphics[width=\linewidth]{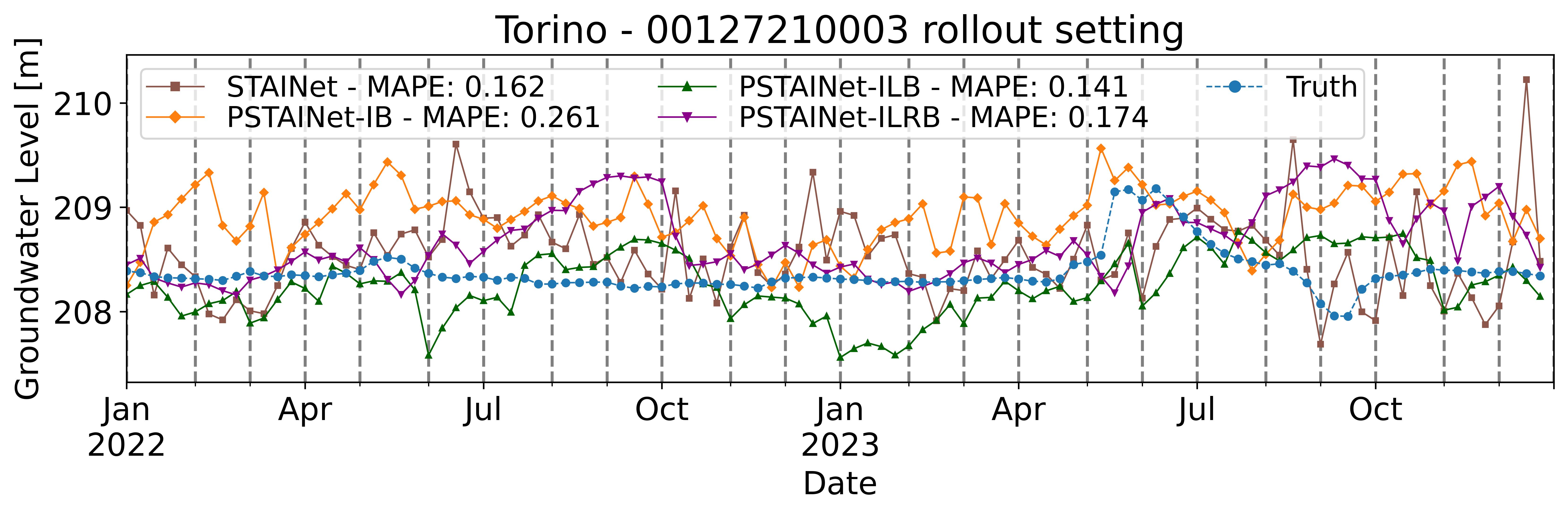}
        \caption{}
        \label{fig:tor3_pred_iter}
    \end{subfigure}
    \hfill
    \begin{subfigure}{0.49\linewidth}
        \centering
        \includegraphics[width=\linewidth]{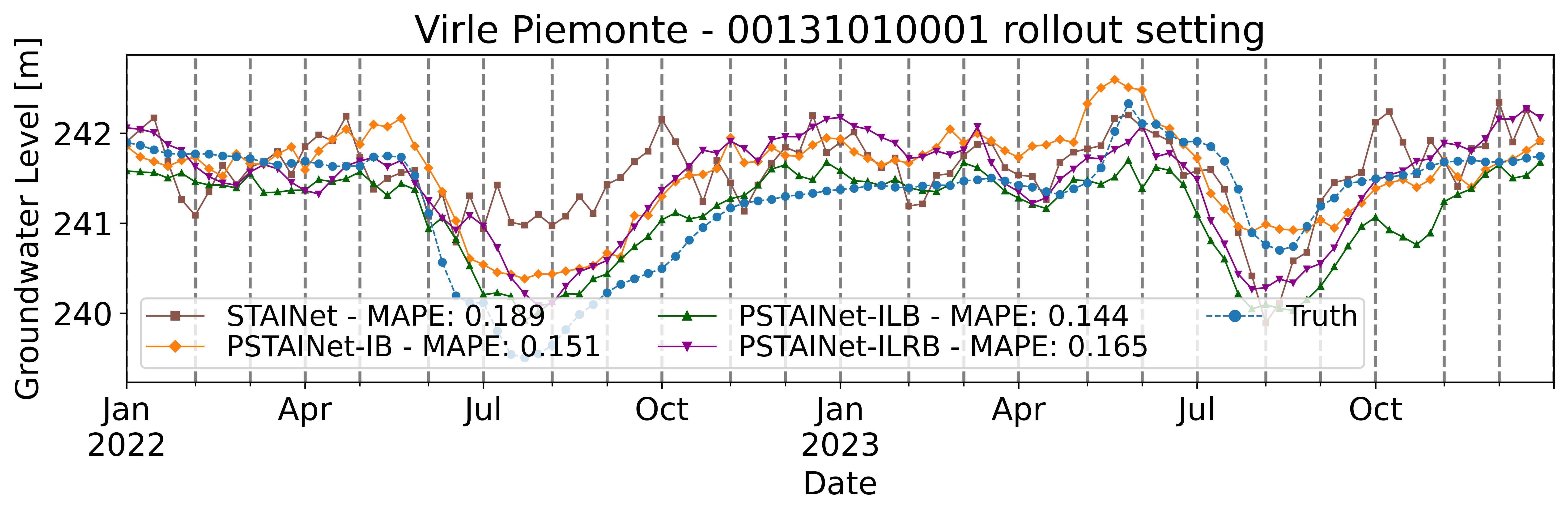}
        \caption{}
        \label{fig:vir_pred_iter}
    \end{subfigure}
    \hfill
    \begin{subfigure}{0.49\linewidth}
        \centering
        \includegraphics[width=\linewidth]{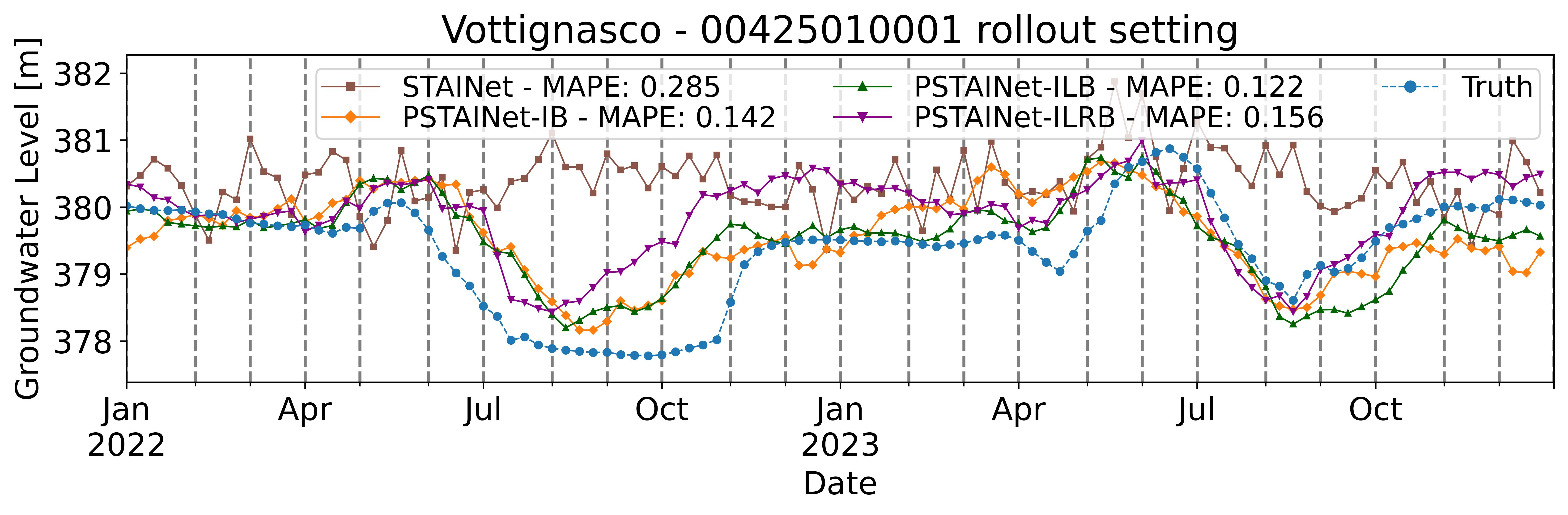}
        \caption{}
        \label{fig:vott_pred_iter}
    \end{subfigure}
    \hfill
\caption{Test set predictions in the rollout setting.}
\label{fig:all_pred_iter}
\end{figure}
\end{landscape}

\begin{table*}[h]
\centering
\caption{PSTAINet-ILB reconstruction metrics per sensor in the rollout setting.}
\begin{adjustbox}{width=0.9\textwidth}
\begin{tabular}{l|ccccc}
\hline
\textbf{Sensor (Municipality - Sensor ID)} &  \textbf{NBIAS} & \textbf{RMSE[m]} & \textbf{MAPE[\%]} & \textbf{NSE} & \textbf{KGE} \\
\cmidrule(lr){1-6}
\morecmidrules
\cmidrule(lr){1-6}
\textbf{Bricherasio - 00103510001} & -0.030 & 0.451 & 0.103 & 0.651 & 0.659 \\
\textbf{Buriasco - 00104110001} & -0.035 & 0.920 & 0.264 & 0.737 & 0.852 \\
\textbf{Candiolo - 00105110001} & -0.123 & 0.433 & 0.148 & 0.311 & 0.760 \\
\textbf{Carmagnola - 00105910001} & -0.038 & 0.471 & 0.165 & 0.677 & 0.727 \\
\textbf{Carmagnola - 00105910002} & -0.090 & 0.294 & 0.106 & 0.475 & 0.785 \\
\textbf{Cavour - 00107010001} & -0.026 & 1.131 & 0.313 & 0.624 & 0.765 \\
\textbf{Collegno - 00109010001} & -0.015 & 0.577 & 0.163 & 0.738 & 0.872 \\
\textbf{La Loggia - 00112710001} & -0.118 & 0.488 & 0.176 & -0.215 & 0.368 \\
\textbf{Orbassano - 00117110001} & -0.033 & 1.213 & 0.309 & 0.540 & 0.724 \\
\textbf{Poirino - 00119710001} & -0.030 & 0.339 & 0.107 & 0.548 & 0.743 \\
\textbf{Scalenghe - 00126010001} & -0.018 & 0.472 & 0.149 & 0.807 & 0.807 \\
\textbf{Torino - 00127210001} & -0.120 & 0.542 & 0.191 & 0.042 & 0.629 \\
\textbf{Torino - 00127210003} & -0.125 & 0.424 & 0.161 & -0.360 & 0.375 \\
\textbf{Virle Piemonte - 00131010001} & -0.058 & 0.339 & 0.112 & 0.671 & 0.820 \\
\textbf{Barge - 00401210001} & -0.057 & 0.377 & 0.087 & 0.407 & 0.709 \\
\textbf{Bra - 00402910001} & -0.025 & 0.257 & 0.066 & 0.442 & 0.602 \\
\textbf{Busca - 00403410001} & -0.024 & 1.154 & 0.196 & 0.621 & 0.792 \\
\textbf{Caramagna Piemonte - 00404110001} & -0.021 & 0.311 & 0.096 & 0.593 & 0.661 \\
\textbf{Cavallermaggiore - 00405910001} & -0.025 & 0.262 & 0.069 & 0.732 & 0.769 \\
\textbf{Cuneo - 00407810001} & 0.003 & 0.928 & 0.148 & 0.760 & 0.872 \\
\textbf{Fossano - 00408910001} & -0.080 & 0.355 & 0.085 & 0.484 & 0.671 \\
\textbf{Fossano - 00408910002} & 0.004 & 0.740 & 0.135 & 0.638 & 0.750 \\
\textbf{Moretta - 00414310002} & -0.016 & 0.175 & 0.055 & 0.749 & 0.840 \\
\textbf{Racconigi - 00417910001} & -0.023 & 0.447 & 0.141 & 0.651 & 0.655 \\
\textbf{Savigliano - 00421510001} & -0.079 & 0.200 & 0.052 & 0.423 & 0.768 \\
\textbf{Scarnafigi - 00421710001} & -0.011 & 0.644 & 0.169 & 0.704 & 0.709 \\
\textbf{Tarantasca - 00422510001} & 0.001 & 0.774 & 0.140 & 0.725 & 0.856 \\
\textbf{Vottignasco - 00425010001} & 0.015 & 0.421 & 0.084 & 0.752 & 0.793 \\
\hline
\end{tabular}
\end{adjustbox}
\label{tab:rec_metrics_all}
\end{table*}

\begin{landscape}
\begin{figure}[ht]
\vspace{-3.75cm}
    \begin{subfigure}{0.49\linewidth}
        \centering
        \includegraphics[width=\linewidth]{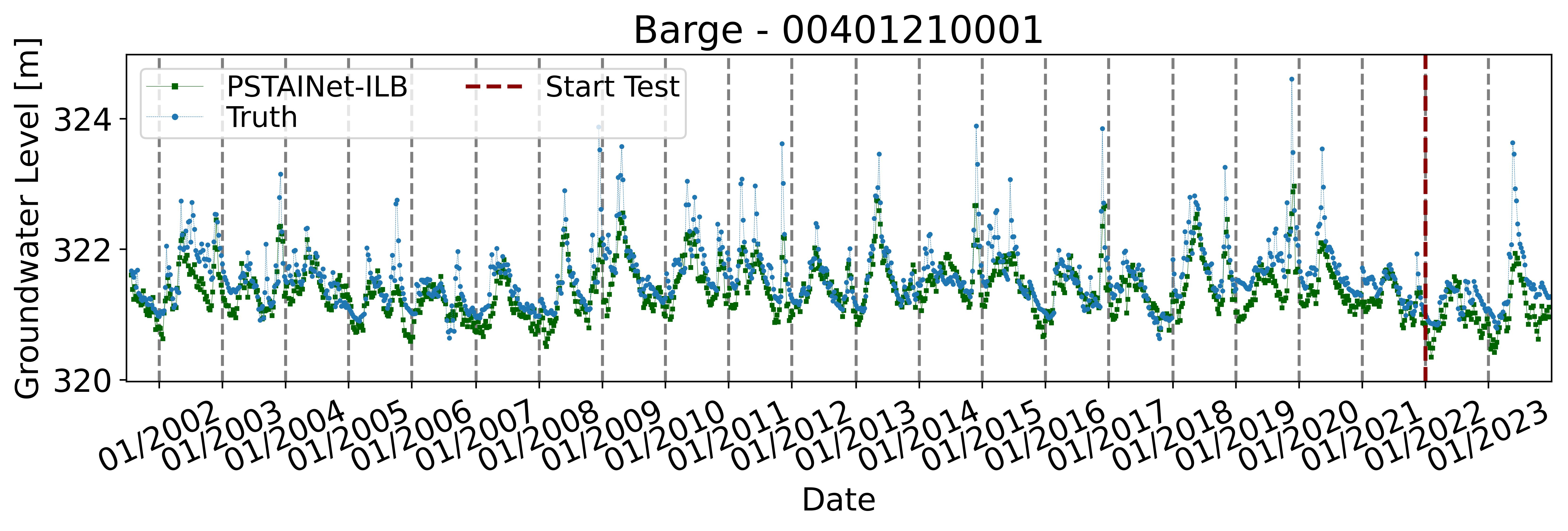}
        \caption{}
        \label{fig:bar_rec}
    \end{subfigure}
    \hfill
    \begin{subfigure}{0.49\linewidth}
        \centering
        \includegraphics[width=\linewidth]{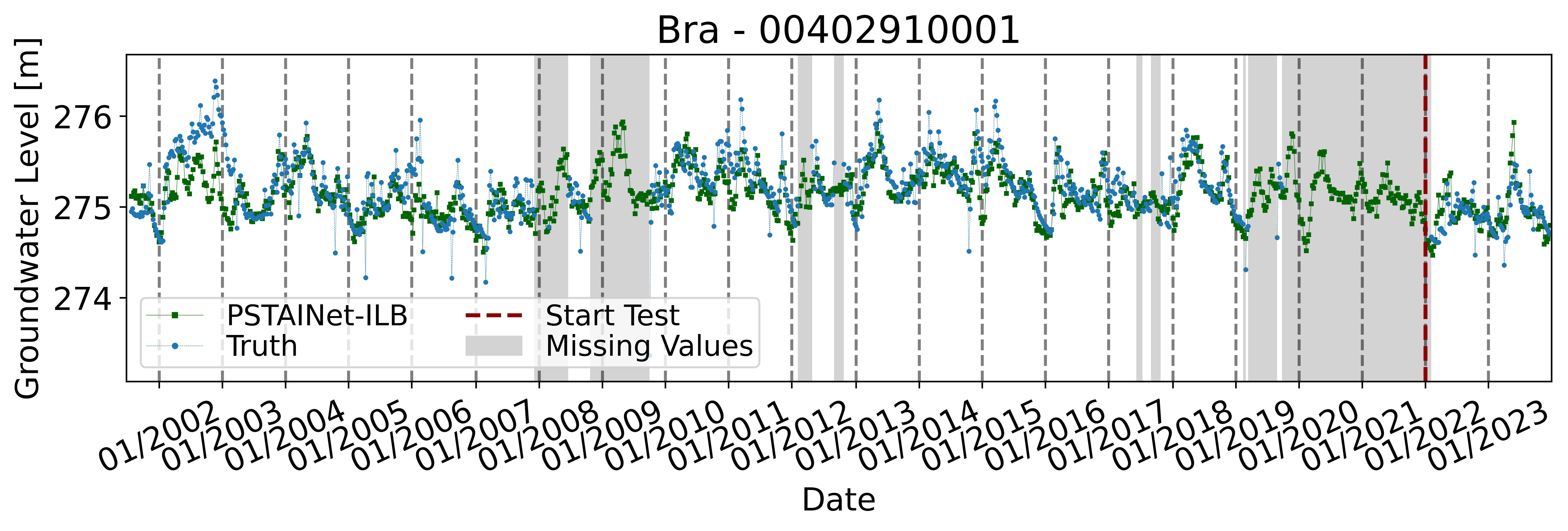}
        \caption{}
        \label{fig:bra_rec}
    \end{subfigure}
    \hfill
    \begin{subfigure}{0.49\linewidth}
        \centering
        \includegraphics[width=\linewidth]{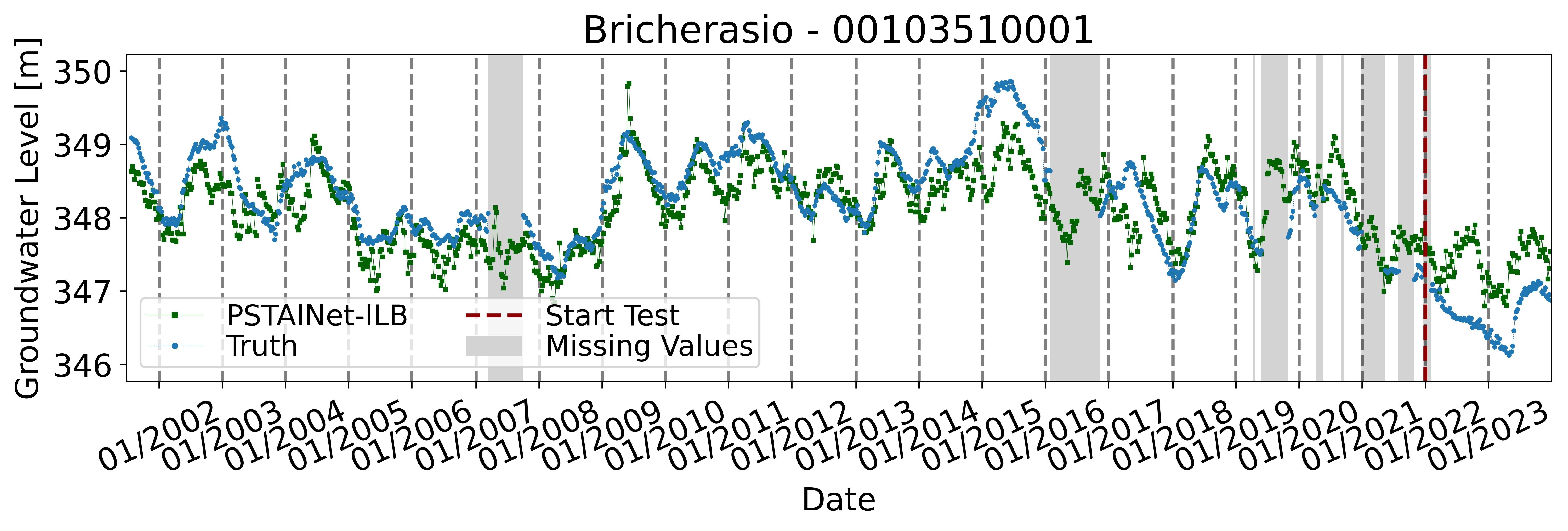}
        \caption{}
        \label{fig:bri_rec}
    \end{subfigure}
    \hfill
    \begin{subfigure}{0.49\linewidth}
        \centering
        \includegraphics[width=\linewidth]{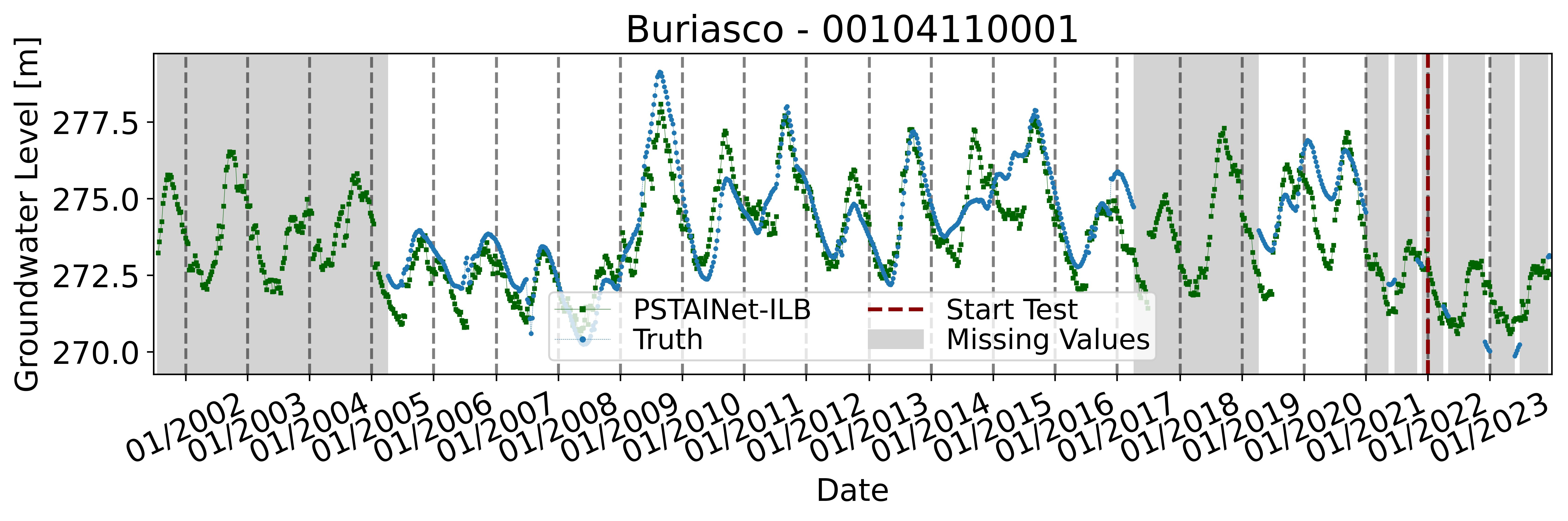}
        \caption{}
        \label{fig:bur_rec}
    \end{subfigure}
    \hfill
    \begin{subfigure}{0.49\linewidth}
        \centering
        \includegraphics[width=\linewidth]{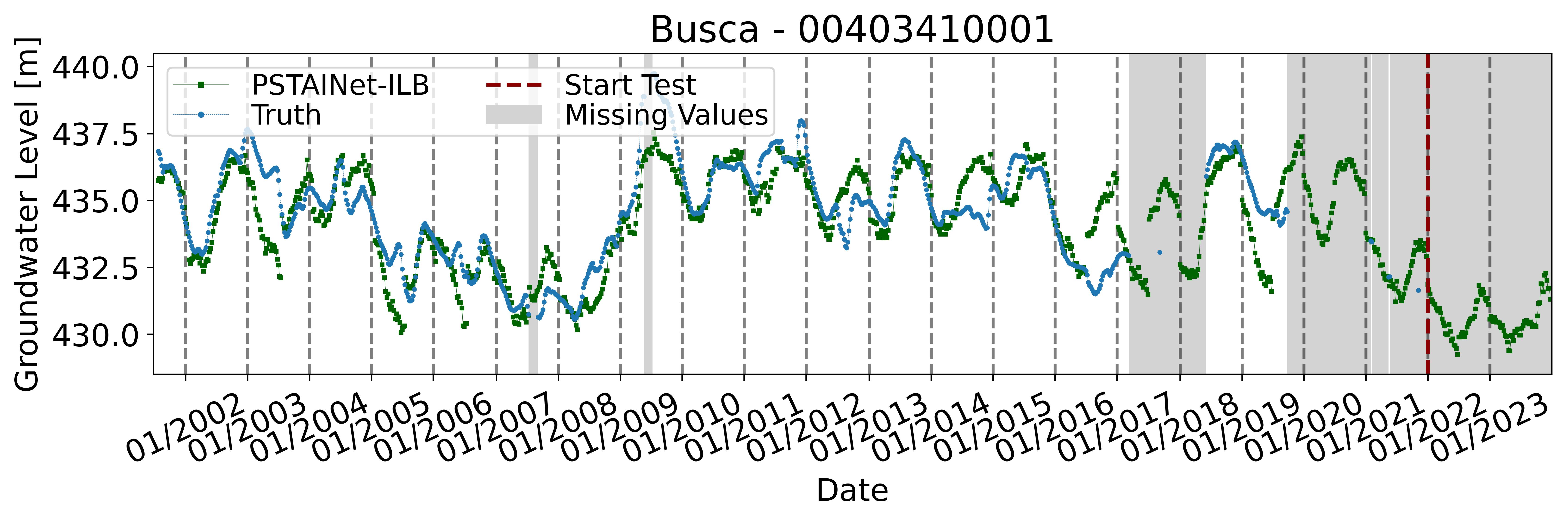}
        \caption{}
        \label{fig:busca_rec}
    \end{subfigure}
    \hfill
    \begin{subfigure}{0.49\linewidth}
        \centering
        \includegraphics[width=\linewidth]{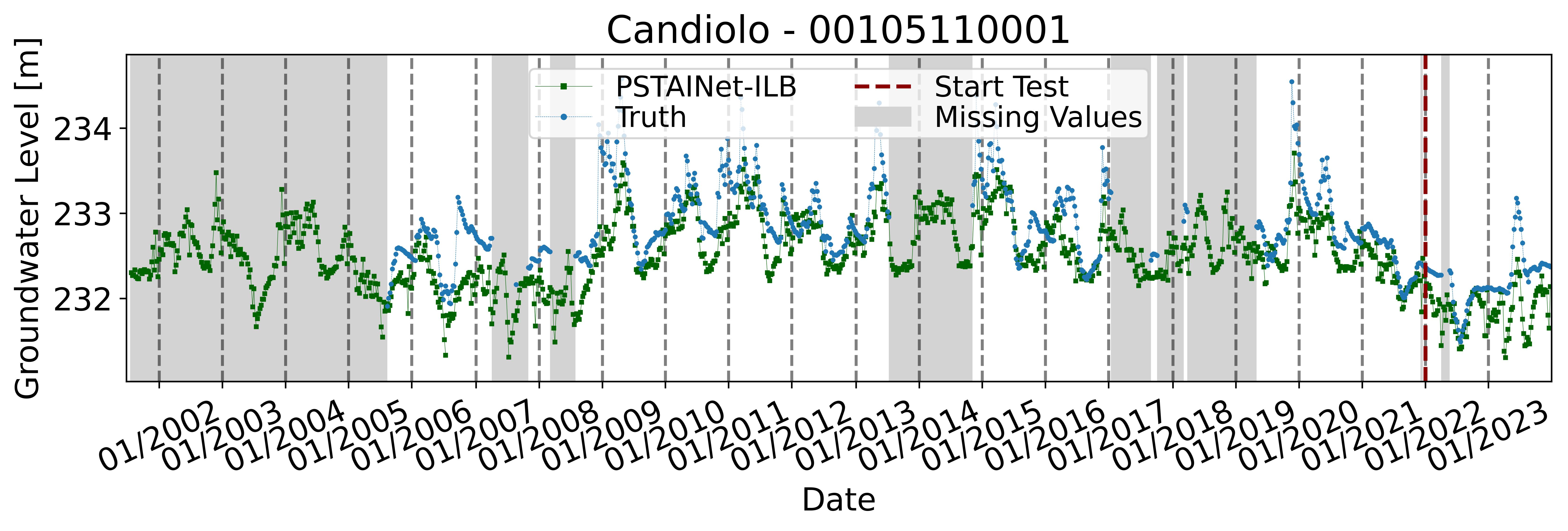}
        \caption{}
        \label{fig:cand_rec}
    \end{subfigure}
    \hfill
    \begin{subfigure}{0.49\linewidth}
        \centering
        \includegraphics[width=\linewidth]{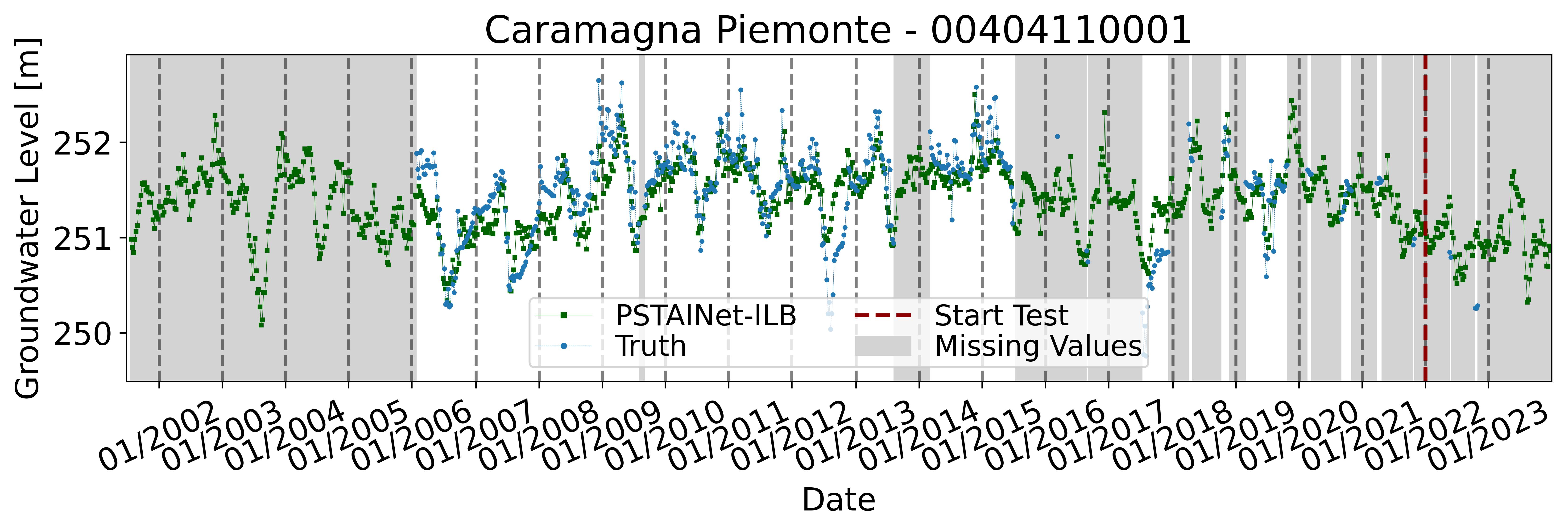}
        \caption{}
        \label{fig:cara_rec}
    \end{subfigure}
    \hfill
    \begin{subfigure}{0.49\linewidth}
        \centering
        \includegraphics[width=\linewidth]{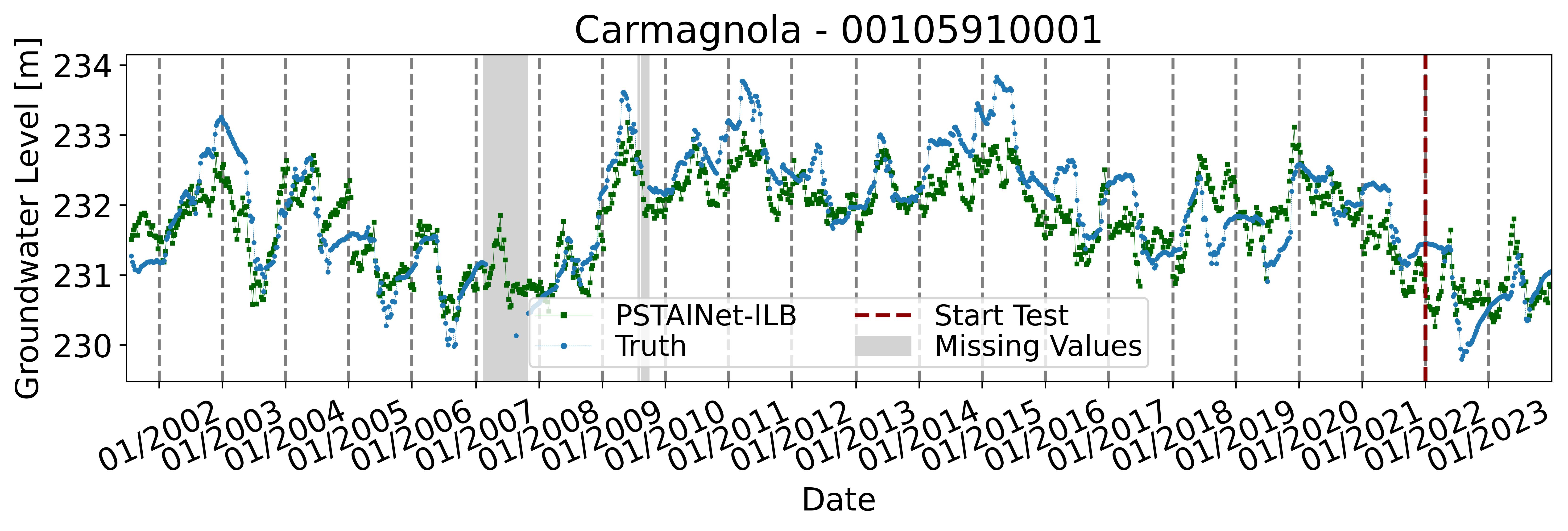}
        \caption{}
        \label{fig:carm1_rec}
    \end{subfigure}
\end{figure}
\end{landscape}

\begin{landscape}
\begin{figure}[ht]
\vspace{-3.75cm}
    \addtocounter{figure}{-1}  
    \begin{subfigure}{0.49\linewidth}
        \centering
        \addtocounter{subfigure}{8} 
        \includegraphics[width=\linewidth]{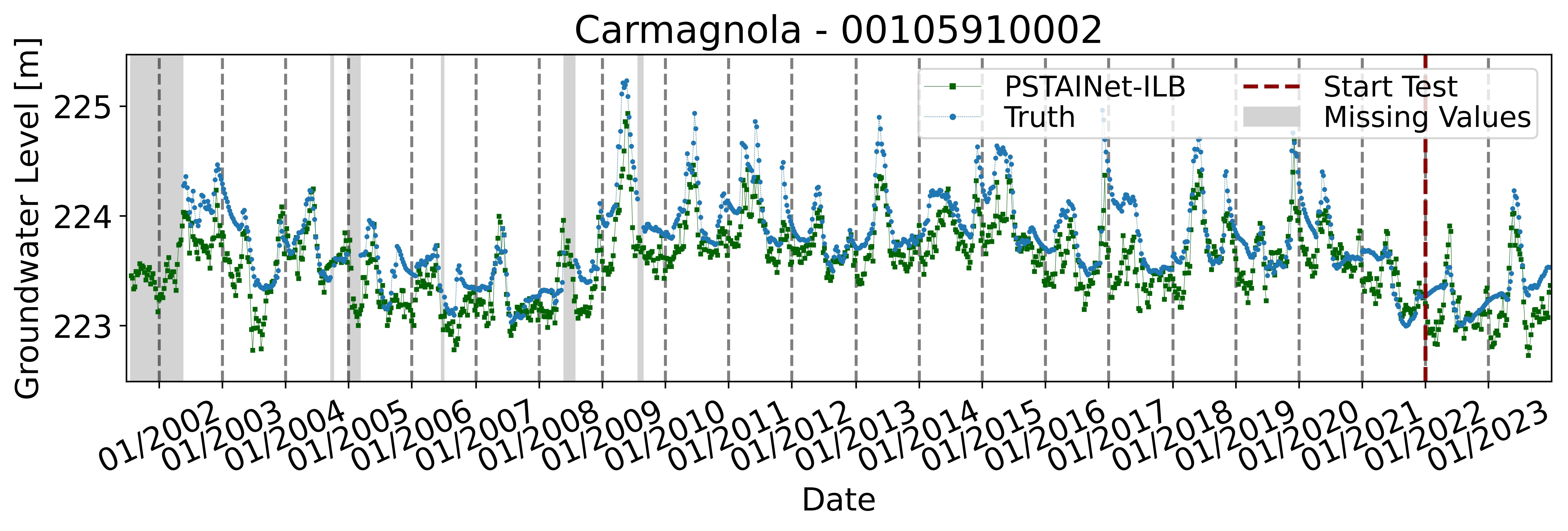}
        \caption{}
        \label{fig:carm2_rec}
    \end{subfigure}
    \hfill
    \begin{subfigure}{0.49\linewidth}
        \centering
        \includegraphics[width=\linewidth]{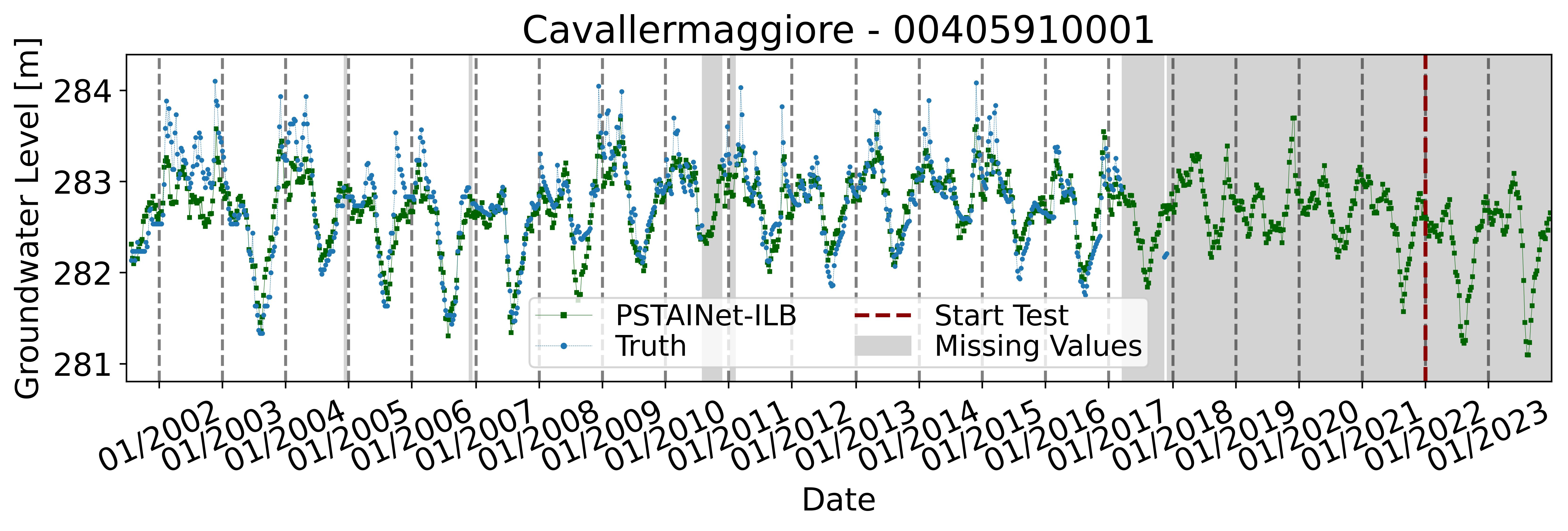}
        \caption{}
        \label{fig:cava_rec}
    \end{subfigure}
    \hfill
    \begin{subfigure}{0.49\linewidth}
        \centering
        \includegraphics[width=\linewidth]{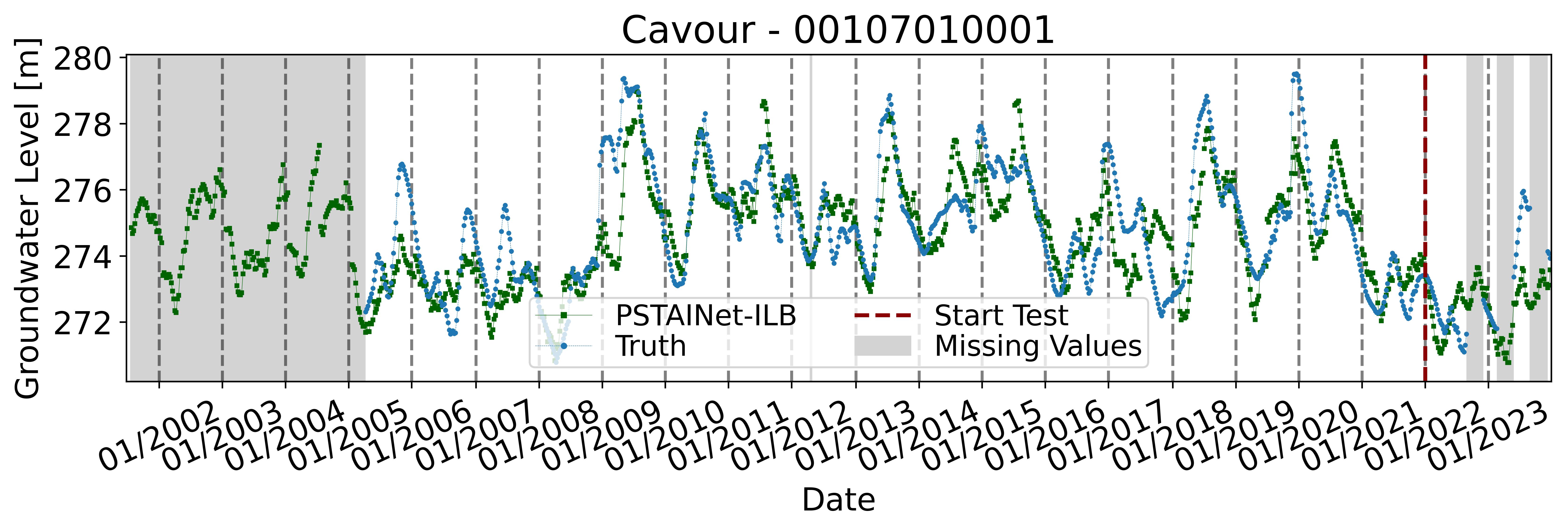}
        \caption{}
        \label{fig:cavo_rec}
    \end{subfigure}
    \hfill
    \begin{subfigure}{0.49\linewidth}
        \centering
        \includegraphics[width=\linewidth]{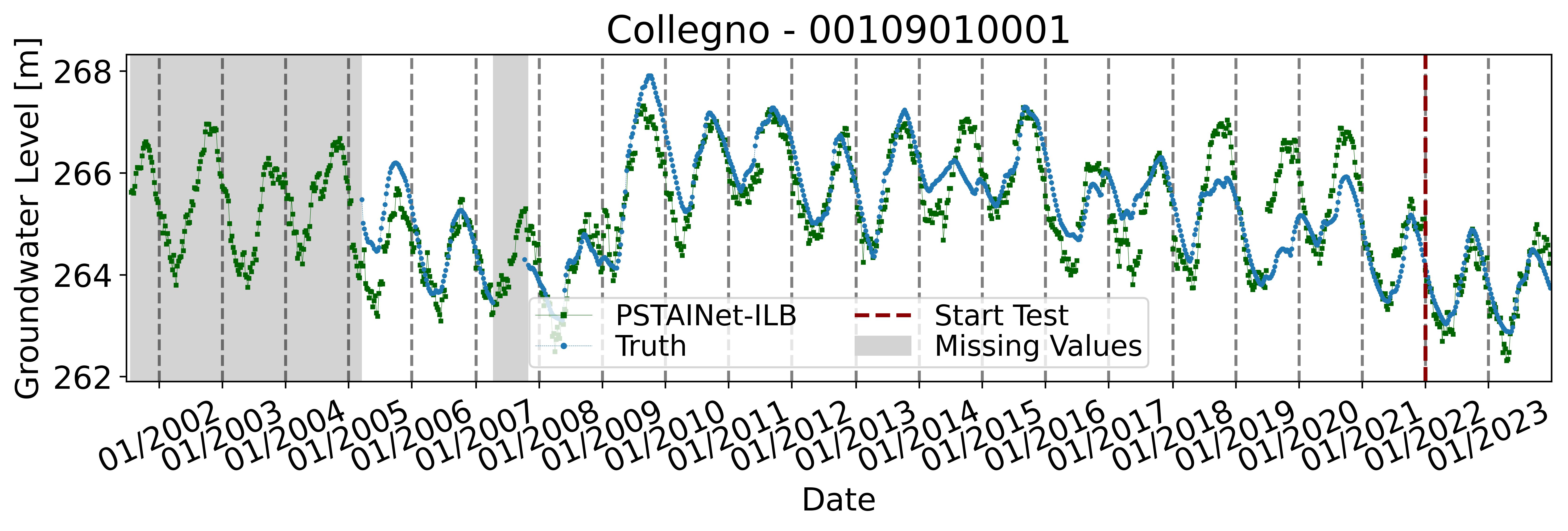}
        \caption{}
        \label{fig:coll_rec}
    \end{subfigure}
    \hfill
    \begin{subfigure}{0.49\linewidth}
        \centering
        \includegraphics[width=\linewidth]{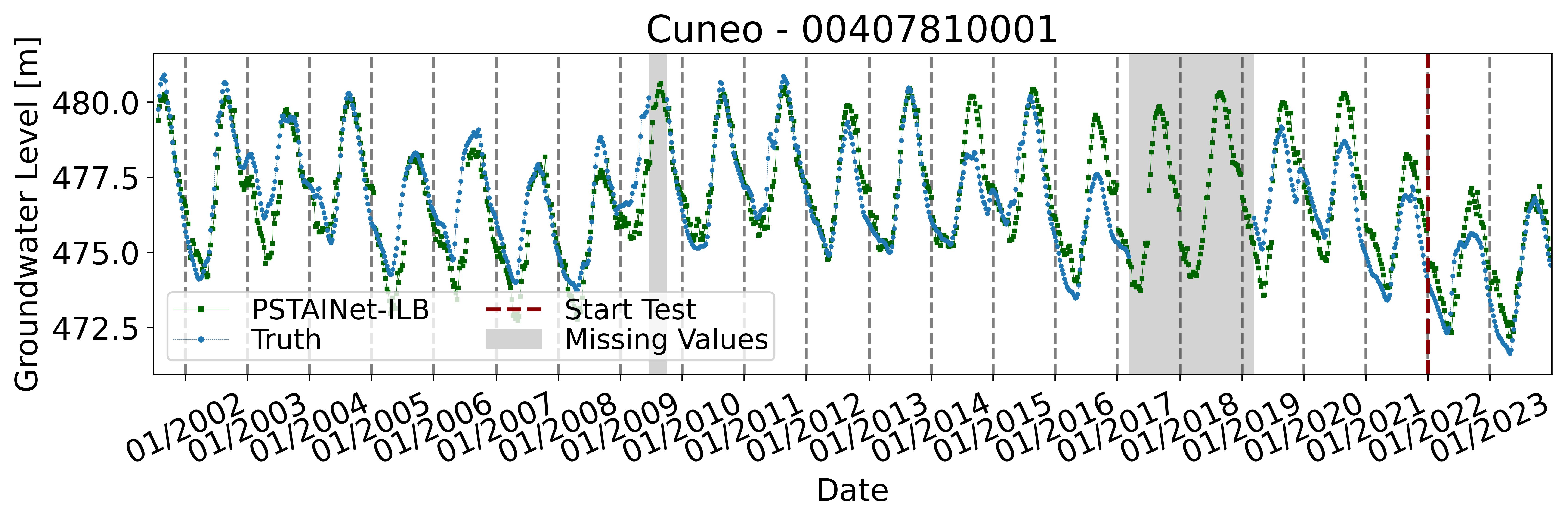}
        \caption{}
        \label{fig:cuneo_rec}
    \end{subfigure}
    \hfill
    \begin{subfigure}{0.49\linewidth}
        \centering
        \includegraphics[width=\linewidth]{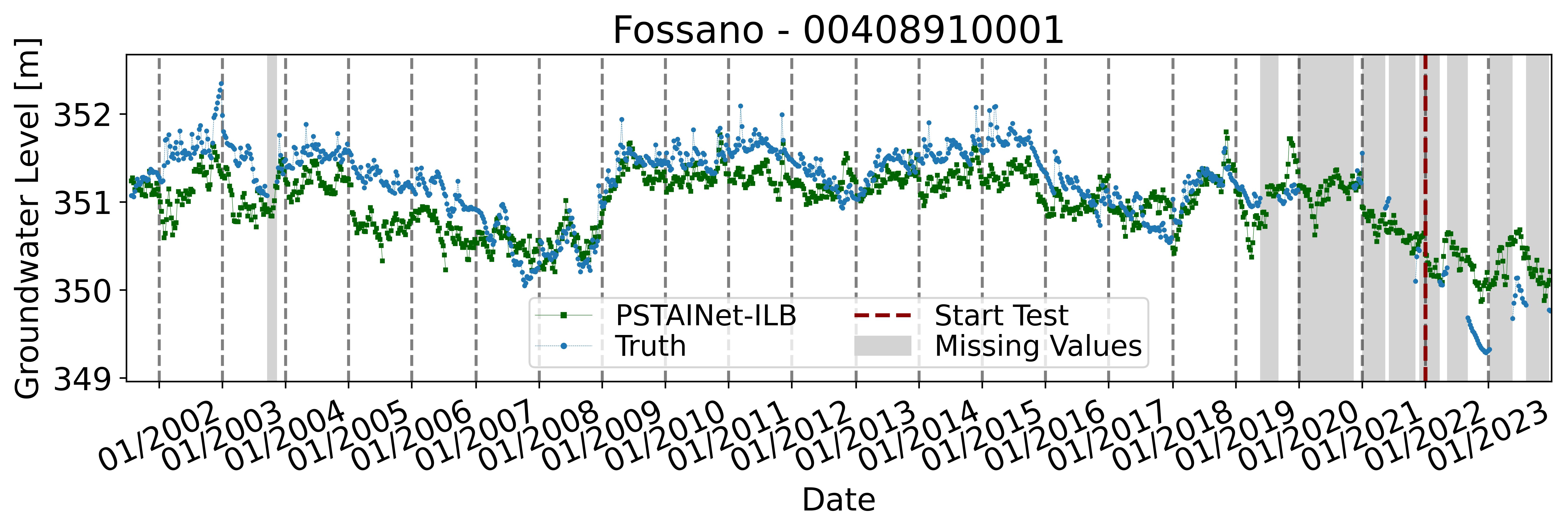}
        \caption{}
        \label{fig:foss1_rec}
    \end{subfigure}
    \hfill
    \begin{subfigure}{0.49\linewidth}
        \centering
        \includegraphics[width=\linewidth]{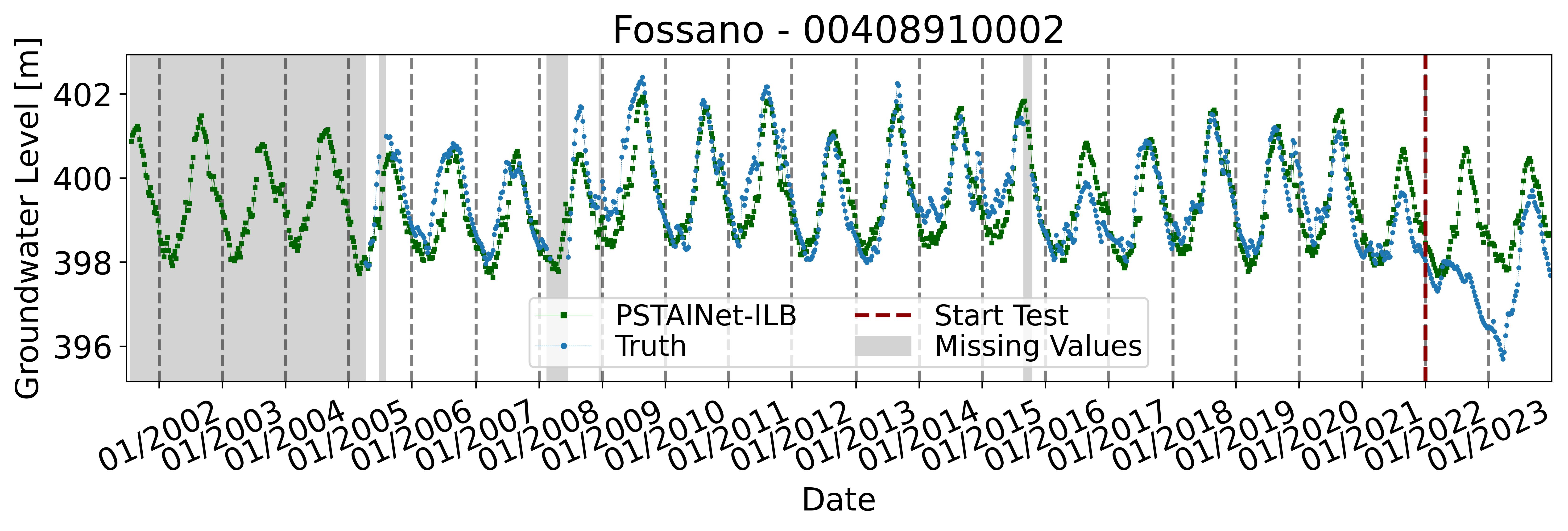}
        \caption{}
        \label{fig:foss2_rec}
    \end{subfigure}
    \hfill
    \begin{subfigure}{0.49\linewidth}
        \centering
        \includegraphics[width=\linewidth]{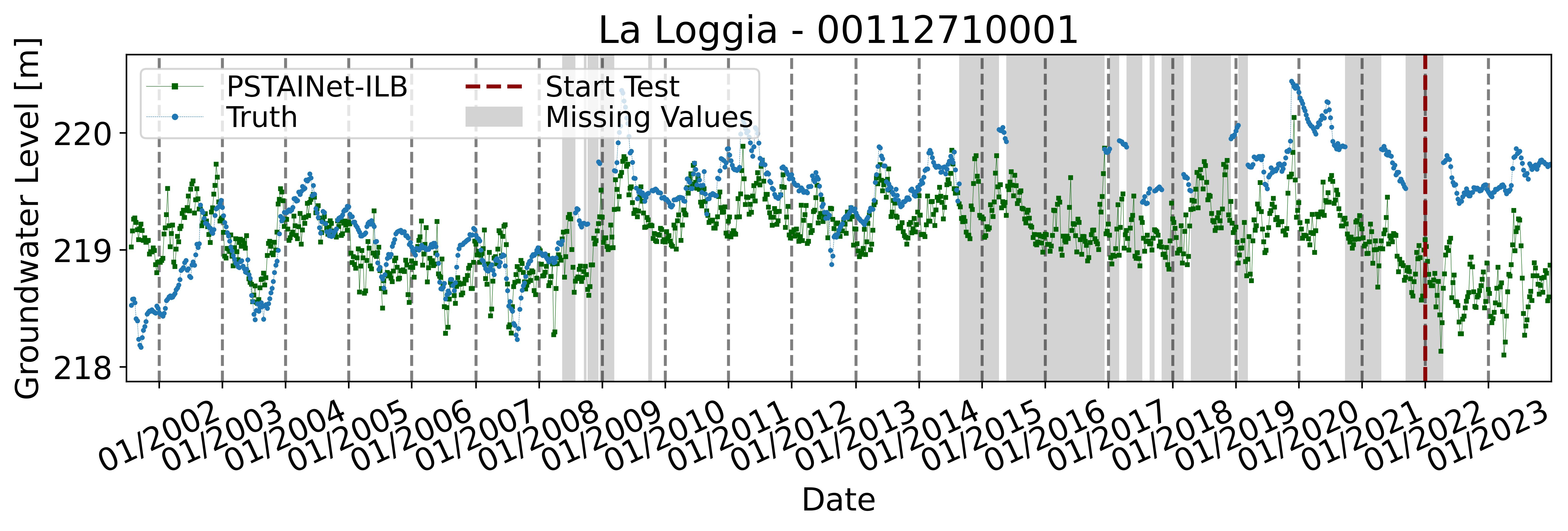}
        \caption{}
        \label{fig:lal_rec}
    \end{subfigure}
    \hfill
\end{figure}
\end{landscape}

\begin{landscape}
\begin{figure}[ht]
\vspace{-3.75cm}
    \addtocounter{figure}{-1}  
    \begin{subfigure}{0.49\linewidth}
        \centering
        \addtocounter{subfigure}{16} 
        \includegraphics[width=\linewidth]{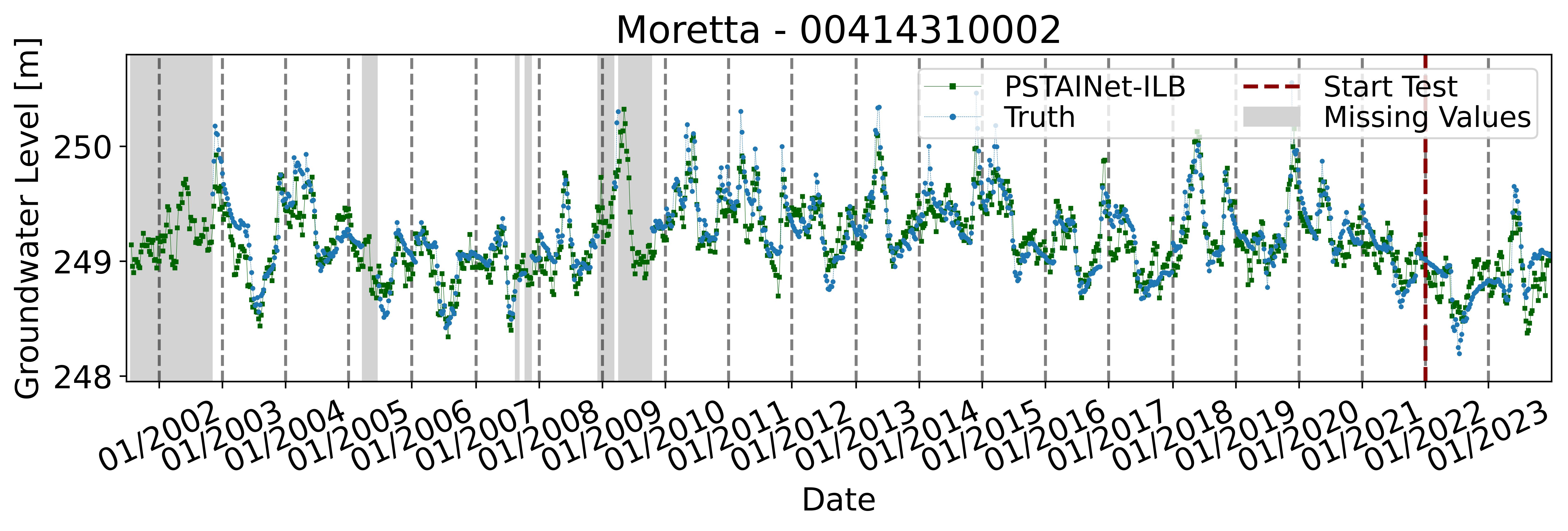}
        \caption{}
        \label{fig:mor_rec}
    \end{subfigure}
    \hfill
    \begin{subfigure}{0.49\linewidth}
        \centering
        \includegraphics[width=\linewidth]{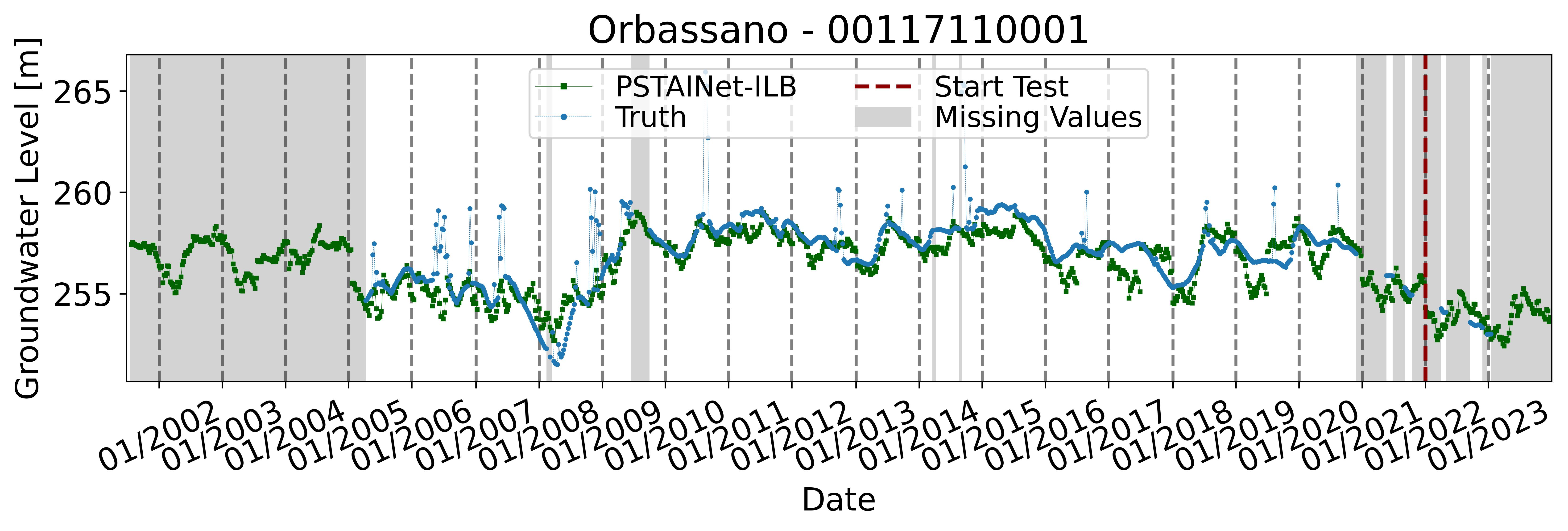}
        \caption{}
        \label{fig:orb_rec}
    \end{subfigure}
    \hfill
    \begin{subfigure}{0.49\linewidth}
        \centering
        \includegraphics[width=\linewidth]{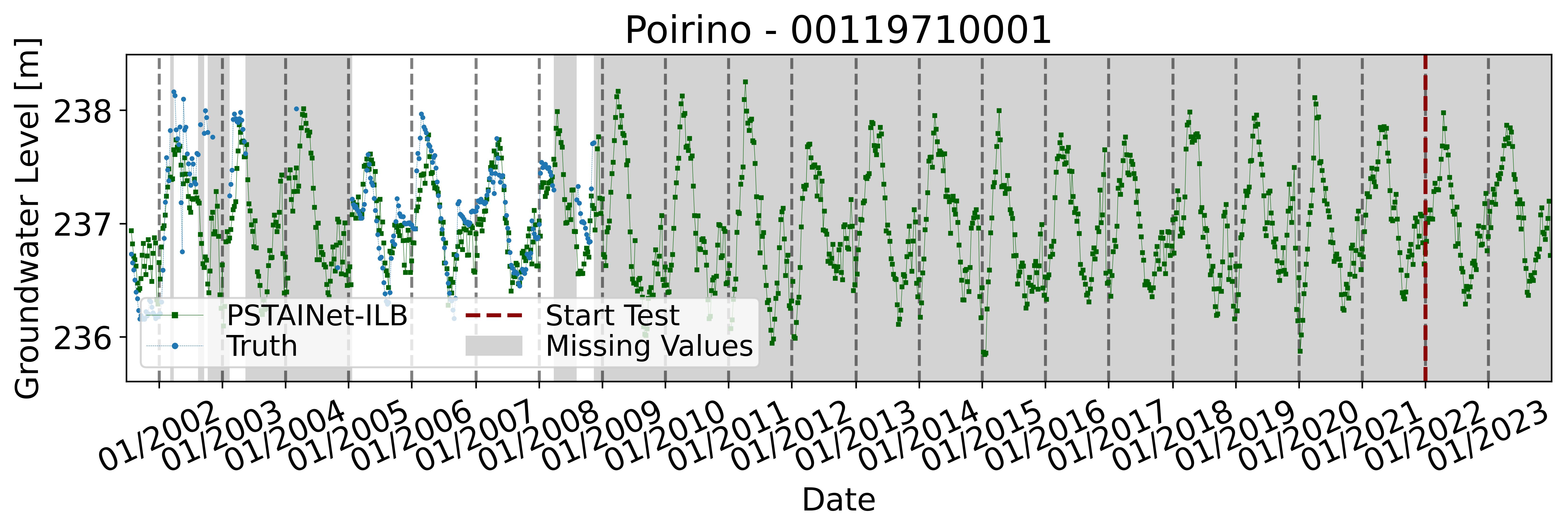}
        \caption{}
        \label{fig:poi_rec}
    \end{subfigure}
    \hfill
    \begin{subfigure}{0.49\linewidth}
        \centering
        \includegraphics[width=\linewidth]{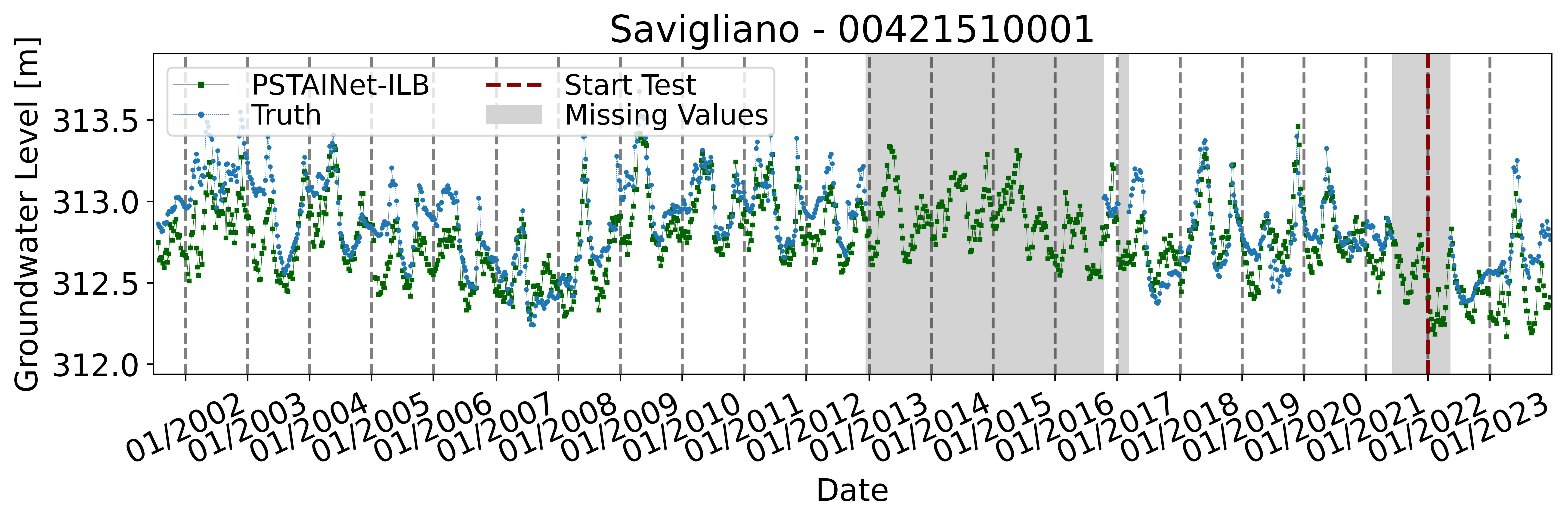}
        \caption{}
        \label{fig:sav_rec}
    \end{subfigure}
    \hfill
    \begin{subfigure}{0.49\linewidth}
        \centering
        \includegraphics[width=\linewidth]{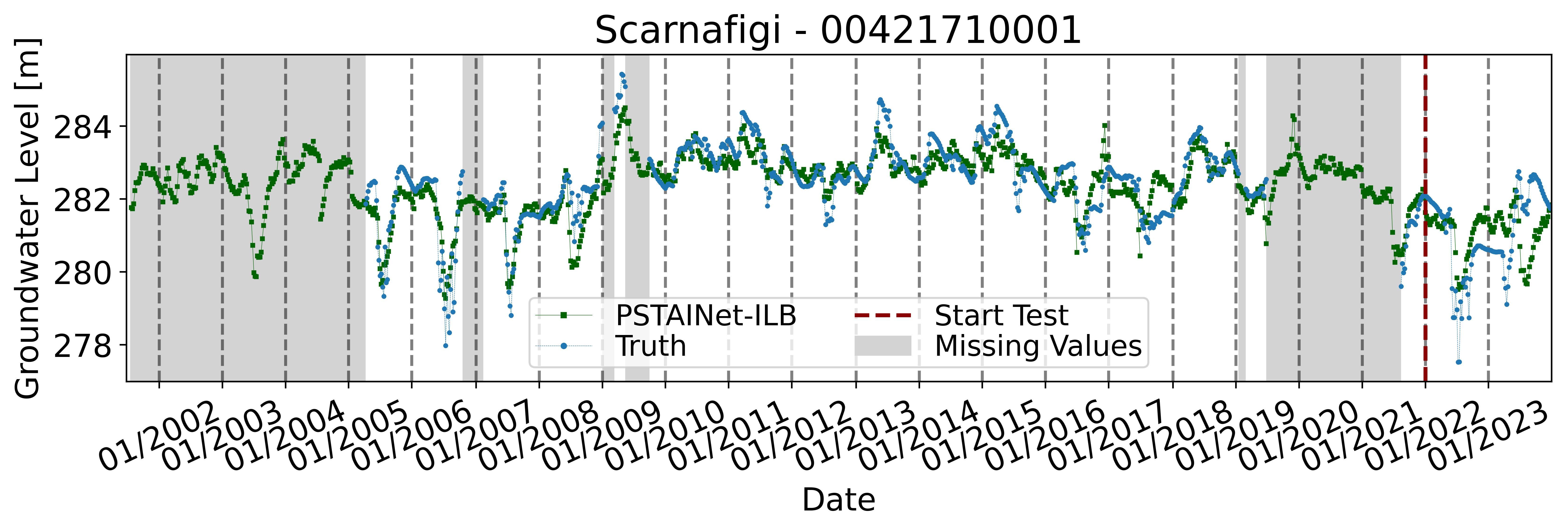}
        \caption{}
        \label{fig:scarn_rec}
    \end{subfigure}
    \hfill
    \begin{subfigure}{0.49\linewidth}
        \centering
        \includegraphics[width=\linewidth]{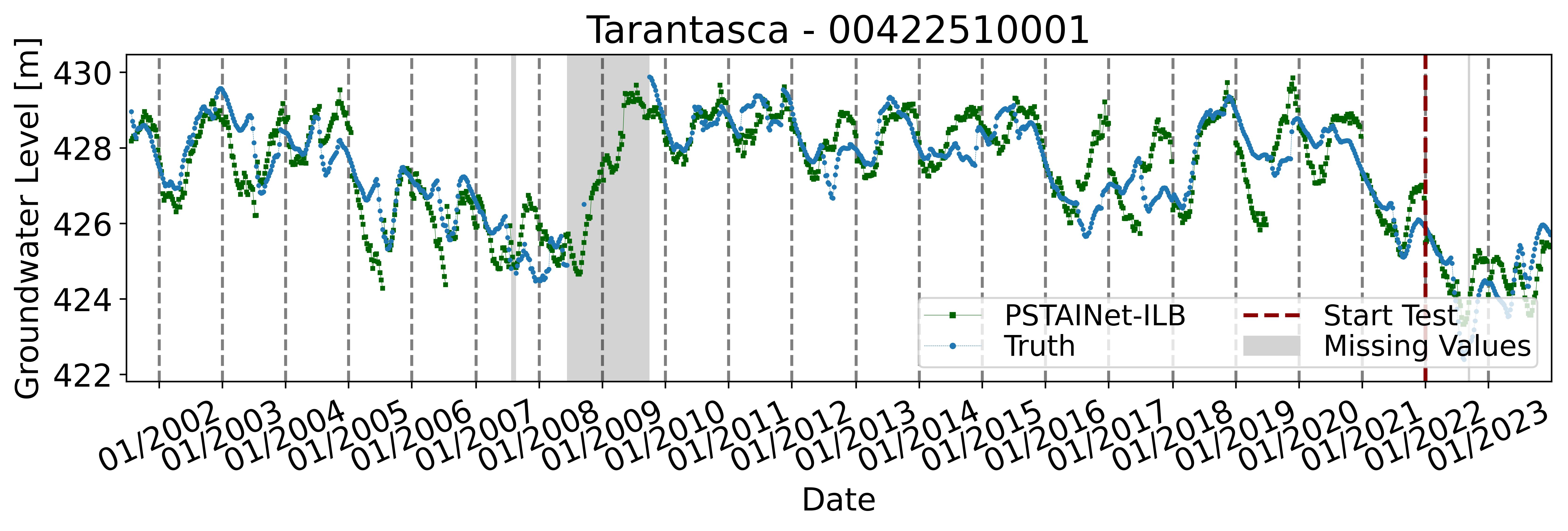}
        \caption{}
        \label{fig:tara_rec}
    \end{subfigure}
    \hfill
    \begin{subfigure}{0.49\linewidth}
        \centering
        \includegraphics[width=\linewidth]{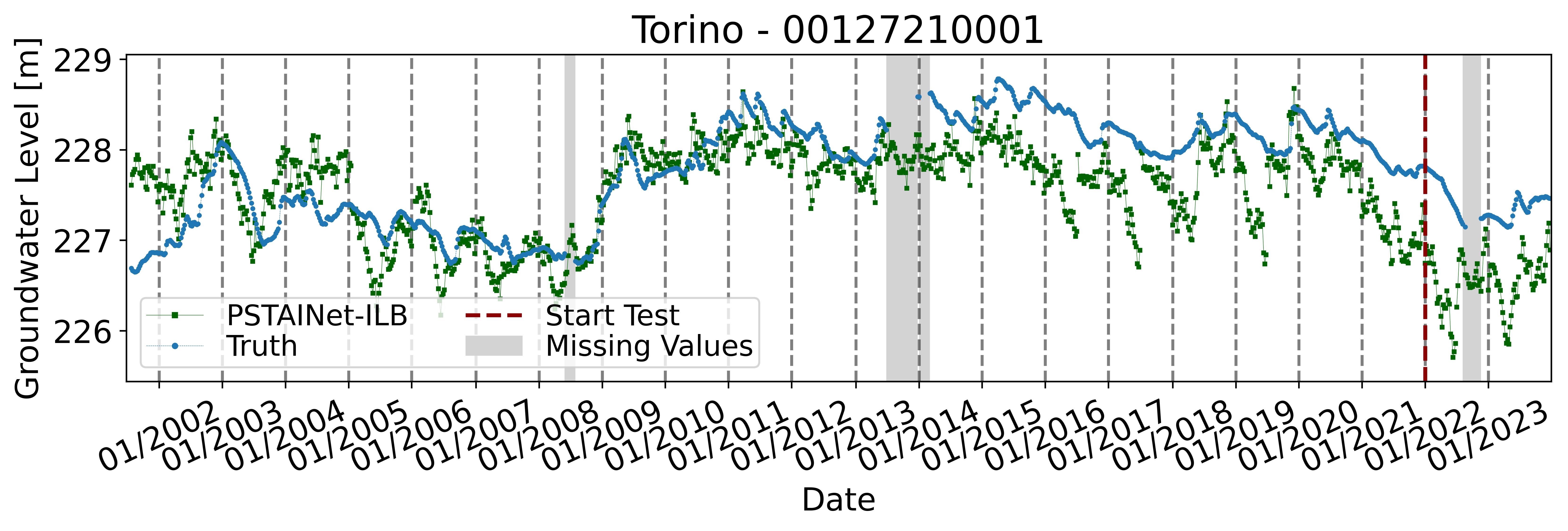}
        \caption{}
        \label{fig:tor1_rec}
    \end{subfigure}
    \hfill
    \begin{subfigure}{0.49\linewidth}
        \centering
        \includegraphics[width=\linewidth]{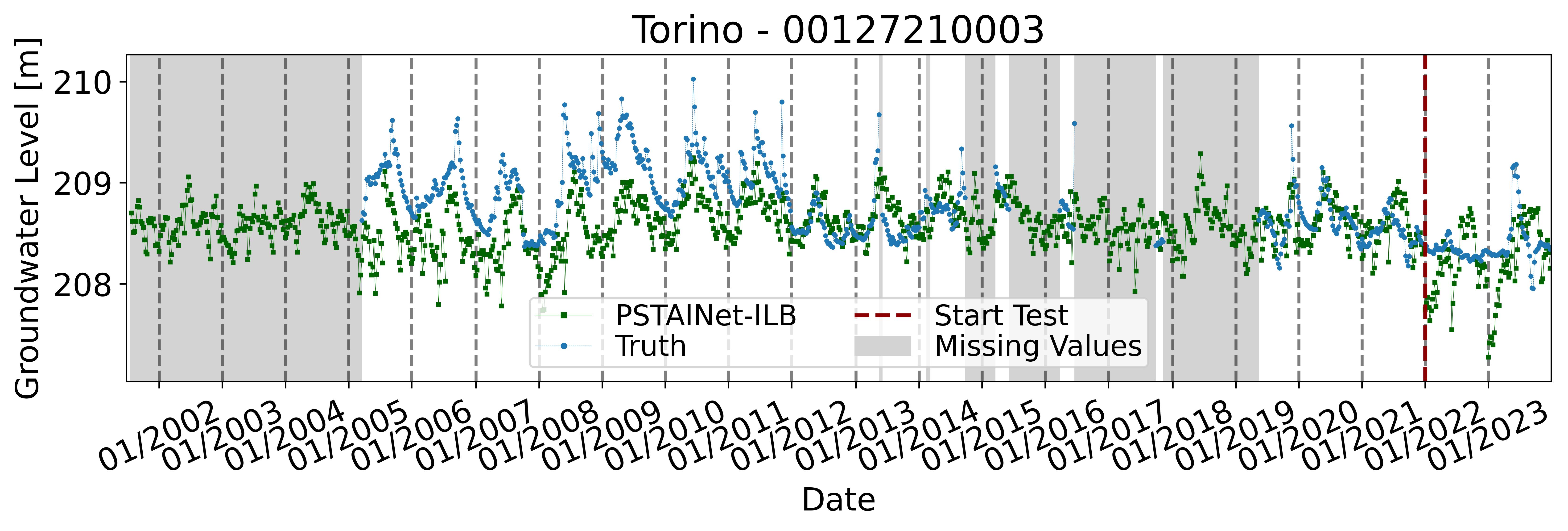}
        \caption{}
        \label{fig:tor3_rec}
    \end{subfigure}
    \hfill
\end{figure}
\end{landscape}

\begin{landscape}
\begin{figure}[ht]
    \addtocounter{figure}{-1}  
    \begin{subfigure}{0.49\linewidth}
        \centering
        \addtocounter{subfigure}{24} 
        \includegraphics[width=\linewidth]{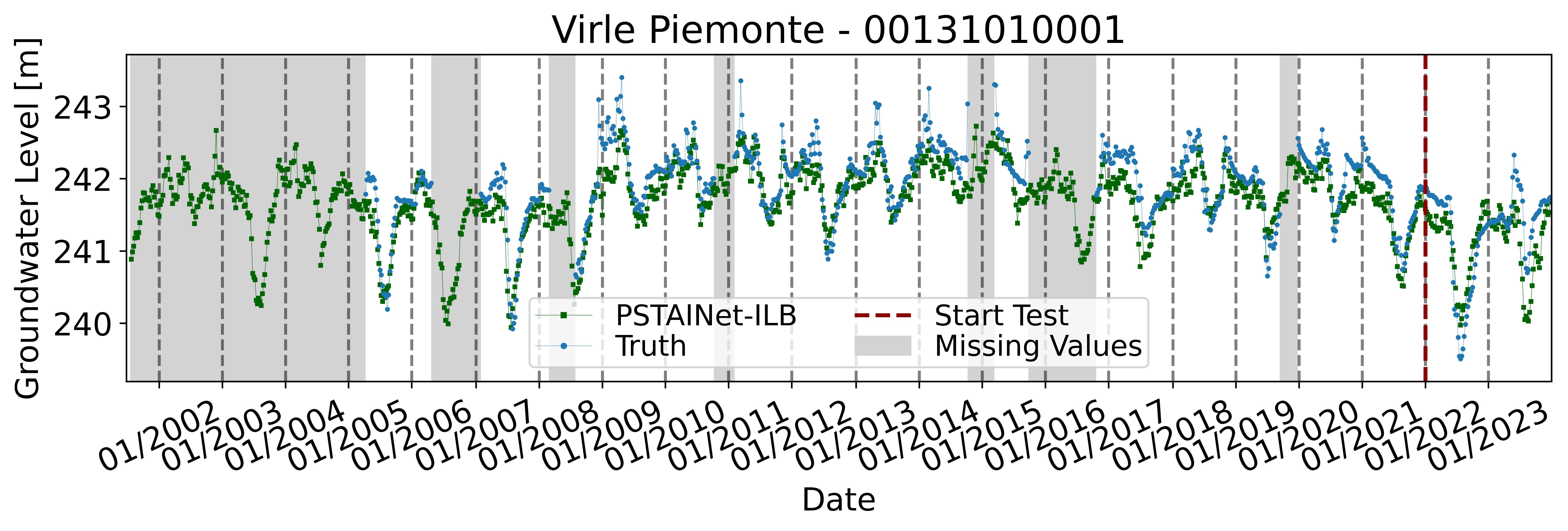}
        \caption{}
        \label{fig:vir_rec}
    \end{subfigure}
    \hfill
    \begin{subfigure}{0.49\linewidth}
        \centering
        \includegraphics[width=\linewidth]{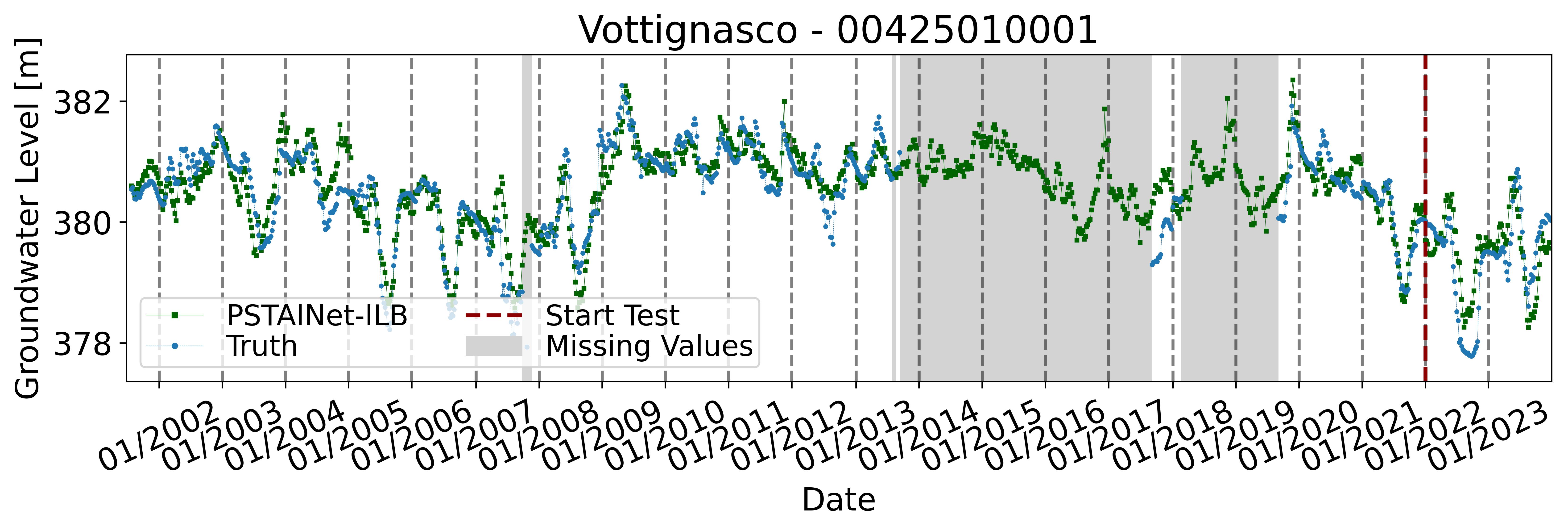}
        \caption{}
        \label{fig:vott_rec}
    \end{subfigure}
\caption{Test set predictions in the rollout setting.}
\label{fig:all_rec}
\end{figure}
\end{landscape}

\end{document}